%% file: main.tex
\documentclass{article}
\PassOptionsToPackage{numbers,square,sort&compress}{natbib}
\usepackage[main,preprint]{neurips_2026}

\usepackage[utf8]{inputenc}
\usepackage[T1]{fontenc}
\usepackage{microtype}
\usepackage{amsmath,amssymb,amsfonts,mathtools,amsthm}
\usepackage{graphicx}
\usepackage{booktabs}
\usepackage{array}
\usepackage{tabularx}
\usepackage{longtable}
\usepackage{subcaption}
\usepackage{float}
\usepackage{placeins}
\usepackage{algorithm}
\usepackage{algpseudocode}
\usepackage{xcolor}
\usepackage{enumitem}
\usepackage{hyperref}
\usepackage{url}
\usepackage{multirow}
\usepackage{caption}
\usepackage{pifont}
\usepackage{wrapfig}
\usepackage{capt-of}
\usepackage{needspace}
\usepackage{colortbl}
\usepackage{ragged2e}
\usepackage{pifont}
\usepackage{dblfloatfix}

\definecolor{CausalBlue}{RGB}{226,241,255}
\definecolor{CausalRow}{gray}{0.95}

\hypersetup{colorlinks=true,citecolor=blue,linkcolor=blue,urlcolor=blue}
\allowdisplaybreaks


\hypersetup{
    colorlinks=false,
    pdfborder={0 0 1},
    citebordercolor={0 1 0}, 
    linkbordercolor={1 0 0}, 
    urlbordercolor={0 0 1}   
}

\title{Online Causal Configuration for Networked Systems via Doubly Robust Steady-State Learning}

\author{
Yuli Liu$^{3}$ \qquad
Zhiheng Zhang$^{1,2}$\thanks{
Accepted at NeurIPS 2026. Corresponding author: Zhiheng Zhang.
Email: \texttt{zhangzhiheng@mail.shufe.edu.cn}.
Project: \url{https://ylliua1116.github.io/causal-ucb/}.
Code: \url{https://github.com/ylliua1116/causal-ucb}.
}\\[0.5em]
\normalfont$^{1}$ School of Statistics and Data Science, Shanghai University of Finance and Economics\\
\normalfont$^{2}$ Institute of Big Data Research, Shanghai University of Finance and Economics\\
\normalfont Shanghai 200433, P.R. China\\
\normalfont$^{3}$ Smart Ocean Technology, University of Macau, Macao SAR, China
}

\newcommand{\Winf}{W_{\infty}}

\newtheorem{theorem}{Theorem}
\newtheorem{lemma}{Lemma}
\newtheorem{proposition}{Proposition}
\newtheorem{corollary}{Corollary}

\begin{document}
\maketitle

\raggedbottom

\begin{abstract}
Measurement exploration reveals how interacting local controllers behave but can alter the steady-state regime a network operator seeks to optimize. Exploratory rewards may therefore fail to identify a configuration's no-exploration value: under full exploration, identical logged distributions can conceal different target values. Identification requires accounting for both action-law mismatch and shifts in the steady-state context distribution.
We introduce CAUSAL-UCB, which selects configurations using optimistic bounds on target causal values. Its interference-aware clipped doubly robust evaluator produces predecision certificates separating statistical variation from calibrated identification, transport, and temporal-dependence errors.
For any finite horizon and predetermined schedule of measurement-exploration rates $\{\eta_t\}$, we prove lower bounds of $\Omega((1+\sum_t\eta_t)^{-1/2})$ on minimax inference error and $\Omega(\sum_t\eta_t)$ on measurement-induced deployment loss in a stationary, potentially slow-mixing family. We bound regret under valid certificates and selected-information growth, and establish an anytime finite-sample certificate with a shared radius for estimation, selection, and reporting in a restricted known-local-interference model.
In a high-mismatch benchmark with 24 paired seeds and 200 rounds, CAUSAL-UCB attains target regret of $1.732$ versus $6.326$ for certified IPS-UCB, with deployment gaps of $30.607$ and $23.217$, respectively. Independent-audit coverage tests and trace-driven semi-synthetic studies further assess certificate coverage and deployment trade-offs.
\end{abstract}

\section{Introduction}

At each slow decision round, a network operator chooses a global configuration, while adaptive, interacting local controllers determine the realized actions and settle into a steady state.
The deployment objective is the configuration's no-exploration steady-state utility, not the reward observed under the measurement policy used to collect data.
Exploration can improve action support, but it also changes local actions and post-burn-in contexts.
Consequently, a configuration that performs well under exploratory logging need not maximize deployment utility.

Different controllers can generate identical exploratory observations while having different no-exploration steady-state values (Proposition~\ref{prop:v8-nonid}).
Identifying the target value therefore requires information beyond an accurate outcome model.
Evaluation requires an appropriate exposure representation for network interference, correction for target-action mismatch, and a justified link between exploratory and target context distributions.
In particular, action overlap supports comparisons conditional on observed contexts, but does not by itself identify the context distribution that would arise without exploration.
These requirements must be distinguished when evaluating configurations and making online choices.

\begin{figure}[!htbp]
\centering
\includegraphics[width=.99\linewidth]{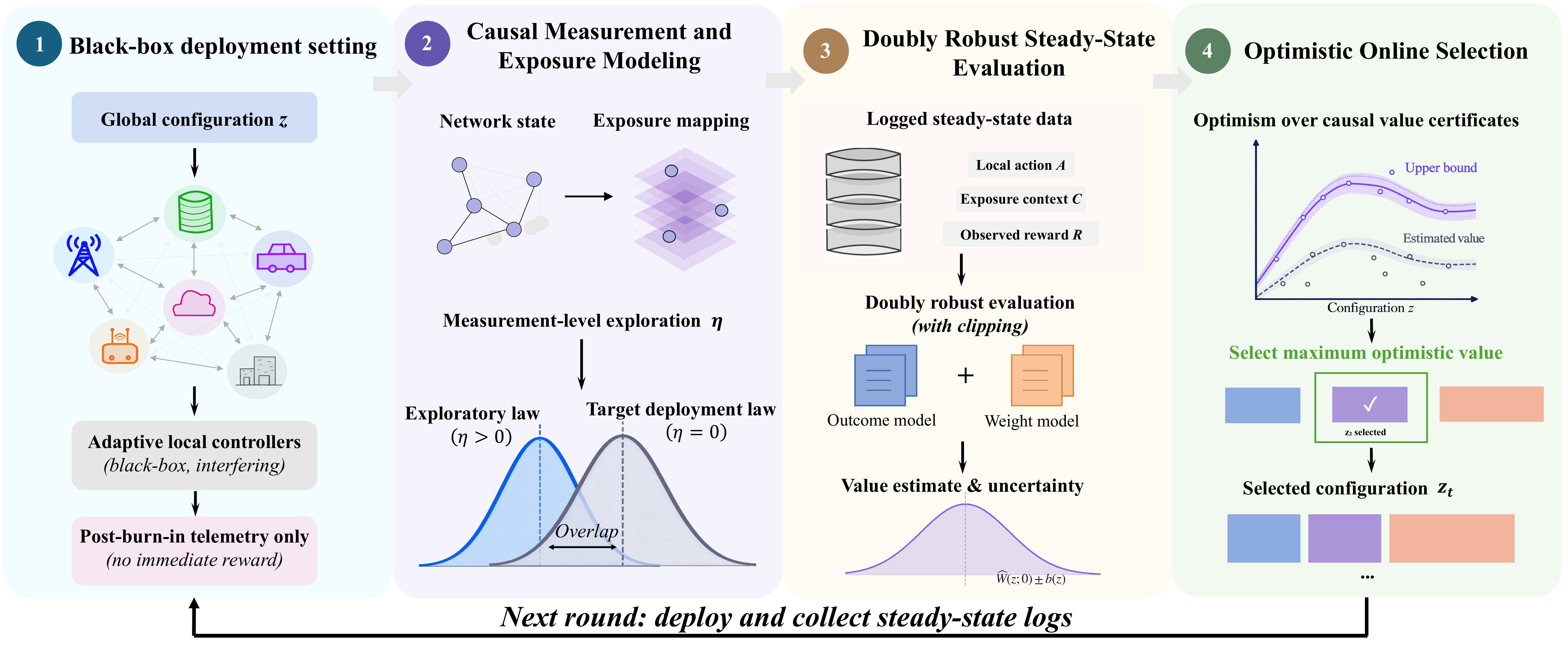}
\caption{\textbf{The CAUSAL-UCB deployment loop.}
A global configuration induces interacting local actions; post-burn-in measurements enter an exposure-aware evaluator, whose value certificate guides the next optimistic configuration choice. Visual elements are conceptual.}
\label{fig:method-overview}
\end{figure}

Our setting connects causal inference under network
interference~\cite{aronow2017estimating,hudgens2008toward,savje2021average},
doubly robust off-policy evaluation and logged-feedback
learning~\cite{dudik2011doubly,wang2017optimal,swaminathan2015counterfactual},
and online decision-making with network
interference~\cite{agarwal2024mabn,jia2025mabi,wang2025design,zhang2025online}.
These research lines provide tools for representing interference, correcting action-law mismatch, and guiding sequential decisions.
Our setting differs in both the decision unit and the target: the operator selects a global configuration, while its value is defined after black-box controllers settle under the no-exploration law.
The resulting problem is to use exploratory
steady-state telemetry for online selection
while accounting for the distinct measurement
and deployment regimes.

We address this problem with CAUSAL-UCB.
An exposure-aware clipped doubly robust evaluator estimates each configuration's target value, and a predecision certificate separates sampling fluctuation from identification, transport, and temporal-dependence errors.
The certificate is used to select configurations by optimistic target values rather than raw exploratory rewards.
The corresponding guarantees rely on the
calibration conditions developed in Section~\ref{sec:certificate-main}.
This separation makes explicit which uncertainties are estimated from logged data and which require additional assumptions or independent information.

Measurement also has two distinct consequences: it affects the information available for identifying target values and can perturb the state whose utility is being optimized.
We formalize this identification--deployment tension through a minimax lower bound for predetermined exploration schedules in a stationary, potentially slow-mixing family.
Separately, we establish a regret guarantee for online selection when the certificates are valid and selected information grows.
Under known local interference, we also derive an anytime finite-sample certificate with a shared predecision radius for estimation, selection, and reporting.

\noindent\textbf{Contributions.} \textbf{(1)} We formalize online causal configuration around the no-exploration steady-state estimand and derive an exposure-aware doubly robust evaluator with a uniform certificate separating sampling fluctuation from identification, transport, and temporal-dependence errors.
\textbf{(2)} For every finite horizon and predetermined exploration schedule in the stated family, we prove minimax lower bounds $\Omega((1+\sum_t\eta_t)^{-1/2})$ for inference error and $\Omega(\sum_t\eta_t)$ for measurement-induced deployment loss.
We also establish a regret guarantee under valid certificates and selected-information growth.
\textbf{(3)} Under known local interference, we establish an anytime finite-sample certificate whose predecision radius is shared by estimation, selection, and reporting.

The remainder of the paper is organized as follows.
Section~\ref{sec:related-work} reviews related work;
Section~\ref{sec:system-model} formalizes the operational setting
and target estimand;
Section~\ref{sec:certificate-main} develops the estimator, certificates,
and theoretical guarantees;
Section~\ref{sec:new-experiments} evaluates decision quality,
identification--deployment trade-offs, and transfer across
network traces.
Section~\ref{sec:conclusion} concludes.
Additional proofs, experimental protocols, and robustness
analyses are provided in the appendices.

\section{Related Work}
\label{sec:related-work}

\noindent\textbf{\textcircled{\scriptsize 1}\ Causal inference with interference.} Potential-outcome methods formalize causal effects under
partial or general interference
\cite{aronow2017estimating,hudgens2008toward,savje2021average}.
Causal representation learning further investigates
interventional structure, temporal effects, and partial
observability
\cite{ahuja2023interventional,lippe2023citris,yao2024multiview}.
Recent design-based work studies covariate adjustment under approximate neighborhood
interference~\cite{lu2026adjusting}, but does not address exploration-induced shifts
in the steady-state context law. We explicitly budget exposure and transport error.

\noindent\textbf{\textcircled{\scriptsize 2}\ Off-policy evaluation and overlap.}
\par
\begin{wraptable}{r}{0.45\columnwidth}
\vspace{-40pt}
\centering
\caption{\textbf{Representative research scope.}}
\label{tab:related-scope}
\vspace{-5pt}
{%
\fontsize{6.8}{7.5}\selectfont
\setlength{\tabcolsep}{9.0pt}
\renewcommand{\arraystretch}{0.96}
\begin{tabular}{@{}lccccc@{}}
\toprule
Research line & I & L & G & S & R \\
\midrule
Interf. estimands
& $\checkmark$ & $\circ$ & -- & -- & -- \\
DR OPE
& -- & $\checkmark$ & -- & $\circ$ & -- \\
Network bandits
& $\checkmark$ & $\circ$ & $\circ$ & -- & $\checkmark$ \\
Design bandits
& $\checkmark$ & $\circ$ & $\circ$ & -- & $\checkmark$ \\
Temporal overlap
& -- & $\checkmark$ & -- & $\circ$ & -- \\
\midrule
\textbf{CAUSAL-UCB}
& $\checkmark$ & $\checkmark$ & $\checkmark$
& $\checkmark$ & $\checkmark$ \\
\bottomrule
\end{tabular}\par
}
\vspace{2pt}

\parbox{\linewidth}{%
\raggedright
\fontsize{6.0}{6.7}\selectfont
I: interference; L: logged evaluation;
G: global configuration; S: target steady state;
R: online cost/regret.
$\checkmark$: explicit; $\circ$: related;
--: outside scope.\par
}

\vspace{-7pt}
\end{wraptable}

Doubly robust evaluation, counterfactual risk
minimization, and orthogonal nuisance fitting provide
complementary tools for learning from logged data
\cite{dudik2011doubly,wang2017optimal,
swaminathan2015counterfactual,chernozhukov2018double}.
Exponential smoothing~\cite{aouali2023smoothing}
and marginal density-ratio estimation~\cite{taufiq2023marginal}
address related off-policy evaluation difficulties,
while risk--return benchmarking studies downstream
policy selection~\cite{kiyohara2024risk}.
Overlap-weighted temporal estimation~\cite{hess2026overlap}
and future-dependent evaluation under partial
observability~\cite{uehara2023future}
address distinct longitudinal challenges.
These methods do not by themselves align exploratory
contexts with the no-exploration steady-state law.
Action correction and context transport therefore
remain separate identification requirements here.

\noindent\textbf{\textcircled{\scriptsize 3}\ Sequential choice and information cost.} Optimism-based bandits use uncertainty to guide
sequential decisions
\cite{auer2002finite,abbasi2011improved,lattimore2020bandit}.
Contextual bandit methods examine cumulative versus
simple regret~\cite{krishnamurthy2023proportional},
heterogeneous multi-agent learning~\cite{do2023heterogeneous},
and structured cross-learning~\cite{schneider2023cross}.
Network-interference bandits and design-based approaches
study interacting rewards and inference--regret trade-offs
\cite{agarwal2024mabn,jia2025mabi,wang2025design,zhang2025online}.
Variance-dependent and second-order regret analyses
provide complementary contextual-bandit guarantees
\cite{jia2024variance,pacchiano2025second}.
These lines do not directly address the identification
of a global configuration's no-exploration steady-state
value from exploration-perturbed telemetry.
Proposition~\ref{prop:v8-nonid} formalizes this identification requirement.

\section{Operational Setup and Causal Identification}
\label{sec:system-model}

We formalize deployment logs and an interference-aware action--exposure representation, then define the no-exploration steady-state value as the target of configuration selection.
This formulation distinguishes exploratory measurements from deployment objectives and makes explicit the identification requirements underlying subsequent causal evaluation and certification.

\subsection{Deployment Logs and the Target Object}
We consider a networked system with $N$ nodes and a compact configuration set $\mathcal Z$.
At slow round $t$, the operator deploys a global configuration $z_t\in\mathcal Z$ and a measurement-level exploration rate $\eta_t\in[0,1]$.
After $T_{\mathrm{burn}}$ fast adaptation steps, the operator records $T_0$ post-burn-in windows.
A logged window contains the realized low-level action $A_{i,s}^{(t)}$, context $C_{i,s}^{(t)}$, and normalized utility $R_{i,s}^{(t)}$ for node $i$ and window $s$, with $|R_{i,s}^{(t)}|\le B_u$ after KPI normalization.
We write $O_{i,s}^{(t)}=(A_{i,s}^{(t)},C_{i,s}^{(t)},R_{i,s}^{(t)})$ for each logged observation, $\mathcal D_t=\{O_{i,s}^{(t)}:i=1,\ldots,N,\ s=1,\ldots,T_0\}$ for the batch collected in round $t$, and $\mathcal D_{\le t}=\bigcup_{r=1}^t\mathcal D_r$ for the cumulative log.

\subsection{Exposure Interface for Network Interference}
A direct formulation over the full node-level action profile $A_{1:N,s}^{(t)}$ is statistically and operationally impractical because local controllers are black-box and mutually interfering.
We instead use an auditable exposure interface:
\begin{equation}
E_{i,s}^{(t)}
=
\phi_i\!\left(
\{A_{j,s}^{(t)}\}_{j\in\mathcal N_i(s)},
S_s^{(t)},G_s^{(t)}
\right),
\qquad
\widetilde A_{i,s}^{(t)}
=
\bigl(A_{i,s}^{(t)},E_{i,s}^{(t)}\bigr).
\label{eq:exposure}
\end{equation}
Here $\widetilde A_{i,s}^{(t)}$ is the joint action/exposure treatment unit.
The pre-action context $C_{i,s}^{-,(t)}$ contains only information available before the focal action is generated; contemporaneous telemetry used by $\phi_i$ belongs to the observed exposure.
We score $\widetilde O_{i,s}^{(t)}=\bigl(\widetilde A_{i,s}^{(t)},C_{i,s}^{-,(t)},R_{i,s}^{(t)}\bigr)$, while retaining the raw telemetry in $\mathcal D_t$.
The map $\phi_i$ may summarize neighbor load, interference power, top-$k$ congestion, or graph features \citep{hudgens2008toward,aronow2017estimating,savje2021average}; $S_s^{(t)}$ and $G_s^{(t)}$ denote system telemetry and graph state.
Exposure summarizes realized neighboring actions, not an assigned treatment.

Let $\widetilde{\mathcal A}$ be the joint action/exposure space.
For candidate $z$, $g_{z,0}(\widetilde a\mid c^-)$ is the target law and $g^{\mathrm{obs}}_{z,\eta}(\widetilde a\mid c^-)$ is the law induced by the actual logging mechanism; a round-$r$ observation uses its source-regime law $(z_r,\eta_r)$, fitted as $\widehat g_{b,r}$.
Because exposure depends on multiple contemporaneous actions, the joint law need not inherit the marginal focal-action mixture.
The DR score therefore requires the actual joint action/exposure probability or density, not merely the marginal propensity of $A_i$.

Identification requires consistency for this treatment/exposure unit, conditional exchangeability and overlap given $C^-$, absolute continuity of the target joint law with respect to the source behavior law, and calibrated response transport across source regimes.
These assumptions are not implied by relabeling context.
These identification conditions follow standard
causal principles~\cite{pearl2009causality}.
The calibrated budget $\epsilon_{\exp}(z)$
accounts for exposure approximation and
response-transport discrepancies (Section~\ref{sec:certificate-main}).
We audit exposure sufficiency by comparing $\phi_i$ with an enriched map $\phi_i^+$, using the discrepancy $\widehat\Delta_{\mathrm{exp},t}(z)=\bigl|\widehat W_t^{\phi^+}(z;0)-\widehat W_t^\phi(z;0)\bigr|$.
A large discrepancy motivates enrichment or a larger
calibrated budget; the comparison is a sensitivity
diagnostic, not an independently valid certificate.

\subsection{Identification-Only Exploration and the No-Exploration Benchmark}
The operator ultimately wants the value of a configuration without measurement perturbations.
Let $\nu_{z,\eta}$ be the post-burn-in distribution of pre-action contexts $C^-$ induced by configuration $z$ and exploration rate $\eta$.
Let $m_z(\widetilde a,c^-)$ denote the no-exploration target-regime conditional response for the joint action/exposure unit.
The target value is
\begin{equation}
W_\infty(z;0)
=
\mathbb E_{C^-\sim\nu_{z,0}}
\left[
\int_{\widetilde{\mathcal A}}
m_z(\widetilde a,C^-)\,
g_{z,0}(d\widetilde a\mid C^-)
\right],
\qquad
z^\star\in\arg\max_{z\in\mathcal Z}W_\infty(z;0).
\label{eq:target-value}
\end{equation}
For discrete $\widetilde{\mathcal A}$, the integral is the corresponding sum.
The deployment-facing regret is
\begin{equation}
\mathrm{Reg}_T
=
\sum_{t=1}^{T}
\left[
W_\infty(z^\star;0)-W_\infty(z_t;0)
\right].
\label{eq:target-regret}
\end{equation}

The local-action logging mechanism retains its mixture form $\pi^{\mathrm{obs}}_{z,\eta}(a\mid c^-)=(1-\eta)\pi_{z,0}(a\mid c^-)+\eta\xi(a\mid c^-)$, with $\xi(a\mid c^-)>0$.
Its induced joint law is $g^{\mathrm{obs}}_{z,\eta}$, and each historical score uses the behavior law for its actual source regime $(z_r,\eta_r)$.
This marginal mixture does not determine the joint exposure propensity, and action correction does not transport the exploratory context law to $\nu_{z,0}$; see Appendix~\ref{app:target-logging}.
Define the steady-state value under exploration by
\begin{equation}
W_\infty(z;\eta)
=
\mathbb E_{C^-\sim\nu_{z,\eta}}
\left[
\int_{\widetilde{\mathcal A}}
m_z(\widetilde a,C^-)\,
g^{\mathrm{obs}}_{z,\eta}(d\widetilde a\mid C^-)
\right].
\label{eq:exploration-value}
\end{equation}
The source-regime response discrepancy remains part of the calibrated response-transport budget.

\begin{wrapfigure}{r}{0.38\columnwidth}
\vspace{-25pt}
\centering
\includegraphics[width=\linewidth]{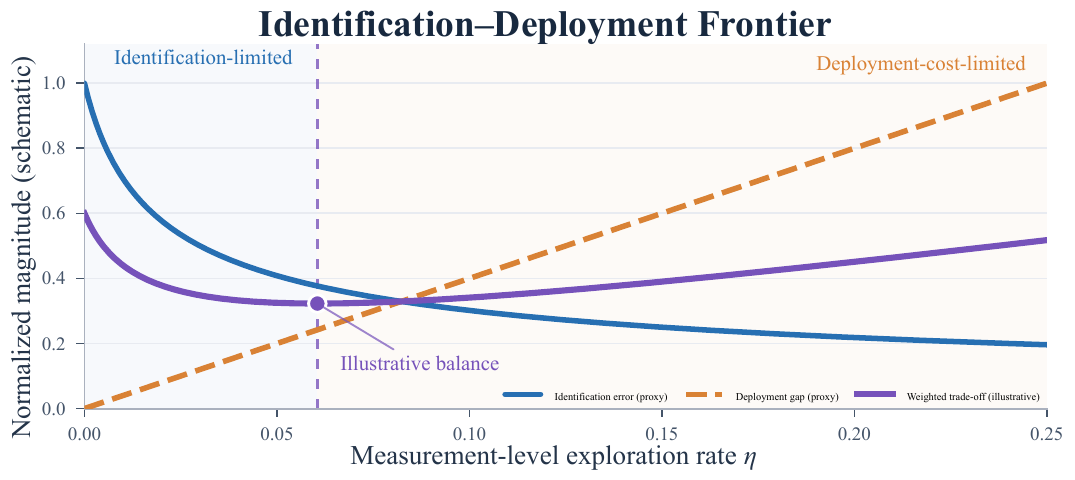}
\vspace{-8pt}
\captionsetup{font=footnotesize,justification=raggedright,singlelinecheck=false}
\caption{\textbf{Identification--deployment trade-off.}
{\scriptsize Blue: error; orange: cost; purple: combined; dashed: illustrative balance.}}
\label{fig:id-deployment}
\vspace{-20pt}
\end{wrapfigure}

\section{Causal Value Certificates and Online Optimization}
\label{sec:certificate-main}

We develop an exposure-aware clipped doubly robust
estimator for the no-exploration steady-state value,
together with calibrated deviation certificates.
We then characterize the identification--deployment
trade-off, analyze optimistic online selection under
explicit information-growth conditions, and give
an anytime certificate for a restricted
known-local-interference model.

\subsection{Clipped Doubly Robust Evaluation}
Evaluation uses the candidate target joint law $\widehat g_{z,0}$, response regression $\widehat m_z$, and source-round behavior joint propensity $\widehat g_{b,r}$.
The target law supplies the direct term and importance numerator; the behavior law supplies the denominator.
These nuisance fits are frozen before the corresponding network window is scored, and stored historical scores are never recomputed using later fits.
For candidate $z$, let
\[
\begin{aligned}
m_z(\widetilde a,c^-)
&=\mathbb E[R\mid \widetilde A=\widetilde a,
C^-=c^-,z,\eta=0],\\
\mu_z(c^-)
&=\int_{\widetilde{\mathcal A}}
m_z(\widetilde a,c^-)\,
g_{z,0}(d\widetilde a\mid c^-).
\end{aligned}
\]
For an observation from round $r$, define the clipped joint importance weight
\begin{equation}
\widehat w_{z,r}(\widetilde a,c^-)
=
\min\left\{
\frac{\widehat g_{z,0}(\widetilde a\mid c^-)}
{\max\{\widehat g_{b,r}(\widetilde a\mid c^-),\pi_{\min}\}},
w_{\max}
\right\}.
\label{eq:clipped-joint-weight}
\end{equation}
The ratio is a probability ratio for discrete treatments and a density ratio (under a common dominating measure) for continuous treatments; $\pi_{\min}$ uses the same convention.
The score and pooled estimator are
\begin{align}
\widehat\psi_{z,r}(\widetilde O)
&=
\widehat\mu_z(C^-)
+
\widehat w_{z,r}(\widetilde A,C^-)
\{R-\widehat m_z(\widetilde A,C^-)\},
\label{eq:dr-score}\\
\widehat\mu_z(C^-)
&=
\int_{\widetilde{\mathcal A}}
\widehat m_z(\widetilde a,C^-)
\widehat g_{z,0}(d\widetilde a\mid C^-),
\notag\\
\widehat W_t(z;0)
&=
\frac{1}{NT_0t}
\sum_{\substack{r=1,\ldots,t\\s=1,\ldots,T_0\\i=1,\ldots,N}}
\widehat\psi_{z,r}\!\left(\widetilde O_{i,s}^{(r)}\right).
\label{eq:estimator}
\end{align}
The inverse-probability component is connected to
classical estimation under incomplete
observation~\cite{robins1994estimation}.
Here, the weighting must additionally account for the
joint action/exposure treatment law and the distinction
between exploratory and target steady-state regimes.

For discrete treatments, the direct-method integrals are sums. The pooled estimator uses $\mathcal D_{\le t}$ and is available after round $t$ for the decision in round $t+1$.
Cross-configuration response differences require the calibrated transport allowance below.
With $\widehat m_z$ clipped to $[-B_u,B_u]$ and weights clipped at $w_{\max}$, $|\widehat\psi_{z,r}(\widetilde O)|\le B_\psi=2B_u(1+w_{\max})$.
Appendices~\ref{app:mixture-ratio}--\ref{app:clipping-detailed} derive the joint-ratio bounds, DR identity, clipping bias, and nuisance-product expansion \citep{dudik2011doubly,swaminathan2015counterfactual,wang2017optimal}.

For each candidate $z$, nuisance fitting weights past rounds by context similarity rather than pooling them uniformly, using $\alpha_{r,z}\propto K_h\!\left(\widehat d_F(\widehat\nu_r,\widehat\nu_{z,0})\right)$, normalized over $r<t$.
The resulting training-context marginal $\widehat\nu^{z}_{\mathrm{train},t}$ defines the fits for $\widehat m_z$ and $\widehat g_{z,0}$.
Weighting addresses context mismatch but does not establish response invariance; a large $\bar\epsilon_{\mathrm{cfg},t}(z)$ triggers local refitting or a wider radius.

\subsection{Uniform Causal Value Certificate}
\label{subsec:certificate}

The certificate combines fluctuation with calibrated error allowances.
For monitoring, let $J_t=T_0t$ and define the network-window scores as $Y_{j,z}=N^{-1}\sum_{i=1}^{N}\widehat\psi_{z,r(j)}(\widetilde O_{i,j})$, where $j$ indexes the time-ordered network windows and $r(j)$ is the deployment round containing window $j$.
An autocorrelation-based effective sample size is estimated from this aggregated sequence as
\[
\widehat n_{\mathrm{eff},t}(z)
=
\frac{J_t}
{1+2\sum_{\ell=1}^{L_t}\max\{\widehat\rho_{\ell,t}(z),0\}}.
\]
The plug-in ESS is diagnostic; $\underline n_{\mathrm{eff},t}(z)$ must be a calibrated lower-information bound for the aggregated network-window sequence, without assuming within-window node independence.
Concentration also requires bounded conditional drift
and validated dependence control, such as mixing or
block coupling, under the calibration conditions
specified in Appendix~\ref{app:ss-blocks} and~\ref{app:concentration-detailed}.
The adjacent-block diagnostic $\widehat\kappa_t(z)$ enters the radius only when calibrated as an upper bound on residual drift.

The calibrated context-shift budget $\bar\epsilon_{\mathrm{shift},t}(z)$ controls score-averaging transport from the pooled source contexts to $\nu_{z,0}$; $\bar\epsilon_{\mathrm{cfg},t}(z)$ controls transfer of bounded context-level nuisance risk from $\widehat\nu^{z}_{\mathrm{train},t}$ to $\nu_{z,0}$.
Joint-ratio and response-nuisance errors require separate calibration under the actual source-window score laws: context-marginal similarity alone bounds neither joint input-law error nor the no-exploration context law.

The budget $\epsilon_{\mathrm{exp}}(z)$ covers exposure-map approximation and cross-configuration/measurement-regime response transport.
For the proof, it must dominate the historical average of both target-context response discrepancy and score-level response residual, uniformly over source regimes and prefixes (Appendices~\ref{app:exposure-majorant}, \ref{app:thm1proof-detailed}).
A narrower exposure-only budget must therefore be expanded and recalibrated.
Together with target-joint-law estimation error, nuisance-product error, clipping bias, context transport, and configuration mismatch, the budgets form
\begin{equation}
\begin{aligned}
\bar\epsilon_{\mathrm{tot},t}(z)
={}&
\epsilon_{m,t}(z)\epsilon_{\pi,t}(z)
+\bar\epsilon_{\mathrm{target},t}(z)
+\epsilon_{\mathrm{clip},t}(z)\\
&+\epsilon_{\mathrm{exp}}(z)
+\bar\epsilon_{\mathrm{shift},t}(z)
+\bar\epsilon_{\mathrm{cfg},t}(z).
\end{aligned}
\label{eq:total-error-budget}
\end{equation}
Here $\epsilon_{m,t}(z)$ is the response-regression error under source-window score laws, while $\epsilon_{\pi,t}(z)$ is the induced clipped joint-ratio error $\|\widehat w_{z,r}-\bar w_{z,r}\|_{L_2(\nu_{z_r,\eta_r},g_{b,r})}$, not raw propensity-fit error.
The independently calibrated $\bar\epsilon_{\mathrm{target},t}(z)$ bounds first-order error in $\widehat g_{z,0}$; calibration details appear in Appendices~\ref{app:ss-blocks}, \ref{app:context-shift-detailed}, and \ref{app:diagnostic-algorithms}.

\begin{theorem}[Unified deviation certificate]
\label{thm:main}
Fix $t$.
Assume bounded, predictable/cross-fitted scores; overlap for the joint action/exposure law; a valid exposure/response, shift, and mismatch error budget; and conditional dependence and drift bounds that certify $\underline n_{\mathrm{eff},t}(z)$ and $\widehat\kappa_t(z)$ (Appendices~\ref{app:ss-blocks} and \ref{app:concentration-detailed}).
With probability at least $1-\delta$, simultaneously for all $z\in\mathcal Z$,
\begin{equation}
\left|
\widehat W_t(z;0)-W_\infty(z;0)
\right|
\le
2B_\psi
\sqrt{
\frac{\operatorname{comp}(\mathcal Z)+\log(2/\delta)}
{\underline n_{\mathrm{eff},t}(z)}
}
+\bar\epsilon_{\mathrm{tot},t}(z)
+\widehat\kappa_t(z).
\label{eq:main-bound}
\end{equation}
\end{theorem}

Theorem~\ref{thm:main} separates sampling error from calibrated nuisance, transport, clipping, and dependence contributions.
Diagnostics guide refitting and radius inflation but do not establish coverage; see Appendices~\ref{app:dr-identity-detailed}--\ref{app:v7-certification} and~\ref{app:certificate-scope-clarification}.

\subsection{The Identification--Deployment Frontier}
Exploration affects both the information available for causal evaluation and the utility experienced during deployment.
The following lower bound quantifies this identification--deployment tension in a slow-mixing steady-state family.

\begin{theorem}[Steady-state identification--deployment frontier]
\label{thm:frontier}
For each finite horizon $T$ and predetermined measurement-exploration schedule $\{\eta_t\}_{t=1}^T$, there exists a two-configuration, two-action stationary steady-state family, with potentially horizon-dependent mixing time, such that for any adaptive configuration-selection rule, the minimax expected absolute error of estimating $W_\infty(z_1;0)$ and the measurement-induced deployment gap satisfy, for universal $c_1,c_2>0$,
\begin{equation}
\mathrm{InferenceError}_T
\ge
\frac{c_1}{\sqrt{1+\sum_{t=1}^T\eta_t}},
\qquad
\mathrm{DeploymentGap}_T
\ge
c_2\sum_{t=1}^T\eta_t.
\label{eq:frontier}
\end{equation}
\end{theorem}

Measurement actions reveal latent state through refresh events while also perturbing deployment.
The adaptive-data KL and two-point proof is in Appendix~\ref{app:frontier-detailed}; the lower bound characterizes the joint information--deployment problem, not Pareto optimality of CAUSAL-UCB.

\subsection{CAUSAL-UCB: Certified Online Optimization}
CAUSAL-UCB selects the largest certified upper value after updating nuisance models and diagnostics; the selection schematic is in Appendix~\ref{app:conceptual-schematics}.

\begin{theorem}[Regret of CAUSAL-UCB]
\label{thm:regret}
Assume the confidence event in Theorem~\ref{thm:main} holds uniformly over rounds (with roundwise failure-probability allocation), and the selected effective sample sizes satisfy square-root summability.
Define $e_t(z)=\bar\epsilon_{\mathrm{tot},t}(z)+\widehat\kappa_t(z)$.
Then
\begin{equation}
\operatorname{Reg}_T
=
\widetilde O\!\left(
B_\psi\sqrt{T\operatorname{comp}(\mathcal Z)}
+\sum_{t=2}^T e_{t-1}(z_t)
\right).
\label{eq:regret}
\end{equation}
If $e_{t-1}(z_t)\le ct^{-\gamma}$ for $0<\gamma<1$, the cumulative error is $O(T^{1-\gamma})$.
First-round regret is at most $2B_u$.
If $W_\infty(z;\eta)$ is uniformly $L_W$-Lipschitz in $\eta$ for every $z\in\mathcal Z$, then the cumulative deployment gap $\sum_{t=1}^T|W_\infty(z_t;\eta_t)-W_\infty(z_t;0)|$ is at most $L_W\sum_{t=1}^T\eta_t$.
\end{theorem}

For finite $\mathcal Z$, assume the per-visit condition $n_{\mathrm{eff},t-1}(z_t)\ge\lambda\{1+N_{z_t}(t-1)\}$.
This yields, up to initialization constants,
\[
\sum_{t\ \mathrm{eligible}}\frac{1}{\sqrt{\underline n_{\mathrm{eff},t-1}(z_t)}}
\le 2\sqrt{\frac{|\mathcal Z|T}{\lambda}},
\]
and hence statistical regret $\widetilde O(B_\psi\sqrt{|\mathcal Z|T\operatorname{comp}(\mathcal Z)/\lambda})$.
For fixed $|\mathcal Z|$ and $\lambda$, this gives the stated square-root rate; growing $|\mathcal Z|$ retains the factor $\sqrt{|\mathcal Z|/\lambda}$ unless stronger square-root summability is assumed.
The information-growth condition is substantive, not implied by optimism; certificate-error and deployment-gap contributions are already stated in Theorem~\ref{thm:regret}.
Proofs appear in Appendices~\ref{app:regret-detailed}, \ref{app:gap-detailed}, and~\ref{app:certificate-scope-clarification}.

\begin{wrapfigure}{r}{0.53\columnwidth}
\vspace{-3pt}
\raggedright
\footnotesize

\refstepcounter{algorithm}
\label{alg:causalucb}

\hrule height 0.7pt
\vspace{2pt}
\noindent
\textbf{Algorithm \thealgorithm}\quad
\textbf{CAUSAL-UCB}
\par\vspace{2pt}
\hrule height 0.45pt
\vspace{2pt}

\begin{algorithmic}[1]
\renewcommand{\algorithmicrequire}{\textbf{Input:}}

\Require $\mathcal Z,T,T_{\mathrm{burn}},T_0,
\{\eta_t\},w_{\max},\pi_{\min}$.

\Statex \textbf{Initialize:}
$\mathcal D_{\le0}=\emptyset$,
$\widehat W_0(z;0)=0$,
$b_0(z)=+\infty$.

\For{$t=1,\ldots,T$}
    \State Fit predictable/cross-fitted nuisance models on $\mathcal D_{\le t-1}$ with configuration-mismatch weighting.
    \State Compute $\widehat W_{t-1}(z;0)$ and $b_{t-1}(z)$ for all $z\in\mathcal Z$ using calibrated diagnostics.
    \State Select $z_t$ maximizing $\widehat W_{t-1}(z;0)+b_{t-1}(z)$ over $\mathcal Z$.
    \State Deploy $(z_t,\eta_t)$; wait $T_{\mathrm{burn}}$; log $T_0$ windows into $\mathcal D_t$ and update $\mathcal D_{\le t}$.
\EndFor

\State \textbf{Return:} Configurations, value estimates, and certificates.
\end{algorithmic}

\vspace{2pt}
\hrule height 0.7pt
\vspace{-5pt}
\end{wrapfigure}

With known local interference, fixed exploration, homogeneous matched rewards, and known noise scale, the general certificate specializes to a computable anytime radius.
In this restricted setting, known neighborhood actions determine whether a logged observation matches the no-exploration target profile.
Because the exploration mechanism is fixed and known, the matching probability can be computed directly.
We specify the corresponding model and certificate below.

Consider $K$ configurations and $N$ nodes.
With no-exploration action fixed at zero and exploration action $\operatorname{Bernoulli}(1/2)$, local actions are independent $\operatorname{Bernoulli}(\eta/2)$ for fixed, known $\eta$.

Let each known interference neighborhood contain $d=2r+1$ nodes.
Define
\[
I_{j,i}=\mathbf 1\{\text{all actions in node $i$'s neighborhood equal zero}\},
\]
with matching probability $q=(1-\eta/2)^d>0$.

Conditional on $I_{j,i}=1$, rewards follow $\operatorname{clip}(\mu_z+\sigma Z_{j,i},0,1)$, with iid Gaussian noise independent of actions and prior deployments and a common law across nodes/windows for each $z$.
Assume known noise-scale upper bound $\sigma$, $w_{\max}\ge1/q$, and no hidden interference, context transport, nonstationarity, or burn-in dependence.
The target profile is the all-zero neighborhood, logged with probability $q$, so the ratio is $1/q$ on matched windows and zero otherwise; at $\eta=0$ all nodes are matched, giving the target value $W_z=\mathbb E[\operatorname{clip}(\mu_z+\sigma Z,0,1)]\in[0,1]$.

For each configuration $z$, index its $n$ logged network windows by $j$.
Let $K_j=\sum_{i=1}^N I_{j,i}$, $h_j=K_j/(Nq)$, and define the \emph{prequential, unclipped, doubly robust} score
\begin{equation}
Y_{j,z}
=
\widehat m_{j-}(z)
+
\frac{1}{Nq}\sum_{i=1}^N
I_{j,i}\bigl(R_{j,i}-\widehat m_{j-}(z)\bigr),
\qquad
\widehat W_{z,n}
=
\frac1n\sum_{j=1}^nY_{j,z}.
\label{eq:restricted-dr}
\end{equation}

Here $\widehat m_{j-}(z)$ is the matched-reward mean available before the deployment round containing window $j$ (set to $1/2$ if none exists); all windows in that round use this frozen mean.
With $M=NTT_0$, define
\begin{align}
L&=\log\frac{2K(M+1)}{\delta},\\
\epsilon(m)&=
\begin{cases}
1/2,&m=0,\\
\min\{1,\sigma\sqrt{2L/m}\},&m\ge1.
\end{cases}
\end{align}
Let $m_{j-}$ be the matched count available before that deployment round, and $m_n=\sum_{j=1}^nK_j$.

\begin{proposition}[Anytime calibrated DR certificate, restricted model]
\label{prop:certified-special}
With probability at least $1-\delta$, simultaneously over every configuration $z\in\mathcal Z$ and every time at which its data are examined, the projected estimator $\widetilde W_{z,n}=\Pi_{[0,1]}\widehat W_{z,n}$ satisfies
\begin{equation}
\label{eq:special-case-certified-radius}
|\widetilde W_{z,n}-W_z|\le b_{z,n},
\qquad
b_{z,n}
=
\frac1n\sum_{j=1}^n|1-h_j|\epsilon(m_{j-})
+\frac{m_n}{nNq}\epsilon(m_n).
\end{equation}
For an as-yet unvisited configuration, use $\widetilde W_{z,0}=1/2$ and $b_{z,0}=1/2$.
The certificate can be computed solely from previously logged actions and outcomes.
In particular, $b_{z,n}$ may be used directly for online selection, and its corresponding interval may be reported as the experiment's confidence interval.
It is not a post-hoc interval.
\end{proposition}

For online selection, let $n_z(t-1)$ count windows under $z$ before round $t$.

\begin{corollary}[Deployment regret conditional on the same certificate]
\label{cor:restricted-regret}
Choose
\[
z_t\in\arg\max_{z\in\mathcal Z}
\{\widetilde W_{z,n_z(t-1)}+b_{z,n_z(t-1)}\},
\]
with optimism capped at $1$ if desired.
After at most $H$ forced initialization rounds, on the simultaneous event of the proposition,
\begin{equation}
\operatorname{Reg}_T
\le
H+2\sum_{t=H+1}^T b_{z_t,n_{z_t}(t-1)}.
\label{eq:restricted-regret}
\end{equation}
\end{corollary}

Appendix~\ref{app:restricted-case-proofs} proves the proposition and corollary using one simultaneous matched-reward event over configurations and examined prefixes, including adaptive decisions.
This is the known-mechanism special case of Theorem~\ref{thm:main}; the general setting instead requires separately calibrated target-law, transport, exposure, and dependence budgets.

\paragraph{Decision implementation.}
The experiment initializes each of the five configurations twice ($H=10$); thereafter it selects the configuration maximizing
\[
\min\{1,\widetilde W_{z,n_z(t-1)}+b_{z,n_z(t-1)}\}
\]
and resolves exact upper-bound ties by the current center estimate.
The same $b$ governs selection, reported confidence width, and the corollary's per-round bound; oracle target values are evaluation-only, while decisions use prior matched observations and predecision radii.

\section{Experiments}
\label{sec:new-experiments}

\begin{wrapfigure}{r}{0.43\columnwidth}
    \vspace{-32pt}
    \centering

    \begin{subfigure}[t]{0.48\linewidth}
        \centering
        \includegraphics[width=\linewidth]{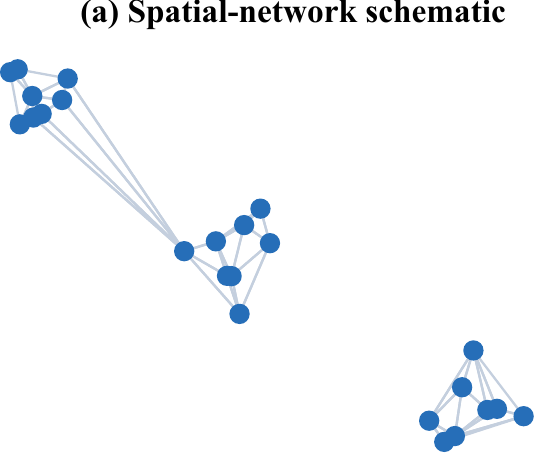}
        \label{fig:main-a}
    \end{subfigure}%
    \hfill
    \begin{subfigure}[t]{0.48\linewidth}
        \centering
        \includegraphics[width=\linewidth]{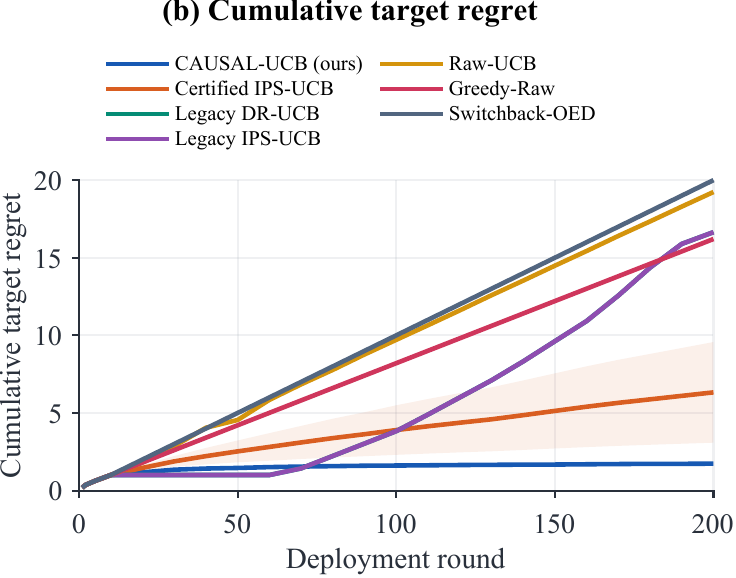}
        \label{fig:main-b}
    \end{subfigure}

    \par\vspace{2pt}

    \begin{subfigure}[t]{0.48\linewidth}
        \centering
        \includegraphics[width=\linewidth]{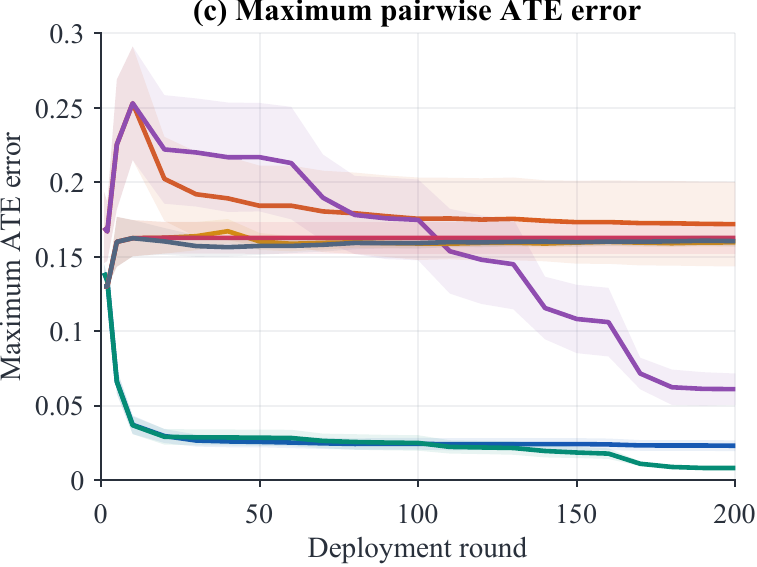}
        \label{fig:main-c}
    \end{subfigure}%
    \hfill
    \begin{subfigure}[t]{0.48\linewidth}
        \centering
        \includegraphics[width=\linewidth]{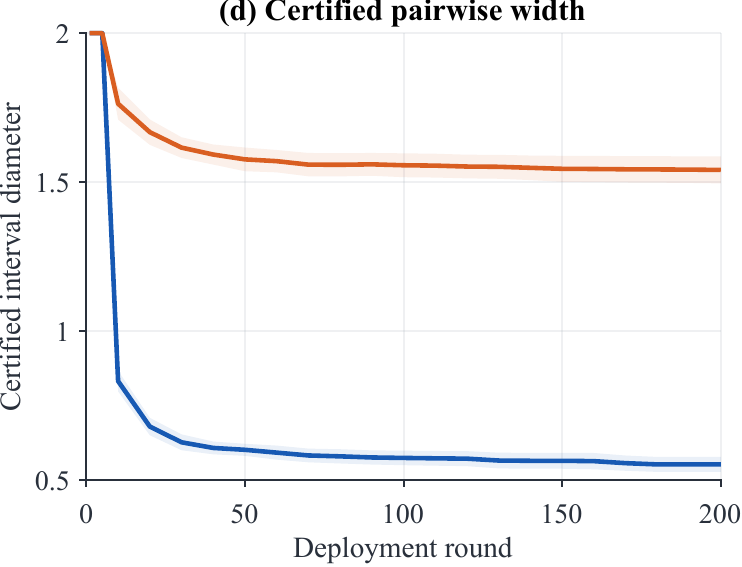}
        \label{fig:main-d}
    \end{subfigure}

    \captionsetup{font=footnotesize,skip=2pt}
    \caption{\textbf{Certified online selection}
    (a) Conceptual spatial-network schematic;
    (b) cumulative target regret;
    (c) maximum pairwise ATE error;
    (d) decision-active certificate diameter
    for Certified DR/IPS.
    Curves show 24-paired-seed means with
    pointwise 95\% Student-$t$ intervals.}
    \label{fig:main-certified}
    \vspace{-20pt}
\end{wrapfigure}

We evaluate CAUSAL-UCB through four theory-driven questions: \textbf{\textcircled{\scriptsize 1} Certified decisions:} Does target-value certification improve online selection under logging--target mismatch?
\textbf{\textcircled{\scriptsize 2} Interference and identification:} How do exposure, joint overlap, and context shift affect causal evaluation?
\textbf{\textcircled{\scriptsize 3} Mechanisms and cost:} Which components matter, and how does measurement exploration trade information against deployment utility?
\textbf{\textcircled{\scriptsize 4} Trace-driven transfer:} Do the findings persist on network-derived traces?

\noindent\textbf{Settings and scope.} The controlled benchmark uses a 24-node ring, five configurations, 200 rounds, and 24 paired seeds.
METR-LA and FlockLab provide topology and context traces for semi-synthetic replays with model-generated counterfactual outcomes (180 rounds, eight independent seeds per condition); these are not live deployments.
A separate 300-seed independent audit assesses full-path certificate coverage (Appendix~\ref{app:seven-term-fullpath-300}).
We report target regret, final optimality gap, pairwise causal-estimation error, uncertainty width, and deployment gap, distinguishing decision-active certificates from diagnostic widths.
Details and robustness checks are in the appendices; empirical performance does not establish the identification assumptions.

\subsection{Certified Decision Quality under Mismatch}
\label{subsec:results-main}

\noindent\textbf{Protocol.} The frozen $N=24$ ring benchmark uses five configurations, $T=200$ rounds, $T_0=2$ retained windows, radius-one interference, exploration rates $\eta\in\{0.55,0.08\}$, and 24 paired seeds.
Oracle target values are evaluation-only.

\begin{wraptable}[10]{r}{0.48\columnwidth}
\vspace{-2.7\baselineskip}
\centering
\setlength{\abovecaptionskip}{1pt}
\setlength{\belowcaptionskip}{3pt}
\caption{\textbf{High-mismatch controlled comparison.}
Means over 24 paired seeds and 200 rounds. Low-mismatch results are in Appendix~\ref{app:metric-pareto}.}
\label{tab:certified-main}

{%
\fontsize{7.1}{8.0}\selectfont
\setlength{\tabcolsep}{1.35pt}
\renewcommand{\arraystretch}{0.98}
\begin{tabularx}{\linewidth}{@{}>{\raggedright\arraybackslash}p{0.35\linewidth}*{5}{>{\raggedleft\arraybackslash}X}@{}}
\toprule
Method & Reg.$\downarrow$ & Gap$\downarrow$ & ATE$\downarrow$ & $W\downarrow$ & Dep.$\downarrow$ \\
\midrule
\rowcolor{CausalBlue}
\textbf{CAUSAL-UCB (ours)} & \textbf{1.732} & \textbf{.000} & .0230 & \textbf{.552} & 30.607\\
Certified IPS-UCB & 6.326 & .022 & .1718 & 1.541 & 23.217\\
Legacy DR-UCB & 16.647 & .060 & \textbf{.0081} & 1.961 & 12.279\\
Legacy IPS-UCB & 16.647 & .060 & .0610 & 1.961 & 12.279\\
Raw-UCB & 19.235 & .085 & .1595 & 1.919 & 8.016\\
Greedy-Raw & 16.207 & .080 & .1626 & 2.000 & \textbf{4.120}\\
Switchback-OED & 20.000 & .110 & .1604 & 1.800 & 9.215\\
\bottomrule
\end{tabularx}
}
\par\vspace{2pt}
{\raggedright\fontsize{6.0}{7.0}\selectfont
ATE: maximum pairwise ATE error; Dep.: expected simulator deployment gap.
$W$ is decision-active only for certified methods; all other widths are diagnostics.\par}
\vspace{-3\baselineskip}
\end{wraptable}

With clipped-Gaussian noise ($\sigma=0.08$), we compare
certified/legacy DR/IPS, Raw-UCB, Greedy-Raw, and
Switchback-OED~\cite{bojinov2023switchback}
on five metrics.
Only certified widths are decision-active; each of five
sensitivity levels uses 24 paired seeds and 200 rounds
(Appendix~\ref{app:metric-pareto}).

\noindent\textbf{Results.} Under high mismatch, CAUSAL-UCB attains the lowest regret ($1.732$), zero terminal gap, and narrowest decision-active width ($0.552$), versus $6.326$ regret and $1.541$ width for Certified IPS-UCB.
Legacy DR-UCB attains the lowest ATE error
(0.0081 versus 0.0230 for CAUSAL-UCB),
whereas CAUSAL-UCB achieves substantially
lower target regret (1.732 versus 16.647).
These results distinguish causal estimation
accuracy from online configuration-selection
performance. CAUSAL-UCB's deployment gap ($30.607$) exceeds Certified IPS-UCB ($23.217$) and Greedy-Raw ($4.120$); under low mismatch, Greedy-Raw slightly lowers regret ($1.000$ vs. $1.018$).
The separate Pareto analysis is in Appendix~\ref{app:metric-pareto}.

\subsection{Interference, Overlap, and Context Transport}
\label{subsec:results-overlap}

\begin{wrapfigure}[11]{r}{0.43\columnwidth}
    \vspace{-15pt}
    \centering
    \includegraphics[width=0.49\linewidth]{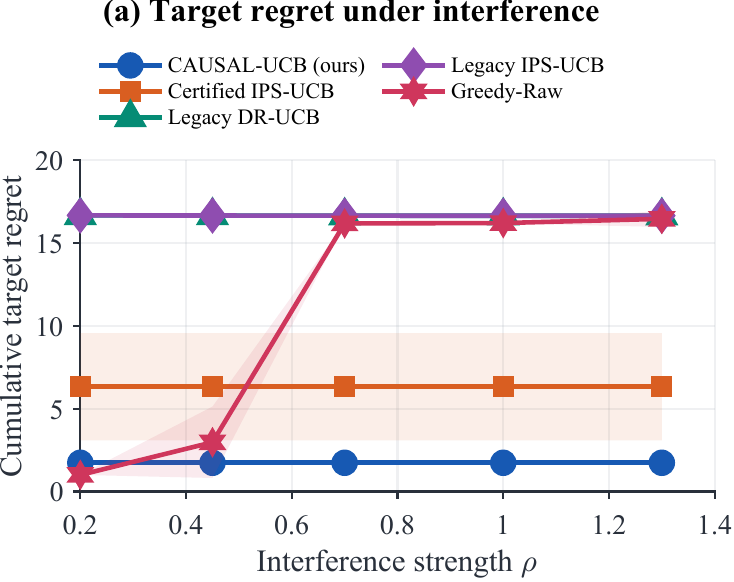}%
    \hfill
    \includegraphics[width=0.49\linewidth]{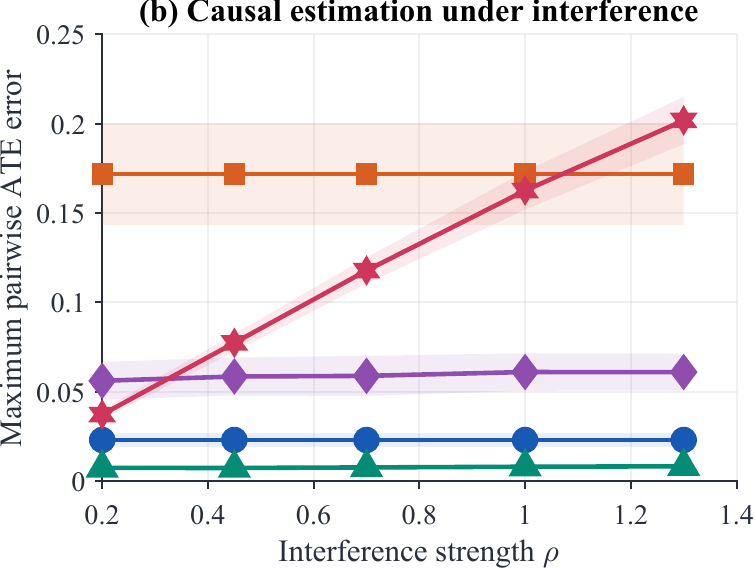}
    \captionsetup{font=footnotesize,skip=2pt}
    \caption{\textbf{Interference sensitivity.}
    (a) Cumulative target regret and (b) maximum pairwise ATE error
    across five interference levels. Curves show means over
    24 paired seeds with pointwise 95\% Student-$t$ intervals.}
    \label{fig:interference}
    \vspace{-5pt}
\end{wrapfigure}
We vary network interference across five levels with 24 paired seeds per condition (Figure~\ref{fig:interference}).
Panels (a) and (b) report cumulative target regret and maximum pairwise ATE error, respectively, distinguishing decision quality from causal-estimation accuracy.
Since neighboring actions affect realized exposure, focal-action propensity alone cannot characterize the joint action/exposure law.
These controlled results assess sensitivity rather than establish identification under arbitrary interference.
Oracle values are evaluation-only.
We next examine joint-action support and its distinction from post-burn-in context transport.

\begin{wrapfigure}{r}{0.46\columnwidth}
    \vspace{-30pt}
    \centering

    \includegraphics[width=0.32\linewidth]{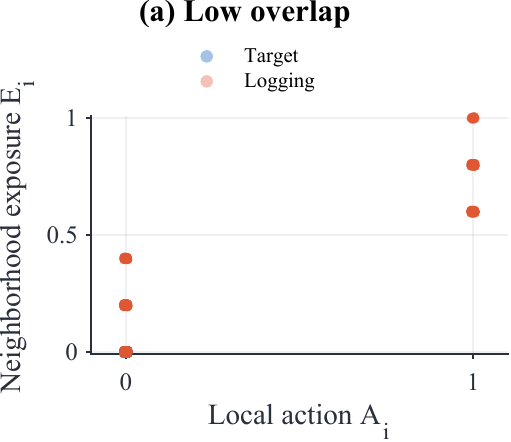}%
    \hfill
    \includegraphics[width=0.32\linewidth]{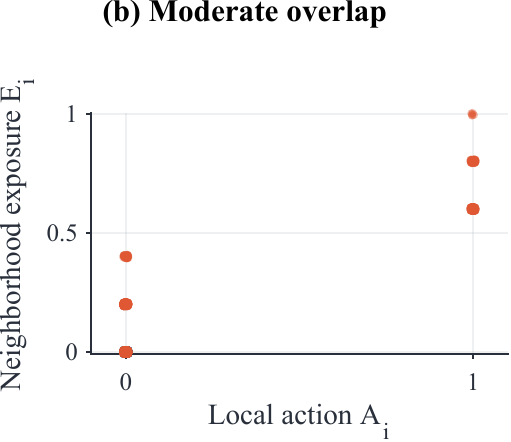}%
    \hfill
    \includegraphics[width=0.32\linewidth]{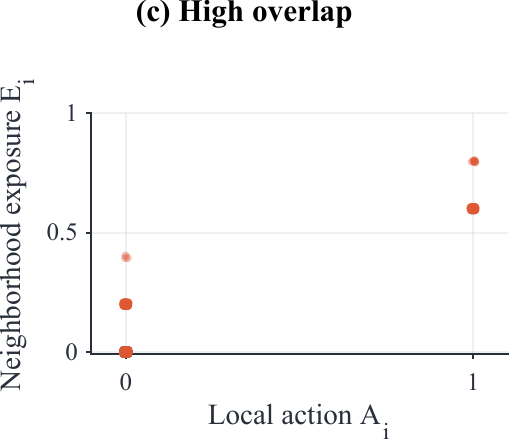}

    \par\vspace{2pt}

    \includegraphics[width=0.32\linewidth]{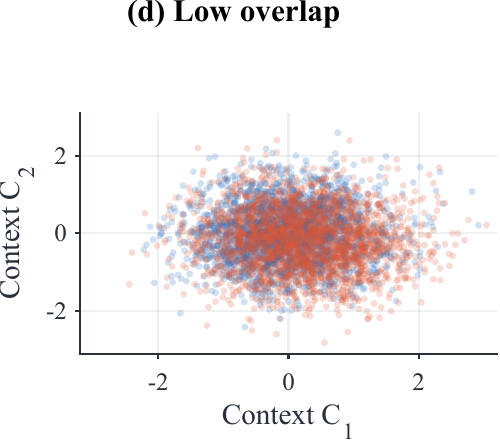}%
    \hfill
    \includegraphics[width=0.32\linewidth]{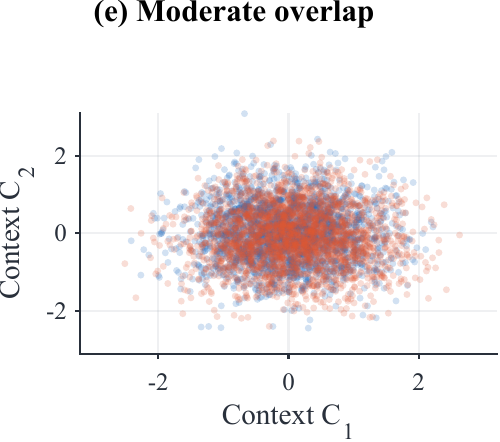}%
    \hfill
    \includegraphics[width=0.32\linewidth]{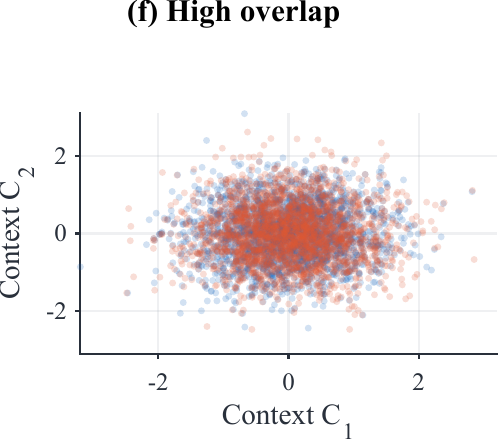}

    \par\vspace{2pt}

    \includegraphics[width=0.32\linewidth]{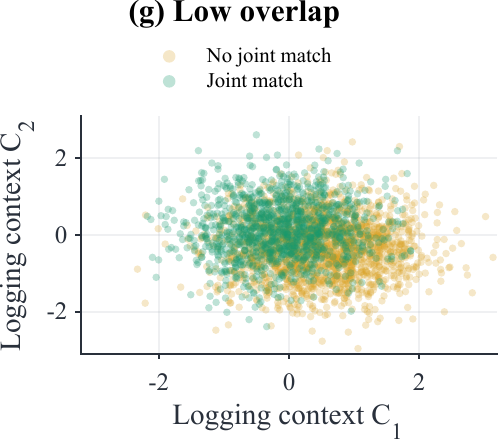}%
    \hfill
    \includegraphics[width=0.32\linewidth]{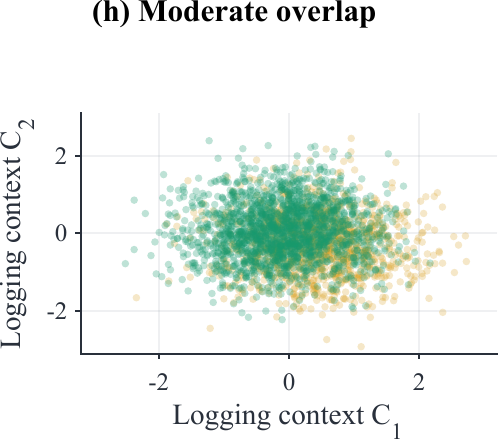}%
    \hfill
    \includegraphics[width=0.32\linewidth]{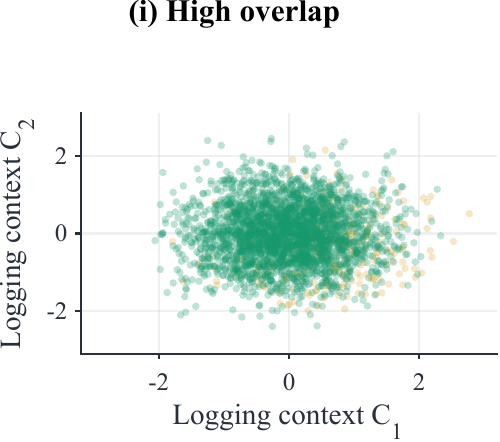}

    \par\vspace{2pt}

    \captionsetup{font=footnotesize,skip=2pt}
    \caption{\textbf{Logging--target support and context shift.}
    (a--c) Action--exposure support;
    (d--f) logging--target context distributions;
    (g--i) joint clean-neighborhood matches at
    $\eta=0.55,0.30,0.08$.
    Blue/red indicate target/logging draws in (a--f);
    green/yellow indicate match/no-match observations in (g--i).}
    \label{fig:joint-support-geometry}
    \vspace{-20pt}
\end{wrapfigure}
Conditional action overlap does not determine the no-exploration context law.
Figure~\ref{fig:joint-support-geometry} illustrates context-dependent logging propensities and exploration-induced post-burn-in responses.
The paired no-exploration target draws are used for visualization and evaluation only; the policies learn from logged observations.
At $\eta=0.55,0.30,0.08$, the joint clean-neighborhood probabilities are $0.402,0.623,0.891$, respectively.
The corresponding 10th-percentile match probabilities are $0.250,0.511,0.848$, and descriptive node-level IPW effective counts are $4{,}228,7{,}036,10{,}255$.

These diagnostics indicate that action/exposure matching improves as measurement exploration decreases in this controlled setting.
However, better joint support does not imply that the exploratory post-burn-in context distribution matches the no-exploration target distribution.
Thus, action-law correction and context transport remain separate requirements.
The plotted comparisons are sensitivity diagnostics, not independent guarantees of identification or certificate coverage.
Additional generator details and support analyses are provided in Appendix~\ref{app:joint-support-geometry}.

\subsection{Mechanisms and the Identification--Deployment Trade-off}
\label{subsec:results-mechanisms}

Five controlled variants (24 paired seeds, 200 rounds) assess DR correction, exposure, optimism, and exploration.

\begin{wrapfigure}[10]{r}{0.43\textwidth}
    \vspace{-20pt}
    \centering
    \includegraphics[width=0.49\linewidth]{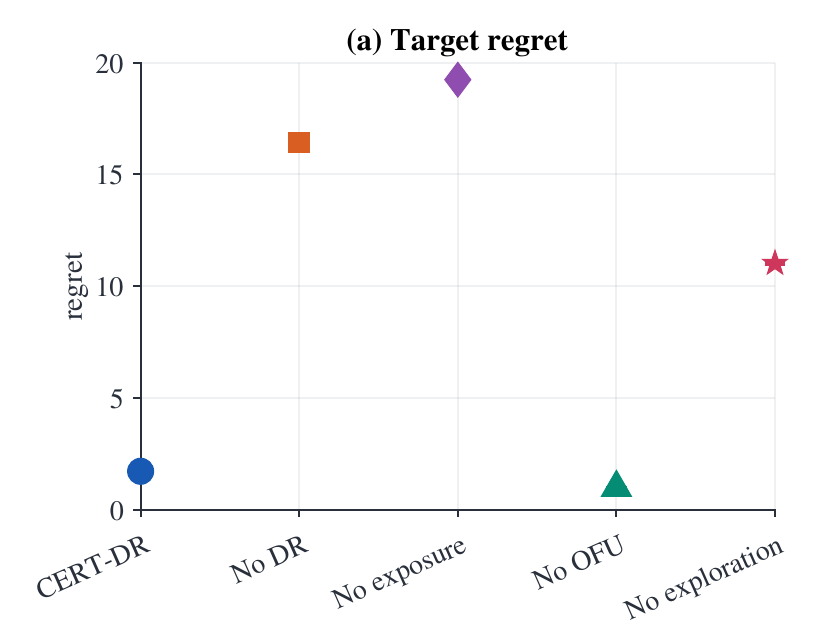}%
    \hfill
    \includegraphics[width=0.49\linewidth]{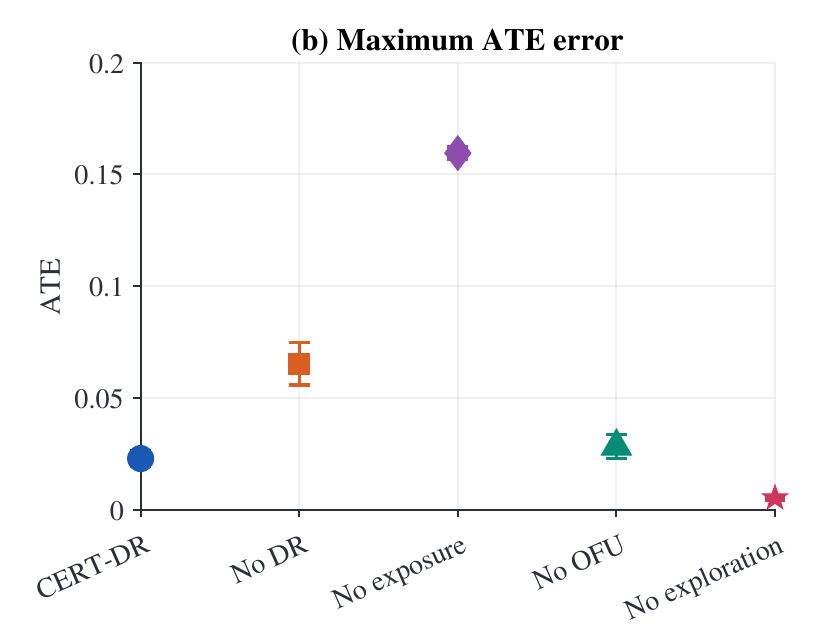}
    \captionsetup{font=footnotesize,skip=2pt,justification=centering}
    \caption{\textbf{Mechanism ablations.}
    (a) Cumulative target regret and
    (b) maximum pairwise ATE error across controlled
    variants. Results illustrate the distinction between
    sequential decision quality and causal-estimation accuracy.}
    \label{fig:ablation-regret}
    \vspace{-40pt}
\end{wrapfigure}

Removing DR correction or exposure raises regret; omitting exposure yields the largest ATE error (about $0.16$).
No exploration lowers ATE error (about $0.005$) but raises regret.
The no-OFU ablation changes more than optimism, so it cannot isolate the OFU effect.

The paired optimism study (Appendix~\ref{app:certified-optimism-initial-info}) compares Certified-OFU, Certified-Greedy, and Greedy+Decaying-Forced across five initial-information regimes (96 paired seeds each) with shared fits and audits.
With an under-sampled optimum, OFU lowers regret from $0.467$ to $0.241$; across 33 initially misranked seeds, regret is $0.232$ versus $1.077$ (paired difference $0.845$, 95\% CI $[0.621,1.068]$).
Certified-Greedy wins in the other four regimes.

\begin{wrapfigure}[9]{r}{0.43\textwidth}
    \vspace{-25pt}
    \centering
    \includegraphics[width=0.49\linewidth]{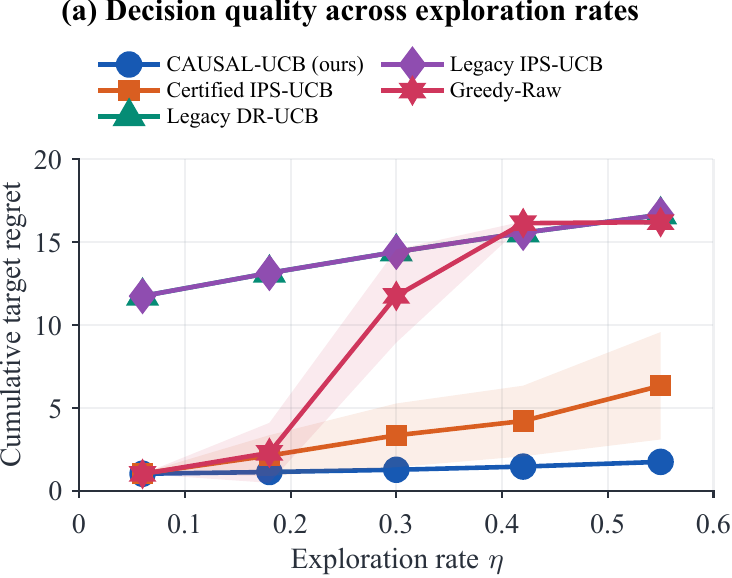}%
    \hfill
    \includegraphics[width=0.49\linewidth]{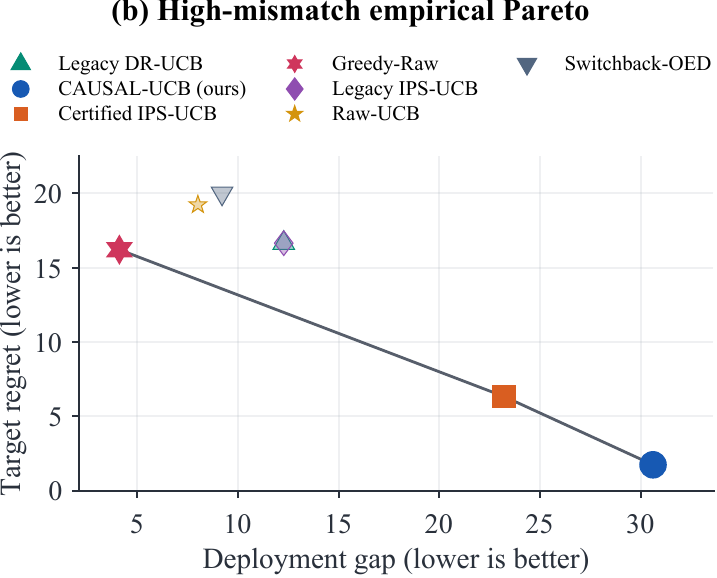}
    \captionsetup{font=footnotesize,skip=2pt,justification=centering}
    \caption{\textbf{Exploration and deployment trade-off.}
    (a) Cumulative target regret versus exploration rate
    $\eta$ and (b) high-mismatch Pareto frontier comparing
    cumulative target regret with deployment gap.}
    \label{fig:exploration-regret}
    \vspace{-5pt}
\end{wrapfigure}

We next sweep five fixed exploration rates.
Regret increases for several baselines but remains relatively stable for Certified DR (Figure~\ref{fig:exploration-regret}(a)); this illustrates, rather than proves, Theorem~\ref{thm:frontier}.
The high-mismatch regret/deployment-gap pairs in panel (b) are $(1.732,30.607)$ for CAUSAL-UCB, $(16.207,4.120)$ for Greedy-Raw, and $(6.326,23.217)$ for Certified IPS-UCB, exposing distinct decision and measurement costs across these methods.

A matched-log $2\times2$ DR/IPS-by-exposure study at $\rho=1.40$ separately examines estimator and exposure effects (Appendix~\ref{app:matched-log-studies}, Figure~\ref{fig:matched-log-dr-exposure}).
Using identical logs across the four conditions separates these effects from differences in data collection.

\subsection{Trace-Driven Transfer and Scope}
\label{subsec:real-validation}

\begin{wraptable}[15]{r}{0.43\columnwidth}
\vspace{-3\baselineskip}
\centering
\setlength{\abovecaptionskip}{1pt}
\setlength{\belowcaptionskip}{3pt}
\caption{\textbf{High-mismatch semi-synthetic comparison.}
Mean cumulative target regret $\pm$ 95\% Student-$t$ CI
half-width over eight seeds ($T=180$). Low-mismatch results
are summarized in Appendix~\ref{app:trace-comparisons}.}
\label{tab:semi-syn-high}

{%
\fontsize{6.2}{7.0}\selectfont
\setlength{\tabcolsep}{1.0pt}
\renewcommand{\arraystretch}{0.98}
\begin{tabularx}{\linewidth}{
@{}
>{\raggedright\arraybackslash}p{0.34\linewidth}
>{\raggedleft\arraybackslash}p{0.34\linewidth}
*{3}{>{\raggedleft\arraybackslash}X}
@{}
}
\toprule
Method
& Regret $\downarrow$
& Gap $\downarrow$
& ATE $\downarrow$
& Dep. $\downarrow$ \\
\midrule

\multicolumn{5}{@{}l}{\textit{METR-LA}}\\[-1pt]
\rowcolor{CausalBlue}
\textbf{CAUSAL-UCB}
& \textbf{0.754 $\pm$ 0.035}
& \textbf{0.000}
& \textbf{0.0466}
& 12.681 \\

IPS
& 0.774 $\pm$ 0.036
& \textbf{0.000}
& 0.0670
& 12.663 \\

Raw
& 8.438 $\pm$ 0.090
& 0.026
& 0.0703
& 4.177 \\

Greedy
& 5.640 $\pm$ 0.097
& 0.030
& 0.0743
& \textbf{3.567} \\

Switch
& 8.745 $\pm$ 0.000
& 0.055
& 0.0713
& 4.116 \\

\midrule
\multicolumn{5}{@{}l}{\textit{FlockLab}}\\[-1pt]
\rowcolor{CausalBlue}
\textbf{CAUSAL-UCB}
& \textbf{0.735 $\pm$ 0.000}
& \textbf{0.000}
& \textbf{0.0218}
& 12.716 \\

IPS-UCB
& 1.612 $\pm$ 1.582
& 0.004
& 0.1284
& 11.295 \\

Raw-UCB
& 8.532 $\pm$ 0.034
& 0.041
& 0.0735
& 4.108 \\

Greedy-Raw
& 5.685 $\pm$ 0.000
& 0.030
& 0.0701
& \textbf{3.435} \\

Switchback-OED
& 8.745 $\pm$ 0.000
& 0.055
& 0.0724
& 4.116 \\

\bottomrule
\end{tabularx}
}
\par\vspace{2pt}
{\raggedright\fontsize{5.5}{6.3}\selectfont
Gap: final target gap; ATE: pairwise ATE error;
Dep.: cumulative deployment gap.\par}
\vspace{-2pt}
\end{wraptable}

We test transfer on METR-LA~\cite{li2018dcrnn_traffic}
and FlockLab~\cite{jacob2019flocklab}
topology/context traces with model-generated
counterfactual outcomes (180 rounds, eight independent
seeds per condition).
These are semi-synthetic replays, not live deployments.

Under high mismatch, CAUSAL-UCB reduces regret relative to IPS-UCB on METR-LA ($0.754$ vs. $0.774$) and FlockLab ($0.735$ vs. $1.612$), but has higher deployment gaps than Greedy-Raw.
Under low mismatch, Greedy-Raw wins on both traces ($2.010$ vs. $3.126$; $0.735$ vs. $5.472$), although CAUSAL-UCB attains the lowest FlockLab ATE error ($0.0048$).
The benefits are regime-dependent.
Controlled-ring, trace-replay, and 300-seed audit protocols remain separate; eight-seed trace intervals are marginal, not paired.
General certificates require calibrated target-action, exposure, context-transport, clipping, and dependence budgets; the anytime result remains restricted to known interference (Appendices~\ref{app:frontier-detailed}, \ref{app:certificate-scope-clarification}, and~\ref{app:v8-bias-growth}).

\section{Conclusion}
\label{sec:conclusion}

CAUSAL-UCB selects by certified no-exploration steady-state value.
We prove a uniform certificate, an identification--deployment lower bound, and regret under calibrated conditions; known local interference yields an anytime radius for estimation, selection, and reporting.
Experiments show high-mismatch gains, deployment costs, and low-mismatch reversals across experimental regimes.

\begin{ack}
Zhiheng Zhang acknowledges support from the Fundamental
Research Funds for the Central Universities at Shanghai
University of Finance and Economics (Grant No. 2025110602)
and the Shanghai Engineering Research Center of Finance
Intelligence (Grant No. 19DZ2254600).

The authors declare no competing interests.
\end{ack}

\clearpage
\appendix

\setcounter{topnumber}{4}
\setcounter{bottomnumber}{2}
\setcounter{totalnumber}{6}

\renewcommand{\topfraction}{0.92}
\renewcommand{\bottomfraction}{0.85}
\renewcommand{\textfraction}{0.05}
\renewcommand{\floatpagefraction}{0.80}

\renewcommand{\thesection}{\Alph{section}}

\section{Problem Origin and Detailed Closest-Work Comparison}
\label{app:problem-origin-positioning}
The main text establishes the decision primitive: an operator selects a global configuration from post-burn-in exploratory telemetry, but evaluates its \emph{no-exploration steady-state value}.
The identification obstruction and its precise assumptions are developed in Appendix~\ref{app:v8-bias-growth}; this section locates the objective relative to the closest research lines.
\input{appendix/30_related_work_scope_matrix.tex}

\section{Extended Setup, Notation, and Estimation Foundations}
\input{appendix/02_detailed_notation_timescales_and_logged_probability_spa.tex}
\input{appendix/01_technical_foundations_and_proof_roadmap.tex}

\input{appendix/04_target_value_and_logging_law_derivation.tex}
\input{appendix/03_post_burn_in_blocks_effective_sample_calibration_and_ex.tex}

\input{appendix/05_mixture_logging_ratio_bounds.tex}

\input{appendix/06_oracle_doubly_robust_identity.tex}

\input{appendix/07_clipping_bias_and_boundedness.tex}
\input{appendix/08_plug_in_nuisance_error_exact_three_term_decomposition.tex}
\input{appendix/09_exposure_approximation_and_enrichment_diagnostic.tex}
\input{appendix/10_context_shift_and_cross_configuration_mismatch.tex}
\input{appendix/11_configuration_aware_nuisance_fitting.tex}

\section{Assumptions, Calibration, and Information Growth}
\input{appendix/15_finite_sample_conditions_for_causal_value_certification.tex}
\input{appendix/16_calibrated_error_budgets_and_information_growth.tex}

\input{appendix/29_identification_calibration_scope_and_growth.tex}

\section{Full Proofs and Decision-Theoretic Guarantees}
\input{appendix/12_fixed_configuration_concentration.tex}
\input{appendix/13_uniformization_over_the_configuration_space.tex}

\input{appendix/14_proof_of_theorem_1_conditional_certificate.tex}
\input{appendix/17_identification_deployment_frontier_a_slow_mixing_steady.tex}

\input{appendix/18_causal_ucb_regret_proof.tex}

\input{appendix/19_deployment_gap_corollary.tex}

\input{appendix/28_restricted_case_certificate_proofs.tex}

\section{Algorithms and Operational Diagnostics}
\input{appendix/20_diagnostic_algorithms.tex}

\section{Experimental Protocol, Data Provenance, and Metrics}
\input{appendix/21_code_and_evaluation_protocol.tex}

\input{appendix/27_experimental_data_sources_and_evaluation_scope.tex}
\input{appendix/26_metrics_paired_comparisons_and_pareto.tex}

\section{Additional Experiments, Robustness, and Mechanism Studies}
\input{appendix/22_five_level_robustness_and_parameter_sweeps.tex}

\input{appendix/23_trace_driven_semi_synthetic_comparisons_as_grouped_bars.tex}

\input{appendix/24_seed_level_distributions_and_pareto_summary.tex}
\input{appendix/25_matched_log_mechanism_studies.tex}

\input{appendix/31_certified_optimism_initial_information.tex}
\input{appendix/appendix_xxv3_ofu_18cell.tex}

\section{Certificate Audits, Misspecification, and Failure Boundaries}
\input{appendix/appendix_xxv2_coverage_300x160.tex}
\input{appendix/appendix_xxv5_calibration_misspecification.tex}
\input{appendix/appendix_xxv5_severity_addendum.tex}

\input{appendix/32_limitations_and_societal_implications.tex}

\end{document}

%% file: appendix/30_related_work_scope_matrix.tex
\subsection{Closest-Work Comparison and Scope Boundaries}
\label{app:related-work-scope}
The main text summarizes the literature by problem dimensions.
Here we clarify the closest methodological connections and the distinct estimand and guarantee regimes.

\begin{table}[!htbp]
\centering
\caption{Representative connections and precise scope distinctions.}
\label{tab:related-scope-detailed}
\footnotesize
\setlength{\tabcolsep}{2.5pt}
\renewcommand{\arraystretch}{1.06}
\begin{tabular}{@{}p{0.265\linewidth}p{0.315\linewidth}p{0.355\linewidth}@{}}
\toprule
\textbf{Research line} & \textbf{Core contribution} & \textbf{Distinction in this work}\\
\midrule
Interference and designs \citep{hudgens2008toward,aronow2017estimating,bojinov2023switchback}
& Potential outcomes, exposure-based inference, and temporal treatment designs.
& A global configuration induces local actions and a post-burn-in no-exploration target regime.\\

DR and orthogonal OPE \citep{dudik2011doubly,wang2017optimal,chernozhukov2018double}
& Logged-action correction and nuisance-robust value estimation.
& The certificate separately budgets exposure, target-context transport, and dependence errors.\\

Longitudinal / latent evaluation \citep{uehara2023future,hess2026overlap}
& Evaluation under partial observability and overlap-weighted temporal treatment effects.
& The estimand here is a global configuration's steady-state value without measurement exploration.\\

Contextual and network bandits \citep{schneider2023cross,agarwal2024mabn,jia2025mabi}
& Online learning with unknown context laws or interacting rewards.
& Selection uses certified causal values from exploration-perturbed network telemetry.\\

Design-based inference--regret \citep{wang2025design}
& Pareto trade-offs and anytime asymptotic inference under network interference.
& Our lower bound concerns identification versus measurement-induced deployment loss in a specified stationary, potentially slow-mixing family.\\
\bottomrule
\end{tabular}
\end{table}

Additional connections arise from heterogeneous causal
inference and network geometry.
Collaborative causal estimation under heterogeneous
data distributions~\cite{guo2024collaborative}
addresses cross-site statistical heterogeneity,
whereas geometric analyses of network causal
inference~\cite{farzam2024geometry}
study how graph structure can affect the reliability
of causal estimation.
Neither line directly provides a predecision
certificate for the no-exploration steady-state
value of a global configuration under
measurement-induced regime changes.

%% file: appendix/02_detailed_notation_timescales_and_logged_probability_spa.tex
\subsection{Notation and Logged Probability Space}
\label{app:notation-detailed}

\paragraph{Compact notation table.}
Table~\ref{tab:master-notation} summarizes the main symbols and experimental settings.
Related symbols are grouped by model component; theorem-calibrated quantities and diagnostic-only proxies are distinguished explicitly.
The main causal evaluation uses the pre-action context $C^-$ and the joint action/exposure variable $\widetilde A=(A,E)$.
The original logged tuple $O$ is distinguished from the derived scoring record $\widetilde O$.

\begingroup
\fontsize{7.5}{8.6}\selectfont
\setlength{\LTleft}{0pt}
\setlength{\LTright}{0pt}
\setlength{\LTcapwidth}{\textwidth}
\setlength{\tabcolsep}{2pt}
\renewcommand{\arraystretch}{0.80}
\begin{longtable}{@{}>{\raggedright\arraybackslash}p{0.215\textwidth}>{\raggedright\arraybackslash}p{0.265\textwidth}>{\raggedright\arraybackslash}p{0.215\textwidth}>{\raggedright\arraybackslash}p{0.265\textwidth}@{}}
\caption{\textbf{Notation and experimental protocols.}}\label{tab:master-notation}\\
\toprule
\textbf{Symbol(s)} & \textbf{Meaning} & \textbf{Symbol(s)} & \textbf{Meaning}\\
\midrule
\endfirsthead
\caption[]{Notation and experimental protocols (continued)}\\
\toprule
\textbf{Symbol(s)} & \textbf{Meaning} & \textbf{Symbol(s)} & \textbf{Meaning}\\
\midrule
\endhead
\midrule
\multicolumn{4}{r}{\footnotesize Continued on next page}\\
\endfoot
\bottomrule
\endlastfoot
\multicolumn{4}{@{}l}{\textbf{\textit{Network, logging, and causal target}}}\\[3pt]
$N,\mathcal Z,M$ & Nodes, configuration set, and its size. & $i,t,T$ & Node, deployment round, horizon.\\
$z,z_t,z^\star$ & Candidate, chosen, and optimal configuration ($\arg\max_z W_\infty(z;0)$). & $\tau,T_{\mathrm{burn}},s,T_0$ & Fast time, burn-in, retained-window index/count.\\
$Y_\tau^{(t)},H_{t-1}$ & Latent network state and predecision history. & $\mathcal D_t,O_{i,s}^{(t)}$ & Round-$t$ logs and raw $(A,C,R)$ tuple.\\
$A,R,B_u,C^-$ & Local action, response ($|R|\le B_u$), and pre-action context. & $E,\phi,\phi^+$ & Exposure and baseline/enriched maps.\\
$\widetilde A,\widetilde O$ & Joint treatment $(A,E)$ and score record $(\widetilde A,C^-,R)$. & $\eta_t,\eta,\xi$ & Exploration rates and exploratory action law.\\
$\pi_{z,0},\pi^{\mathrm{obs}}_{z,\eta}$ & Target and logged local-action laws. & $g_{z,0},g^{\mathrm{obs}}_{z,\eta}$ & Target and logged joint action/exposure laws.\\
$g_{b,r},\widehat g_{b,r}$ & Behavior joint law and its estimate. & $\nu_{z,\eta},\nu_{z,0}$ & Pre-action context laws with/without exploration.\\
$\nu_b^z,\nu_{\mathrm{tr}}^z,\widehat\nu$ & Logging/training context laws and an estimated context law. & $d_{\mathcal F},K_h$ & Context discrepancy and its error-envelope map.\\
$m_z,m_{\mathrm{log}}^z,\widehat m_z$ & Target/logging response regressions and fitted nuisance. & $\mu_z,\widehat\mu_z$ & Target-law response average and fitted counterpart.\\
$W_\infty,\widehat W_t,\widehat W_{z,n}$ & No-exploration value and time/prefix estimates (indexed by $z$). & $\widehat\Delta_{\mathrm{exp},t}$ & Exposure-enrichment sensitivity diagnostic.\\
$\mathrm{Reg}_T,\mathrm{Gap}$ & Cumulative target regret; terminal optimality gap. & $\mathrm{ATE},\mathrm{ATE\ error}$ & Treatment effect; maximum pairwise error.\\
$\mathrm{DeploymentGap},L_W$ & Exploration-induced loss; value's exploration-rate Lipschitz constant. & $\rho,\Delta,\sigma$ & Interference strength, value gap, noise scale.\\
\addlinespace[0.4pt]
\multicolumn{4}{@{}l}{\textbf{\textit{DR estimation, certificates, and dependence}}}\\[3pt]
$\widehat\psi_{z,r},\psi_z$ & Fitted and oracle DR scores. & $\widehat w_{z,r},w_{\max},\pi_{\min}$ & Clipped joint ratio, cap, denominator floor.\\
$B_\psi,\delta,\delta_t$ & Score bound $2B_u(1+w_{\max})$; $\sum_t\delta_t\le\delta$. & $\operatorname{comp}(\mathcal Z)$ & Configuration-class complexity.\\
$b_t(z),b_{z,n}$ & Predecision/finite-prefix certificate radius. & $E_t(z),e_t(z)$ & Total predecision error, including excess drift.\\
$\epsilon_{m,t},\epsilon_{\pi,t}$ & Response-nuisance and fitted joint-ratio errors. & $\bar\epsilon_{\mathrm{target},t},\epsilon_{\mathrm{clip},t}$ & Target-law error and clipping bias.\\
$\epsilon_{\mathrm{exp}},\bar\epsilon_{\mathrm{shift},t}$ & Exposure/response-transport and context-transport budgets. & $\bar\epsilon_{\mathrm{cfg},t},\bar\epsilon_{\mathrm{tot},t}$ & Configuration mismatch and total calibrated budget.\\
$\widehat\kappa_t,N_z(t)$ & Excess block-drift diagnostic; times $z$ is selected. & $\gamma_\ell,\rho_\ell$ & Score lag covariance and autocorrelation.\\
$\widehat n_{\mathrm{eff},t},\widetilde n_{\mathrm{eff},t}$ & Plug-in ESS and operational shrinkage proxy: diagnostics only. & $\underline n_{\mathrm{eff},t}$ & Dependence-calibrated ESS lower bound used in the theorem.\\
$B_t,G_t,L_t,c_{\mathrm{ess}},c_B$ & Block/group/lag settings and operational constants. & $\beta(k),\tau_{\mathrm{mix}}$ & Absolute regularity and mixing time.\\
$c_1,c_2,\lambda,\gamma$ & Lower-bound constants; information growth and approximation exponent. & $\mathrm{CI}_{95}$ & 95\% interval (paired-$t$ or Wilson, by study).\\
\addlinespace[0.4pt]
\multicolumn{4}{@{}l}{\textbf{\textit{Finite-state audit and supplemental diagnostics}}}\\[3pt]
$J,K,D$ & Contexts, actions, and audit-cell/union-bound dimensions. & $p_z,\widetilde p,\widetilde b,\widetilde m$ & Target action law and independent audit means.\\
$n_{\min},\alpha,r,\epsilon(m)$ & Audit counts, error budget, radius, and mismatch envelope. & $e_p,e_b,e_m,e_w,e_\nu$ & Audit errors: actions, response, weights, context.\\
$q,K_j,m_{j-},m_n,h_j$ & Neighborhood-match probabilities/indicators, counts, weights. & $\text{budget}$ & Scored-window budget per decision.\\
\addlinespace[0.4pt]
\multicolumn{4}{@{}l}{\textbf{\textit{Experimental protocols (not theoretical assumptions)}}}\\[3pt]
Main benchmark & $N=24$, $M=5$, $T=200$, $T_0=2$; 24 paired seeds; $\eta\in\{.55,.08\}$, $\sigma=.08$. & Seven-term coverage & 300 seeds, 160 rounds; 2,000 scored windows/decision; 10k fit/200k audit observations per cell.\\
OFU--Greedy & Five regimes, 96 paired seeds, 80 rounds; 400 blocks/round; 10k fit/200k audit observations per cell. & Gap--noise--budget & $\Delta\in\{.005,.016,.030\}$, $\sigma\in\{.2,.6,1\}$; 200/400 windows, 24 seeds, 80 rounds/cell.\\
Concealed misspecification & 64 paired seeds, 120 rounds, 1,200 windows/round; $\xi\in\{0,.10,.20,.35,.50\}$. & METR-LA / FlockLab & Observed topology/context traces; model-generated outcomes (semi-synthetic).\\
\end{longtable}
\endgroup

\paragraph{Slow and fast time.}
The operator chooses a deployment at slow round $t=1,\ldots,T$.
Within each round, the system evolves for $\tau=1,\ldots,T_{\mathrm{burn}}+T_0$: the first $T_{\mathrm{burn}}$ steps are burn-in, and the final $T_0$ windows are retained.
The raw observation for node $i$ and retained window $s$ is
\[
O_{i,s}^{(t)} = (A_{i,s}^{(t)},C_{i,s}^{(t)},R_{i,s}^{(t)}).
\]
For causal scoring, the raw log is supplemented with the exposure constructed from the observed network telemetry.
We write
\[
E_{i,s}^{(t)} = \phi_i\!\left(
\{A_{j,s}^{(t)}\}_{j\in\mathcal N_i(s)},
S_s^{(t)},G_s^{(t)} \right),
\qquad
\widetilde A_{i,s}^{(t)}
= (A_{i,s}^{(t)},E_{i,s}^{(t)}),
\]
and define the corresponding scoring record as
\[
\widetilde O_{i,s}^{(t)}
= \left(
\widetilde A_{i,s}^{(t)},
C_{i,s}^{-,(t)}, R_{i,s}^{(t)} \right).
\]
Here $C_{i,s}^{-,(t)}$ contains only information available strictly before the focal action is generated.
Contemporaneous neighboring actions and telemetry used to construct $E_{i,s}^{(t)}$ belong to the observed joint action/exposure variable, not to $C_{i,s}^{-,(t)}$.
The raw log retains the telemetry needed to construct this scoring record.

\paragraph{Prequential scoring convention.}
The main-text notation suppresses the fitting cutoff on the target response regression and target joint law.
In the cumulative estimator, $\widehat\psi_{z,r}$ denotes the score rule stored for source round $r$: its target-law fit, response-regression fit, and behavior joint-law fit are frozen before the corresponding scoring window is observed.
Scores from earlier rounds are retained as stored; they are not retrospectively recomputed using nuisance fits trained on later observations.
If nuisance fits are updated within a deployment round, the same convention applies at the network-window level, with a window-specific stored score rule.
This is the predictability convention used in the filtration-based concentration arguments.

\paragraph{Latent state and history.}
The full state $Y_\tau^{(t)}$ may include queues, controller variables, channel histories, routing tables, or local filters; it and its transition kernel are unknown to the operator.
Only logged tuples and runtime diagnostics (e.g., score autocorrelation, block drift, exposure histograms, and context two-sample statistics) are observed.
Before round $t$, the history is
\[
H_{t-1} = \sigma\!\left(
\left\{
z_r,\eta_r,O_{i,s}^{(r)},
\underline n_{\mathrm{eff},r},
\widehat\kappa_r, \bar\epsilon_{\mathrm{shift},r}, \bar\epsilon_{\mathrm{cfg},r}, \widehat\Delta_{\mathrm{exp},r}: r<t,\ i\le N,\ s\le T_0 \right\} \right).
\]
It contains realized deployments and telemetry, not counterfactual observations for configurations that were not deployed.
Their values must therefore be evaluated through the exposure interface, the logging law, and the off-policy estimator.
In particular, causal evaluation of a candidate configuration requires assumptions that relate the source-regime response to the candidate's target response; the behavior propensity correction alone does not establish this cross-configuration response relationship.

\paragraph{Target and logging distributions.}
Let $\nu_{z,\eta}$ be the post-burn-in distribution of pre-action contexts $C^-$ under configuration $z$ and exploration rate $\eta$.
Let $\pi_{z,0}(a\mid c^-)$ denote the target conditional law of the focal local action without measurement exploration, and let $\xi(a\mid c^-)$ be the known exploratory local-action law.
The local-action logging mechanism retains its original mixture form:
\[
\pi^{\mathrm{obs}}_{z,\eta}(a\mid c^-) = (1-\eta)\pi_{z,0}(a\mid c^-) + \eta\xi(a\mid c^-).
\]
This equation describes the conditional law of the focal local action.
It does not, by itself, imply the same mixture identity for the joint action/exposure variable $\widetilde A$, because the exposure may depend on multiple contemporaneous actions and network telemetry.

Write $g_{z,0}(\widetilde a\mid c^-)$ for the target joint action/exposure law under no-exploration deployment, and $g^{\mathrm{obs}}_{z,\eta}(\widetilde a\mid c^-)$ for the joint law induced by the actual logging mechanism.
The latter must be derived from that mechanism; it is not obtained merely by substituting the marginal local-action mixture into a joint propensity.
For a historical observation generated in round $r$, the behavior joint law corresponds to its actual source regime $(z_r,\eta_r)$ and is denoted $g_{b,r}(\widetilde a\mid c^-)$.
Its fitted version is denoted $\widehat g_{b,r}(\widetilde a\mid c^-)$.

The target value uses the context law $\nu_{z,0}$ and the target joint law $g_{z,0}$:
\[
W_\infty(z;0) =
\mathbb E_{C^-\sim\nu_{z,0}}
\left[
\int
m_z(\widetilde a,C^-)\, g_{z,0}(d\widetilde a\mid C^-) \right],
\]
where $m_z(\widetilde a,c^-)$ is the target-regime conditional response under the specified joint action/exposure treatment.
The corresponding target conditional mean is
\[
\mu_z(c^-) =
\int
m_z(\widetilde a,c^-)\, g_{z,0}(d\widetilde a\mid c^-).
\]
For discrete joint treatments, these integrals are sums.
For continuous joint treatments, the propensity ratio is interpreted as a density ratio with respect to a common dominating measure.

The target joint law and context law differ from the laws under which historical exploratory observations are collected.
The joint importance ratio corrects treatment/exposure-law mismatch only under the corresponding identification, overlap, and response-transport conditions.
Context transport from $\nu_{z,\eta}$ to $\nu_{z,0}$ and cross-configuration response transport remain distinct requirements; they are not established by the local-action mixture identity alone.

%% file: appendix/01_technical_foundations_and_proof_roadmap.tex
\subsection{Technical Foundations and Proof Roadmap}\label{app:overview}
This appendix provides derivations and calibration details for the estimator, certificate, frontier, and regret guarantee.
The proof chain is:
\begin{enumerate}[leftmargin=*]
  \item Appendix~\ref{app:notation-detailed} defines the logged probability space, timescales, and random variables.
  \item Appendix~\ref{app:target-logging} derives the no-exploration target and the relationship between target and logging laws.
  \item Appendix~\ref{app:ss-blocks} defines the post-burn-in score sequence, the plug-in effective sample size, its operational shrinkage proxy, and the excess block-drift diagnostic. The certified lower-information quantity $\underline n_{\mathrm{eff},t}$ requires the separate dependence calibration stated there.
  \item Appendix~\ref{app:mixture-ratio} records the mixture-induced ratio bounds and explains why the clipped stochastic term is controlled by $B_\psi$ rather than by an additional uncontrolled $1/\eta$ variance factor.
  \item Appendix~\ref{app:dr-identity-detailed} proves the oracle doubly robust identity.
  \item Appendices~\ref{app:clipping-detailed}--\ref{app:context-shift-detailed} derive clipping, nuisance product error, exposure sensitivity, context-shift proxies, and cross-configuration mismatch proxies.
  \item Appendix~\ref{app:configuration-aware-nuisance-fitting} gives the configuration-aware nuisance fitting rule used in the algorithm.
  \item Appendices~\ref{app:concentration-detailed}--\ref{app:uniform-detailed} prove fixed-configuration and uniform deviation bounds.
  \item Appendices~\ref{app:v7-certification}--\ref{app:gap-detailed} prove the main certificate, frontier theorem, regret theorem, and deployment-gap corollary.
  \item Appendices~\ref{app:diagnostic-algorithms}--\ref{app:new-exp-protocol} give the executable diagnostic and experimental protocol.
\end{enumerate}

%% file: appendix/04_target_value_and_logging_law_derivation.tex
\subsection{Target Value and Logging Law Derivation}
\label{app:target-logging}

The no-exploration target for a candidate configuration $z$ is
\begin{equation}
\Winf(z;0)
=
\mathbb E_{C^-\sim\nu_{z,0}}
\left[
\int_{\widetilde{\mathcal A}}
m_z(\widetilde a,C^-)\,
g_{z,0}(d\widetilde a\mid C^-)
\right].
\label{eq:app-target}
\end{equation}
Here $C^-$ denotes the pre-action context, $\widetilde A=(A,E)$ is the joint action/exposure treatment unit, $g_{z,0}(\cdot\mid C^-)$ is its target conditional distribution under no-exploration deployment, and $m_z(\widetilde a,C^-)$ is the corresponding target-regime conditional response mean.
For discrete joint treatments, the integral is a sum; for continuous joint treatments, it is taken with respect to the target probability measure.

For deployment under configuration $z$ and exploration rate $\eta$, let $\nu_{z,\eta}$ denote the post-burn-in distribution of pre-action contexts.
The observed focal local action follows the original measurement-level mixture
\[
\pi^{\mathrm{obs}}_{z,\eta}(a\mid c^-) = (1-\eta)\pi_{z,0}(a\mid c^-) +\eta\xi(a\mid c^-).
\]
Consequently,
\[
\pi^{\mathrm{obs}}_{z,\eta}(a\mid c^-)
\ge \eta\xi(a\mid c^-).
\]
If $\xi(a\mid c^-)\ge\xi_{\min}>0$ and $\eta>0$, the focal-action target-to-logging ratio obeys
\[
\frac{\pi_{z,0}(a\mid c^-)}
{\pi^{\mathrm{obs}}_{z,\eta}(a\mid c^-)}
\le
\frac{1}{\eta\xi_{\min}}.
\]
This is a positivity bound for the marginal focal-action law only.
It is not, by itself, a bound on the joint action/exposure importance ratio.

Let $g^{\mathrm{obs}}_{z,\eta}(\cdot\mid c^-)$ denote the joint action/exposure law induced by the actual logging mechanism, including the contemporaneous neighboring actions and the exposure mapping.
In general, the local-action mixture above does not imply
\[
g^{\mathrm{obs}}_{z,\eta}(\widetilde a\mid c^-) = (1-\eta)g_{z,0}(\widetilde a\mid c^-) +\eta g_{\xi}(\widetilde a\mid c^-)
\]
for any particular joint exploratory law $g_\xi$ unless that identity is established from the joint action-generation mechanism.
In particular, the exposure component may depend on multiple contemporaneous actions and network telemetry, and the joint law must account for those dependencies.

For a historical observation collected in round $r$, let $(z_r,\eta_r)$ be its actual source regime.
Its behavior joint law is the law induced by that source regime, denoted $g^{\mathrm{obs}}_{z_r,\eta_r}$ when the mechanism is known, or $g_{b,r}$ when represented by the corresponding behavior-law notation.
The candidate target law remains $g_{z,0}$; these laws need not coincide when $z_r\ne z$.

The joint importance ratio used for causal evaluation is
\[
w^{\mathrm{joint}}_{z,r}(\widetilde a,c^-) =
\frac{g_{z,0}(\widetilde a\mid c^-)}
{g_{b,r}(\widetilde a\mid c^-)}.
\]
For continuous joint treatments, this ratio denotes the Radon--Nikodym derivative of the target joint law with respect to the behavior joint law, whenever that derivative exists.
A necessary overlap condition is
\[
g_{z,0}(\cdot\mid c^-)
\ll
g_{b,r}(\cdot\mid c^-)
\]
for the relevant target contexts and historical source regimes.
Any finite upper bound on this joint ratio must be established from the actual joint action/exposure mechanism; it does not follow solely from the marginal mixture formula.
The clipped ratio used by the estimator limits the score's range, while any resulting clipping bias must be included in the calibrated error budget.

Finally, the target context distribution $\nu_{z,0}$ may differ from the context distribution under which historical observations were collected.
Joint action/exposure weighting addresses the treatment-law mismatch conditional on context; it does not, by itself, transport the context distribution or identify a candidate configuration's response from observations generated under a different configuration.
Those discrepancies require the separately stated context-transport and cross-configuration response-transport conditions in the main certificate.

%% file: appendix/03_post_burn_in_blocks_effective_sample_calibration_and_ex.tex
\subsection{Post-Burn-in Blocks, Effective Sample Calibration, and Excess Drift}
\label{app:ss-blocks}
This section defines the post-burn-in network-window score sequence, its plug-in effective sample size, the operational shrinkage proxy, and the excess block-drift diagnostic $\widehat\kappa_t$.
The certified lower- information quantity $\underline n_{\mathrm{eff},t}$ requires a separate dependence calibration.

\subsubsection{From network windows to the clipped score sequence}
For a fixed candidate configuration $z$, flatten the retained windows in observation order into $j=1,\ldots,J_t$, where $J_t=T_0t$ and $j$ indexes the pair $(r,s)$ of deployment round and retained window.
Define the network-window score
\[
Y_{j,z} = \frac1N\sum_{i=1}^N \widehat\psi_{z,r(j)}(\widetilde O_{i,s(j)}^{(r(j))}),
\qquad
\bar Y_t(z)=\frac1{J_t}\sum_{j=1}^{J_t}Y_{j,z}.
\]
The score rule stored for source round $r(j)$ is frozen before the corresponding network window is scored.
By the clipped-score bound in Appendix~\ref{app:clipping-detailed}, $|Y_{j,z}|\le B_\psi$.
The within-window dependence among nodes is unrestricted; the time-series unit for the concentration and autocorrelation calculations is the aggregated network-window score $Y_{j,z}$, not the $N J_t$ node-level records.
This convention matches the normalization of $\widehat W_t(z;0)$ in the main text.

\subsubsection{Variance identity for correlated network windows}
If the post-burn-in network-window sequence is approximately second-order stationary, define
\[
\gamma_\ell(z)=\operatorname{Cov}(Y_{j,z},Y_{j+\ell,z}),
\qquad
\rho_\ell(z)=\gamma_\ell(z)/\gamma_0(z).
\]
Then
\begin{align}
\operatorname{Var}\{\bar Y_t(z)\}
&=
\frac1{J_t^2}
\sum_{j=1}^{J_t}\sum_{j'=1}^{J_t}
\operatorname{Cov}(Y_{j,z},Y_{j',z})\notag\\
&=
\frac{\gamma_0(z)}{J_t}
\left[
1+2\sum_{\ell=1}^{J_t-1}
\left(1-\frac{\ell}{J_t}\right)\rho_\ell(z)
\right].
\end{align}
This motivates the plug-in effective sample size
\begin{equation}
\widehat n_{\mathrm{eff},t}(z)
=
\frac{J_t}
{1+2\sum_{\ell=1}^{L_t}\max\{\widehat\rho_{\ell,t}(z),0\}}.
\label{eq:app-neff}
\end{equation}
Positive autocorrelations reduce the estimated information.
Negative empirical autocorrelations are not used to claim extra precision, because they are often noisy in short sequences.

\subsubsection{Operational ESS shrinkage and calibration requirement}
The plug-in value $\widehat n_{\mathrm{eff},t}(z)$ is an empirical estimate.
Let $B_t$ be the number of post-burn-in blocks used to estimate autocorrelation.
Define the operational shrinkage proxy
\begin{equation}
\widetilde n_{\mathrm{eff},t}(z)
=
\max\!\left\{
1,
\frac{\widehat n_{\mathrm{eff},t}(z)}
{1+c_{\mathrm{ess}}\sqrt{\log(2T/\delta)/B_t}}
\right\}.
\label{eq:app-neff-shrinkage}
\end{equation}
This shrinkage rule is an operational heuristic, not a finite-sample lower confidence bound.
The quantity $\underline n_{\mathrm{eff},t}(z)$ used in Theorem~\ref{thm:main} must instead be supplied by a proved dependence argument or an independently calibrated lower-information bound for the sequence $\{Y_{j,z}\}$.
Such calibration must establish that the reported quantity does not exceed the information certified by the dependence model or audit.
Empirical autocorrelations alone do not establish coverage.
If no valid lower-information calibration is available, the general confidence certificate is not certified.

\subsubsection{Excess block drift}
The autocorrelation correction addresses temporal dependence but not continued drift after burn-in.
Partition the network-window sequence into adjacent blocks $\mathcal B_{1,t},\ldots,\mathcal B_{G_t,t}$ and define
\[
\bar Y_{g,t}(z)
= |\mathcal B_{g,t}|^{-1} \sum_{j\in\mathcal B_{g,t}}Y_{j,z}.
\]
The excess block-drift statistic is
\begin{equation}
\widehat\kappa_t(z)
=
\left[
\max_{2\le g\le G_t}
|\bar Y_{g,t}(z)-\bar Y_{g-1,t}(z)|
-
 c_B B_\psi
\sqrt{
\frac{\log(2TG_t/\delta)}
{\min_g|\mathcal B_{g,t}|}}
\right]_+.
\label{eq:app-kappa}
\end{equation}
If adjacent block means differ only by concentration-scale noise, $\widehat\kappa_t(z)=0$.
If they move beyond that scale, the observed excess flags potential residual drift.
The statistic is computable from the clipped network-window scores, but it enters a theorem-backed radius only when a dependence or audit argument validates it as an upper bound on residual mean drift at the stated confidence level.

%% file: appendix/05_mixture_logging_ratio_bounds.tex
\subsection{Mixture Logging Ratio Bounds}
\label{app:mixture-ratio}

This section records the ratio bounds induced by the measurement-level mixture logging mechanism.
We distinguish the marginal focal-action ratio from the joint action/exposure ratio, because the former does not, by itself, establish a bound on the latter.

\begin{lemma}[Marginal mixture ratio bounds]
\label{lem:mixture-ratio}
Let
\[
\pi^{\mathrm{obs}}_{z,\eta}(a\mid c^-) = (1-\eta)\pi_{z,0}(a\mid c^-) +\eta\xi(a\mid c^-),
\qquad
\xi(a\mid c^-)\ge \xi_{\min}>0,
\]
where $0<\eta<1$.
Then the marginal focal-action target-to-logging ratio
\[
w^\star_{z,\eta}(a,c^-) =
\frac{\pi_{z,0}(a\mid c^-)}
{\pi^{\mathrm{obs}}_{z,\eta}(a\mid c^-)}
\]
satisfies
\[
w^\star_{z,\eta}(a,c^-)
\le \frac{1}{1-\eta}
\quad
\text{whenever }\pi_{z,0}(a\mid c^-)>0,
\]
and also satisfies the conservative positivity bound
\[
w^\star_{z,\eta}(a,c^-)
\le \frac{1}{\eta\xi_{\min}}
\]
for all feasible $(a,c^-)$.
\end{lemma}

\begin{proof}
The first bound follows from
\[
\pi^{\mathrm{obs}}_{z,\eta}(a\mid c^-) = (1-\eta)\pi_{z,0}(a\mid c^-) +\eta\xi(a\mid c^-)
\ge
(1-\eta)\pi_{z,0}(a\mid c^-).
\]
Dividing by the positive denominator gives
\[
w^\star_{z,\eta}(a,c^-)
\le \frac{1}{1-\eta}
\]
whenever $\pi_{z,0}(a\mid c^-)>0$.
The second bound follows from
\[
\pi^{\mathrm{obs}}_{z,\eta}(a\mid c^-)
\ge \eta\xi(a\mid c^-)
\ge \eta\xi_{\min}
\]
and $\pi_{z,0}(a\mid c^-)\le 1$.
\end{proof}

\paragraph{Scope of the marginal bounds.}
Lemma~\ref{lem:mixture-ratio} concerns only the conditional distribution of the focal local action $A$ given the pre-action context $C^-$.
Its bounds do not automatically apply to the joint action/exposure variable $\widetilde A=(A,E)$.
In particular, the exposure $E$ may depend on neighboring actions and contemporaneous network telemetry, so the marginal mixture identity for $A$ does not by itself imply a corresponding mixture identity for $\widetilde A$.

\paragraph{Joint action/exposure ratio.}
For candidate configuration $z$, let $g_{z,0}(\cdot\mid c^-)$ denote the target joint action/exposure law under no-exploration deployment.
Let $g_{b,r}(\cdot\mid c^-)$ denote the behavior joint law associated with the actual source regime $(z_r,\eta_r)$ of historical round $r$.
The joint importance ratio is
\[
w^{\mathrm{joint}}_{z,r}(\widetilde a,c^-) =
\frac{g_{z,0}(\widetilde a\mid c^-)}
{g_{b,r}(\widetilde a\mid c^-)}
\]
when densities with respect to a common dominating measure exist and the denominator is positive on the target support.
More generally, the ratio is the Radon--Nikodym derivative of the target joint law with respect to the behavior joint law.

A necessary overlap condition is
\[
g_{z,0}(\cdot\mid c^-)
\ll
g_{b,r}(\cdot\mid c^-)
\]
for the relevant target contexts and historical source regimes.
This absolute-continuity condition ensures that target events are not assigned zero probability under the corresponding behavior law.
It does not, by itself, imply that the joint importance ratio is uniformly bounded.

For clarity, a bound analogous to the first marginal bound would follow under the additional, explicitly verified joint-mixture condition
\[
g_{b,r}(\cdot\mid c^-) = (1-\eta_r)g_{z,0}(\cdot\mid c^-) +\eta_r q_{z,r}(\cdot\mid c^-),
\qquad 0\le\eta_r<1,
\]
where $q_{z,r}$ is a probability law on the same joint action/exposure space.
Under this condition,
\[
w^{\mathrm{joint}}_{z,r}(\widetilde a,c^-)
\le \frac{1}{1-\eta_r}
\]
on the target support.
However, this bound applies only when the stated joint-mixture identity actually holds.
It cannot be inferred solely from the marginal local-action mixture, and it is not automatically valid when the historical source configuration $z_r$ differs from the candidate configuration $z$.

Accordingly, any finite upper bound on the joint ratio must be justified from the actual joint action-generation mechanism, including the induced exposure distribution and any cross-configuration change.
If such a bound cannot be established, the analysis must instead retain the appropriate overlap and clipping-bias conditions without claiming the marginal mixture bound for the joint ratio.

\paragraph{Interpretation.}
The marginal mixture bounds provide a positivity guarantee for the focal local action.
The joint action/exposure formulation used by the main-text doubly robust estimator requires the corresponding joint-law overlap condition and, where a numerical ratio bound is used, its own justification.
The cap $w_{\max}$ bounds the implemented clipped score; it does not establish joint overlap or eliminate clipping bias.
Any clipping bias must remain accounted for in $\epsilon_{\mathrm{clip},t}(z)$, as specified in the main certificate.

%% file: appendix/06_oracle_doubly_robust_identity.tex
\subsection{Oracle Doubly Robust Identity}
\label{app:dr-identity-detailed}

Fix a candidate configuration $z$.
Let $C^-$ denote the pre-action context and let $\widetilde A=(A,E)$ denote the joint action/exposure treatment unit.
Write $g_{z,0}(\widetilde a\mid c^-)$ for the target joint law under no-exploration deployment and
\[
\mu_z(c^-) = \int_{\widetilde{\mathcal A}} m_z(\widetilde a,c^-)\, g_{z,0}(d\widetilde a\mid c^-),
\]
where $m_z(\widetilde a,c^-)$ is the target-regime conditional response mean.

For a historical observation collected in round $r$, let $g_{b,r}(\widetilde a\mid c^-)$ denote the behavior joint law associated with its actual source regime.
Assume that the target joint law is absolutely continuous with respect to this behavior law on the relevant support.
The oracle joint importance ratio is
\[
w_{z,r}(\widetilde a,c^-) =
\frac{g_{z,0}(\widetilde a\mid c^-)}
{g_{b,r}(\widetilde a\mid c^-)}.
\]
For continuous joint treatments, this expression denotes the Radon--Nikodym derivative with respect to a common dominating measure.

Let $m_{b,r}(\widetilde a,c^-)$ denote the conditional mean of the observed response under the historical source regime.
The oracle DR score for candidate $z$ is
\[
\psi_{z,r}(\widetilde O) = \mu_z(C^-) + w_{z,r}(\widetilde A,C^-)
\{R-m_z(\widetilde A,C^-)\},
\qquad
\widetilde O=(\widetilde A,C^-,R).
\]

Conditionally on $C^-=c^-$, suppose the observed response has conditional mean $m_{b,r}(\widetilde a,c^-)$.
Then, whenever the importance ratio is well-defined and the relevant expectations exist,
\begin{align}
\mathbb E_{b,r}
\left[
\psi_{z,r}(\widetilde O)\mid C^-=c^-
\right]
&=
\mu_z(c^-)
+
\int_{\widetilde{\mathcal A}}
g_{z,0}(d\widetilde a\mid c^-)
\bigl[
m_{b,r}(\widetilde a,c^-)
-
m_z(\widetilde a,c^-)
\bigr].
\label{eq:dr-identity-source}
\end{align}
This follows by conditioning on the joint action/exposure variable and using $w_{z,r}\,g_{b,r}=g_{z,0}$ on the target support.

In particular, if the conditional response is invariant between the historical source regime and the target regime on the relevant support, so that
\[
m_{b,r}(\widetilde a,c^-) = m_z(\widetilde a,c^-),
\]
then the response-discrepancy term in Eq.~\eqref{eq:dr-identity-source} vanishes and
\[
\mathbb E_{b,r}
\left[
\psi_{z,r}(\widetilde O)\mid C^-=c^- \right] = \mu_z(c^-).
\]
Averaging this conditional identity over the target context law $\nu_{z,0}$ gives
\begin{align}
\mathbb E_{C^-\sim\nu_{z,0}}
\mathbb E_{b,r}
\left[
\psi_{z,r}(\widetilde O)\mid C^-
\right]
&=
\mathbb E_{C^-\sim\nu_{z,0}}[\mu_z(C^-)]
\notag\\
&=
W_\infty(z;0).
\label{eq:dr-identity-target}
\end{align}

The conditional identity and its target-context average require distinct qualifications when historical observations come from other configurations or exploration regimes.
If the conditional response differs across source and target regimes, the integral in Eq.~\eqref{eq:dr-identity-source} is a response-transport discrepancy and must be bounded by the separately calibrated response-transport budget.
If the historical context law differs from $\nu_{z,0}$, averaging under the historical context law instead of the target context law introduces an additional context-transport discrepancy.
Neither discrepancy is removed merely by correcting the joint action/exposure distribution.

Consequently, the oracle identity establishes the DR correction under the stated joint-treatment overlap and response conditions; it does not, by itself, identify the counterfactual response of an unobserved configuration.
The plug-in estimator inherits the corresponding target-joint-law estimation, nuisance-product, clipping, exposure/response-transport, context-shift, and cross-configuration mismatch terms, as accounted for in the main-text certificate and the following appendices.

%% file: appendix/07_clipping_bias_and_boundedness.tex
\subsection{Clipping Bias and Boundedness}
\label{app:clipping-detailed}

Fix a candidate configuration $z$ and a historical source round $r$.
Let $C^-$ denote the pre-action context and let $\widetilde A=(A,E)$ denote the joint action/exposure treatment unit.
Write $g_{z,0}(\cdot\mid c^-)$ for the target joint law and $g_{b,r}(\cdot\mid c^-)$ for the behavior joint law associated with the actual source regime in round $r$.

Assume the target joint law is absolutely continuous with respect to the behavior joint law on the relevant support.
Define the oracle joint importance ratio and its clipped version by
\[
w_{z,r}(\widetilde a,c^-) =
\frac{g_{z,0}(\widetilde a\mid c^-)}
     {g_{b,r}(\widetilde a\mid c^-)},
\qquad
\bar w_{z,r}(\widetilde a,c^-)
= \min\{w_{z,r}(\widetilde a,c^-),w_{\max}\}.
\]
For continuous joint treatments, the ratio is understood as the Radon--Nikodym derivative with respect to a common dominating measure.
The ratio is defined on the behavior support; values outside that support require the stated overlap condition.

\paragraph{Boundedness of the clipped score.}
Suppose the normalized response and the fitted response regression satisfy
\[
|R|\le B_u,
\qquad
|\widehat m_z(\widetilde a,c^-)|\le B_u.
\]
Suppose also that the fitted target joint law is a probability distribution and that
\[
\widehat\mu_z(c^-) = \int_{\widetilde{\mathcal A}}
\widehat m_z(\widetilde a,c^-)
\,\widehat g_{z,0}(d\widetilde a\mid c^-),
\]
so that $|\widehat\mu_z(c^-)|\le B_u$.
The fitted clipped weight satisfies $0\le\widehat w_{z,r}\le w_{\max}$.
Therefore, for the joint-treatment DR score
\[
\widehat\psi_{z,r}(\widetilde O) = \widehat\mu_z(C^-) +
\widehat w_{z,r}(\widetilde A,C^-)
\{R-\widehat m_z(\widetilde A,C^-)\},
\]
where $\widetilde O=(\widetilde A,C^-,R)$, we have
\begin{align}
|\widehat\psi_{z,r}(\widetilde O)|
&\le
B_u+w_{\max}\bigl(|R|+|\widehat m_z(\widetilde A,C^-)|\bigr)
\notag\\
&\le B_u(1+2w_{\max})
\le 2B_u(1+w_{\max})
=:B_\psi.
\label{eq:clipped-score-bound}
\end{align}
Thus the score bound used in the main-text concentration argument remains valid after replacing the marginal action weight by the clipped joint action/exposure weight.
The same bound applies to the oracle score when its response regression and target mean satisfy the corresponding boundedness conditions.

\paragraph{Clipping bias.}
Let $m_z(\widetilde a,c^-)$ denote the target-regime conditional response, with $|m_z|\le B_u$.
Define the oracle unclipped and clipped scores, respectively, by
\begin{align*}
\psi^{\mathrm{or}}_{z,r}(\widetilde O)
&=
\mu_z(C^-)
+
w_{z,r}(\widetilde A,C^-)
\{R-m_z(\widetilde A,C^-)\},\\
\psi^{\mathrm{clip}}_{z,r}(\widetilde O)
&=
\mu_z(C^-)
+
\bar w_{z,r}(\widetilde A,C^-)
\{R-m_z(\widetilde A,C^-)\},
\end{align*}
where
\[
\mu_z(c^-) = \int_{\widetilde{\mathcal A}} m_z(\widetilde a,c^-) \,g_{z,0}(d\widetilde a\mid c^-).
\]
Because $|R-m_z(\widetilde A,C^-)|\le 2B_u$, their pointwise difference obeys
\[
\left| \psi^{\mathrm{or}}_{z,r}(\widetilde O) - \psi^{\mathrm{clip}}_{z,r}(\widetilde O) \right|
\le
2B_u \bigl(w_{z,r}(\widetilde A,C^-)-w_{\max}\bigr)_+,
\]
where $(x)_+=\max\{x,0\}$.
Consequently, under any common context law $\nu$ for which the relevant expectations exist,
\begin{align}
&\mathbb E_{\substack{C^-\sim\nu\\
\widetilde A\sim g_{b,r}(\cdot\mid C^-)}}
\left[
\left|
\psi^{\mathrm{or}}_{z,r}(\widetilde O)
-
\psi^{\mathrm{clip}}_{z,r}(\widetilde O)
\right|
\right]
\notag\\
&\qquad\le
2B_u
\mathbb E_{\substack{C^-\sim\nu\\
\widetilde A\sim g_{b,r}(\cdot\mid C^-)}}
\left[
\bigl(w_{z,r}(\widetilde A,C^-)-w_{\max}\bigr)_+
\right].
\label{eq:joint-clipping-bias}
\end{align}
The right-hand side is the clipping-bias allowance for this source regime and context law.
When multiple historical rounds are pooled, the calibrated allowance must cover their corresponding averaged contributions; a bound for one round alone does not automatically bound the entire historical average.

\paragraph{Role in the causal value certificate.}
The clipping allowance $\epsilon_{\mathrm{clip},t}(z)$ must bound the historical-average truncation bias for the joint ratios used by the estimator.
It neither establishes overlap when target-supported events are impossible under the behavior law nor covers source-to-target response transport.
Likewise, fitted-ratio error is not included in the oracle clipping bound~\eqref{eq:joint-clipping-bias}; it belongs to the nuisance decomposition in Appendix~\ref{app:nuisance-detailed}.
The certificate must therefore retain separate calibrated allowances for target-law estimation, nuisance error, context transport, exposure approximation, and response transport.

%% file: appendix/08_plug_in_nuisance_error_exact_three_term_decomposition.tex
\subsection{Plug-in Nuisance Error: Exact Decomposition and Response Transport}
\label{app:nuisance-detailed}

Condition on the nuisance fits stored before a scoring window and on a common evaluation context law $\nu$ for the pre-action context $C^-$.
Fix a candidate configuration $z$ and a historical source round $r$.
For compactness, write
\[
p(\widetilde a\mid c^-)=g_{z,0}(\widetilde a\mid c^-),
\qquad
b(\widetilde a\mid c^-)=g_{b,r}(\widetilde a\mid c^-),
\]
for the target and behavior joint action/exposure laws.
Let $m_z(\widetilde a,c^-)$ be the target-regime conditional response mean and $m_{b,r}(\widetilde a,c^-)$ the conditional response mean under the historical source regime.

The fitted quantities below are the versions used by the stored score $\widehat\psi_{z,r}$; their fitting-cutoff indices are suppressed only for readability.
Let $\widehat p$, $\widehat b$, and $\widehat m_z$ denote the fitted target joint law, behavior joint law, and target response regression.
Define
\[
\widehat\mu_z(c^-) = \int_{\widetilde{\mathcal A}}
\widehat m_z(\widetilde a,c^-)
\widehat p(d\widetilde a\mid c^-),
\]
and suppose the fitted joint laws are valid probability distributions.
For continuous joint treatments, all ratios are density ratios with respect to a common dominating measure, and denominator regularization must be defined consistently with that measure.

Write
\[
w=\frac{p}{b},
\qquad
\bar w=\min\{w,w_{\max}\},
\qquad
\widehat w=
\min\left\{
\frac{\widehat p}{\max\{\widehat b,\pi_{\min}\}},
 w_{\max} \right\},
\]
where ratios are evaluated at $(\widetilde a,c^-)$.
Define
\[
\delta_m=\widehat m_z-m_z,
\qquad
\delta_w=\widehat w-\bar w,
\qquad
\Delta_{b,r,z}=m_{b,r}-m_z.
\]
Thus $\delta_w$ is the error in the *induced clipped joint importance ratio*, not merely the raw behavior-density estimation error.

The fitted joint-treatment DR score is
\[
\widehat\psi_{z,r}(\widetilde O) = \widehat\mu_z(C^-) +
\widehat w(\widetilde A,C^-)
\{R-\widehat m_z(\widetilde A,C^-)\},
\qquad
\widetilde O=(\widetilde A,C^-,R).
\]
Conditionally on $C^-=c^-$, the observed response has mean $m_{b,r}(\widetilde a,c^-)$ under the behavior law.
The exact identity is
\begin{align}
&\mathbb E_{b,r}
  [\widehat\psi_{z,r}(\widetilde O)\mid C^-=c^-]
-\int_{\widetilde{\mathcal A}}
  p(\widetilde a\mid c^-)
  m_z(\widetilde a,c^-)\,d\widetilde a
\notag\\
={}&
\int_{\widetilde{\mathcal A}}
  [\widehat p-p](\widetilde a\mid c^-)
  \widehat m_z(\widetilde a,c^-)\,d\widetilde a
\notag\\
&+
\int_{\widetilde{\mathcal A}}
  [p-b\bar w](\widetilde a\mid c^-)
  \delta_m(\widetilde a,c^-)\,d\widetilde a
\notag\\
&-
\int_{\widetilde{\mathcal A}}
  b(\widetilde a\mid c^-)
  \delta_w(\widetilde a,c^-)
  \delta_m(\widetilde a,c^-)\,d\widetilde a
\notag\\
&+
\int_{\widetilde{\mathcal A}}
  b(\widetilde a\mid c^-)
  \widehat w(\widetilde a,c^-)
  \Delta_{b,r,z}(\widetilde a,c^-)\,d\widetilde a.
\label{eq:repaired-nuisance}
\end{align}
For discrete joint treatments, the integrals above are sums.
The first three terms are the target-joint-law, clipping, and nuisance-product contributions.
The fourth term is a separate source-to-target response-transport contribution; it vanishes only under a justified conditional response-invariance assumption.
Correcting the joint action/exposure distribution alone does not remove this fourth term.

\paragraph{Bounds for the nuisance terms.}
Suppose
\[
|R|\le B_u,\qquad |m_z|\le B_u,\qquad |\widehat m_z|\le B_u,\qquad |m_{b,r}|\le B_u.
\]
For the chosen context law $\nu$, define
\[
\mathbb E_{\nu,b}[f]
=
\mathbb E_{C^-\sim\nu}
\mathbb E_{\widetilde A\sim b(\cdot\mid C^-)}
[f(\widetilde A,C^-)],
\qquad
\|f\|_{L_2(\nu,b)} = \left(\mathbb E_{\nu,b}[f^2]\right)^{1/2}.
\]
The first term is bounded by the first-order target-joint-law error
\begin{equation}
\epsilon_{\mathrm{target}}
=
B_u\,
\mathbb E_{C^-\sim\nu}
\left[
\|\widehat p(\cdot\mid C^-)-p(\cdot\mid C^-)\|_1
\right].
\label{eq:joint-target-nuisance-bound}
\end{equation}
Since
\[
p-b\bar w=b(w-w_{\max})_+,
\qquad (x)_+=\max\{x,0\},
\]
the second term is bounded by
\begin{equation}
\epsilon_{\mathrm{clip},r}
=
2B_u\,\mathbb E_{\nu,b}[(w-w_{\max})_+].
\label{eq:joint-clipping-nuisance-bound}
\end{equation}
Here $|\delta_m|\le2B_u$ has been used.

By Cauchy--Schwarz, the third term is bounded by
\begin{equation}
\epsilon_{m\pi,r}
=
\|\delta_m\|_{L_2(\nu,b)}
\|\delta_w\|_{L_2(\nu,b)}.
\label{eq:joint-product-nuisance-bound}
\end{equation}
Accordingly, the main-text $\epsilon_{m,t}(z)$ and $\epsilon_{\pi,t}(z)$ must control these two norms under the actual source-window score laws.
A sufficient pooled-prefix convention is to use uniform upper bounds over all source rounds and windows included in the prefix; alternatively, a directly calibrated historical average of the products may be used.
A bound on the raw behavior law alone is not sufficient unless it is shown to imply the required bound on $\|\widehat w-\bar w\|_{L_2(\nu,b)}$.

\paragraph{Source-to-target response transport.}
The fourth term in Eq.~\eqref{eq:repaired-nuisance} is bounded by
\begin{align}
\epsilon_{\mathrm{resp},r}(z)
&=
\mathbb E_{C^-\sim\nu}
\mathbb E_{\widetilde A\sim b(\cdot\mid C^-)}
\left[
\widehat w(\widetilde A,C^-)
|\Delta_{b,r,z}(\widetilde A,C^-)|
\right]
\notag\\
&\le
w_{\max}\,\mathbb E_{\nu,b}
[|\Delta_{b,r,z}|].
\label{eq:joint-response-transport-bound}
\end{align}
This bound controls the response-transport term when the score is compared directly with the target conditional mean.
In the five-term proof telescope, where the third term is compared with the oracle source-response functional $V_{z,r}^{\mathrm{log}}$, the corresponding score-level residual is instead the signed integral involving $(b\widehat w-p)\Delta_{b,r,z}$; it is defined and bounded in Appendix~\ref{app:thm1proof-detailed}.
The two expressions serve different decompositions and must not be substituted for one another without the associated algebra.

If the main-text exposure/response budget $\epsilon_{\mathrm{exp}}(z)$ is calibrated to dominate the required source-response terms for all admissible historical source regimes and all prefixes, those terms are included there.
Otherwise, an additional explicit response-transport allowance is required in the theorem's error budget.
Such an allowance must not be silently omitted or counted twice.

\paragraph{Implications for the certificate.}
After pooling, target-law, clipping, nuisance-product, and source-response contributions must be calibrated under the same source regimes, target support, and historical averaging scheme; a one-round bound is insufficient.
Target-law error remains first order, and neither cross-fitting nor estimating the marginal local-action mixture identifies the joint action/exposure law or an undeployed configuration's response.

%% file: appendix/09_exposure_approximation_and_enrichment_diagnostic.tex
\subsection{Exposure Approximation and Enrichment Diagnostic}
\label{app:exposure-majorant}

Let $L$ denote the full interference state, which may contain network information not fully captured by the observed exposure mapping.
Let $C^-$ denote the context available strictly before the focal action, and let $\widetilde A=(A,E)$ denote the observed joint action/exposure treatment unit, where $E$ is constructed from the available network telemetry.

Under any single fixed probability law, the tower property gives
\[
\mathbb E[\mathbb E[R\mid L,z]]
=
\mathbb E[R\mid z]
=
\mathbb E[\mathbb E[R\mid\widetilde A,C^-,z]].
\]
Consequently, the difference between these unconditional means under the same law cannot, by itself, quantify exposure approximation error.
Such error concerns whether the observed context and exposure representation is sufficient for the conditional response and transport arguments required by the causal estimand, rather than a difference between two representations of the same unconditional expectation.

Let $P_{z,0}$ denote the target-regime law under configuration $z$ without measurement exploration.
Write $\nu_{z,0}$ for its pre-action context law and $g_{z,0}(\cdot\mid c^-)$ for its conditional joint action/exposure law.
When the target interventional response is identified on this representation, denote it by $m_z^0(\widetilde a,c^-)$.
Thus,
\[
W_\infty(z;0) =
\mathbb E_{C^-\sim\nu_{z,0}}
\left[
\int_{\widetilde{\mathcal A}} m_z^0(\widetilde a,C^-) g_{z,0}(d\widetilde a\mid C^-) \right].
\]
This representation requires the corresponding causal identification assumptions under interference; it is not implied merely by defining an observed exposure mapping.
If the response $m_z$ used in the main-text score is not equal to this identified target response, the resulting representation discrepancy must be included in the exposure/response budget.

For a historical observation collected in source round $r$, let $m_{b,r}(\widetilde a,c^-)$ denote the conditional response under that historical source regime, evaluated at the target context and joint action/exposure arguments.
This notation presumes that the source response is meaningful on the relevant target support; the assumptions or error bounds needed to justify that transport must be stated separately.
Define the target-joint-law average of the source response by
\[
V_{z,r}^{\mathrm{log}}(c^-) = \int_{\widetilde{\mathcal A}} m_{b,r}(\widetilde a,c^-) g_{z,0}(d\widetilde a\mid c^-).
\]
The source index $r$ is retained because the response mechanism may vary with the deployed configuration and exploration regime.

The target-context source-response discrepancy is
\begin{equation}
\epsilon_{\mathrm{exp},r}^{\mathrm{target}}(z)
=
\left|
W_\infty(z;0)
-
\mathbb E_{C^-\sim\nu_{z,0}}
 V_{z,r}^{\mathrm{log}}(C^-)
\right|.
\label{eq:exposure-majorant}
\end{equation}
If the target response is identified as above and the relevant integrals exist, then
\begin{align}
\epsilon_{\mathrm{exp},r}^{\mathrm{target}}(z)
&=
\left|
\mathbb E_{C^-\sim\nu_{z,0}}
\int_{\widetilde{\mathcal A}}
[m_z^0(\widetilde a,C^-)-m_{b,r}(\widetilde a,C^-)]
 g_{z,0}(d\widetilde a\mid C^-)
\right|
\notag\\
&\le
\mathbb E_{C^-\sim\nu_{z,0}}
\int_{\widetilde{\mathcal A}}
|m_z^0(\widetilde a,C^-)-m_{b,r}(\widetilde a,C^-)|
 g_{z,0}(d\widetilde a\mid C^-).
\label{eq:exposure-response-majorant}
\end{align}
This is a sufficient bound for the response discrepancy under the target context and target joint action/exposure laws.
It may be zero under a justified conditional response-invariance assumption, but that invariance does not follow from importance weighting alone.

\paragraph{Scope of the main-text exposure/response budget.}
The quantity in Eq.~\eqref{eq:exposure-majorant} is only the target-context component of the transport argument.
In the proof of Theorem~\ref{thm:main}, the third telescope term also contains the score-level response residual defined in Eq.~\eqref{eq:score-level-response-transport} of Appendix~\ref{app:thm1proof-detailed}.
Therefore the single main-text budget $\epsilon_{\mathrm{exp}}(z)$ must be calibrated to dominate the exposure-representation discrepancy, the historical average of the score-level response residuals, and the historical average of the terms $\epsilon_{\mathrm{exp},r}^{\mathrm{target}}(z)$, uniformly over the admissible source regimes and prefixes used by the algorithm.
Equivalently, one may directly calibrate a tighter bound on the combined response/exposure contribution in the exact telescope.
A bound for Eq.~\eqref{eq:exposure-majorant} alone is not sufficient unless the score-level residual is separately shown to vanish or is otherwise covered.

The component terms are used to define one calibrated envelope $\epsilon_{\mathrm{exp}}(z)$ in the main theorem; they are not added again as separate terms in $\bar\epsilon_{\mathrm{tot},t}(z)$.
This avoids counting the same response-transport contribution twice.
If no single envelope has been calibrated for the actual scoring mechanism, the current theorem's stated error budget is not yet certified for that implementation.

\paragraph{Exposure enrichment as a sensitivity diagnostic.}
Let $\phi$ denote the baseline exposure mapping and let $\phi^+$ denote an enriched mapping using additional observed network information.
The associated change in the fitted value estimate, $\widehat\Delta_{\mathrm{exp},t}(z)$, can be used to assess sensitivity to the chosen exposure representation.
However, a small change between two fitted regressors is not a certified upper bound on exposure approximation error: both regressors may omit the same latent component of $L$.

Accordingly, the exposure/response transport term in Theorem~\ref{thm:main} is a valid error allowance only when its calibration covers the relevant representation and source-to-target response discrepancies over the historical windows used by the estimator.
The enrichment diagnostic provides supporting sensitivity evidence, not an independent high-probability guarantee.
Context -distribution shift and nuisance-training mismatch remain separate terms unless the calibration explicitly covers them.

%% file: appendix/10_context_shift_and_cross_configuration_mismatch.tex
\subsection{Context Shift and Cross-Configuration Mismatch}
\label{app:context-shift-detailed}

This section distinguishes context-distribution shift in score averaging from context-marginal mismatch in nuisance training.
Neither term is a substitute for the joint action/exposure overlap condition or for the source-to-target conditional response discrepancy defined in Appendix~\ref{app:exposure-majorant}.

\paragraph{Context-shift discrepancy.}
Fix a candidate configuration $z$ and a historical source round $r$, whose deployed configuration and exploration rate are $(z_r,\eta_r)$.
Let $\nu_{z,0}$ denote the target pre-action context law and let $\nu_{z_r,\eta_r}$ denote the source pre-action context law.
Define
\[
V_{z,r}^{\mathrm{log}}(c^-) = \int_{\widetilde{\mathcal A}} m_{b,r}(\widetilde a,c^-) g_{z,0}(d\widetilde a\mid c^-),
\]
where $m_{b,r}$ is the conditional response under the historical source regime and $g_{z,0}$ is the target joint action/exposure law.
This expression presumes that the source response is meaningful on the relevant target support and that the required integrals exist.

The context-shift discrepancy for this source round is
\begin{equation}
\epsilon_{\mathrm{shift},r}(z)
=
\left|
\mathbb E_{C^-\sim\nu_{z_r,\eta_r}}
 V_{z,r}^{\mathrm{log}}(C^-)
-
\mathbb E_{C^-\sim\nu_{z,0}}
 V_{z,r}^{\mathrm{log}}(C^-)
\right|.
\label{eq:context-shift-discrepancy}
\end{equation}
It compares the same source-response functional under two context laws, so it isolates context-distribution shift from conditional response transport.
Combining this definition with Eq.~\eqref{eq:exposure-majorant}, the triangle inequality gives
\begin{align}
\left|
W_\infty(z;0)
-
\mathbb E_{C^-\sim\nu_{z_r,\eta_r}}
 V_{z,r}^{\mathrm{log}}(C^-)
\right|
\le
\epsilon_{\mathrm{exp},r}^{\mathrm{target}}(z)
+
\epsilon_{\mathrm{shift},r}(z).
\label{eq:context-response-transport}
\end{align}

\paragraph{Distributional calibration of context shift.}
Let $d_{\mathcal F}$ be a context-distribution discrepancy, such as a suitably normalized integral probability metric.
If a validated calibration map $K_h$ satisfies
\[
\left|\mathbb E_\nu h(C^-)-\mathbb E_{\nu'}h(C^-)\right|
\le K_h\!\left(d_{\mathcal F}(\nu,\nu')\right)
\]
for the relevant function $h=V_{z,r}^{\mathrm{log}}$, then
\begin{equation}
\epsilon_{\mathrm{shift},r}(z)
\le
K_h\!\left(
d_{\mathcal F}(\nu_{z_r,\eta_r},\nu_{z,0})
\right).
\label{eq:context-shift-calibration}
\end{equation}
The calibration map and its conditions must be justified for the relevant function class.
An empirical two-sample statistic alone does not establish this high-probability bound, and an observed context discrepancy does not identify an otherwise unobserved target context law.

For the historical data available before round $t$, define
\begin{equation}
\bar\epsilon_{\mathrm{shift},t-1}(z)
=
\frac{1}{t-1}\sum_{r=1}^{t-1}\epsilon_{\mathrm{shift},r}(z),
\label{eq:averaged-context-shift}
\end{equation}
when the same source context law applies to all $T_0$ retained windows within each deployment round.
If the relevant conditional laws vary across retained windows, the allowance must instead be averaged over those windows with the same normalization as the historical estimator.
The bound used by the certificate must cover the actual historical average, not just a single source round.

\paragraph{Cross-configuration nuisance-training context mismatch.}
The main text assigns past rounds weights based on their context-law discrepancy from the target context law.
Let $\nu_{z,t}^{\mathrm{tr}}$ denote the population context marginal of that weighted training sample, and let $\nu_{z,0}$ be the target context law.
Let $\mathcal H_{\mathrm{cfg}}$ be the class of bounded context-level conditional score or nuisance-risk functions for which transfer is calibrated, normalized so that $|h(c^-)|\le B_\psi$.
Define
\begin{equation}
\bar\epsilon_{\mathrm{cfg},t}(z)
=
\sup_{h\in\mathcal H_{\mathrm{cfg}}}
\left|
\mathbb E_{C^-\sim\nu_{z,t}^{\mathrm{tr}}}h(C^-)
-
\mathbb E_{C^-\sim\nu_{z,0}}h(C^-)
\right|.
\label{eq:configuration-nuisance-mismatch}
\end{equation}
If a validated calibration map $K_{\mathrm{cfg}}$ controls this function class, then
\begin{equation}
\bar\epsilon_{\mathrm{cfg},t}(z)
\le
K_{\mathrm{cfg}}\!\left(
d_{\mathcal F}(\nu_{z,t}^{\mathrm{tr}},\nu_{z,0})
\right).
\label{eq:configuration-mismatch-calibration}
\end{equation}
The context-discrepancy weighting in the main text is an operational way to construct $\nu_{z,t}^{\mathrm{tr}}$; the empirical discrepancy itself is not a high-probability calibration unless the relevant function-class bound is justified.

This $\bar\epsilon_{\mathrm{cfg},t}(z)$ is a context-marginal nuisance- transfer allowance, matching the main-text definition and the finite- state context-audit construction in Appendix~\ref{app:v8-bias-growth}.
It does not, by itself, bound a mismatch between joint action/exposure input laws or between conditional response functions.
Therefore, $\epsilon_{m,t}(z)$ and $\epsilon_{\pi,t}(z)$ must be calibrated under the actual source-window score measures used in Appendix~\ref{app:nuisance-detailed}, or their transfer from the fitting measure must be justified separately.
In particular, a raw behavior propensity error alone is not a substitute for a bound on the induced clipped joint-ratio error.

\paragraph{Avoiding omissions and double counting.}
The context-shift term controls the change in expectation of the source-response functional between the source and target context laws.
The configuration term controls only the calibrated context-marginal transfer of nuisance-risk functions.
The exposure/response budget $\epsilon_{\mathrm{exp}}(z)$ controls the response/exposure contributions specified in Appendix~\ref{app:exposure-majorant} and the exact proof in Appendix~\ref{app:thm1proof-detailed}.
A source-to-target conditional response discrepancy must not be counted again as a configuration mismatch if it is already included in $\epsilon_{\mathrm{exp}}(z)$.
Conversely, an input-law or nuisance error not covered by the direct nuisance bounds cannot be silently omitted just because the context marginals are close.

The quantities above are population discrepancies or calibrated upper bounds, not automatic consequences of observed telemetry.
Empirical context distances, exposure-enrichment comparisons, and fitted-model diagnostics support sensitivity analysis but do not independently establish the required transport or coverage guarantees.

%% file: appendix/11_configuration_aware_nuisance_fitting.tex
\subsection{Configuration-Aware Nuisance Fitting}
\label{app:configuration-aware-nuisance-fitting}

Fix a candidate configuration $z$ and a predecision round $t$.
The nuisance functions used to evaluate $z$ target the no-exploration causal estimand, although their fitting sample may contain historical observations collected under other source configurations.
This section distinguishes target nuisance functions, the weighted training-context law, and the assumptions required to transport historical response information to the target configuration.

\paragraph{Target nuisance functions.}
Let $C^-$ denote the pre-action context and $\widetilde A=(A,E)$ the joint action/exposure treatment unit.
Write $g_{z,0}(\cdot\mid c^-)$ for the target joint law under configuration $z$ without measurement exploration.
Let $m_z(\widetilde a,c^-)$ be the target-regime conditional response mean and define
\[
\mu_z(c^-) = \int_{\widetilde{\mathcal A}} m_z(\widetilde a,c^-) g_{z,0}(d\widetilde a\mid c^-),
\qquad
W_\infty(z;0)=\mathbb E_{C^-\sim\nu_{z,0}}[\mu_z(C^-)].
\]
The fitted response regression and target joint law induce
\[
\widehat\mu_{z,r}(c^-) = \int_{\widetilde{\mathcal A}}
\widehat m_{z,r}(\widetilde a,c^-)
\widehat g_{z,0,r}(d\widetilde a\mid c^-).
\]
Here the subscript $r$ records the fitting cutoff used for the score stored for round $r$; the main text suppresses this cutoff inside $\widehat\psi_{z,r}$ for readability.
The fitted joint law must be a valid probability distribution.
For continuous joint treatments, all ratios and integrals use a common dominating measure.

\paragraph{Historical training observations and context weighting.}
Let $z_r$ and $\eta_r$ denote the deployed configuration and exploration rate in historical round $r$.
The scoring record is
\[
\widetilde O_{i,s}^{(r)}
= (\widetilde A_{i,s}^{(r)},C_{i,s}^{-,(r)},R_{i,s}^{(r)}).
\]
The corresponding behavior joint law is $g_{b,r}(\widetilde a\mid c^-)$ and its fitted version is $\widehat g_{b,r}(\widetilde a\mid c^-)$.

As specified in the main text, historical rounds receive fitting weights based on their context-law discrepancy from the target context law.
Let $\nu_{z,t}^{\mathrm{tr}}$ denote the population context marginal of the resulting weighted training sample.
The full training input law may also differ from the target joint input law
\[
\mu_{z,0}(dc^-,d\widetilde a) =\nu_{z,0}(dc^-)g_{z,0}(d\widetilde a\mid c^-).
\]
The context weights do not automatically identify or reconstruct the joint action/exposure law.
Thus the induced clipped-ratio error and the target-joint-law error must be calibrated separately, under the joint laws used by the actual score.

\paragraph{Predictable fitting and scoring.}
For the concentration arguments, the nuisance predictions used to score network window $j$ must be fixed using information available before that window is scored.
Accordingly, $\widehat\psi_{z,r}$ denotes the stored score rule for source round $r$: its target joint-law fit, response-regression fit, and behavior joint-law fit are frozen before the corresponding scoring window(s).
Historical scores are not retrospectively recomputed using fits that depend on later observations.
If the nuisance rule is updated within a round, the fitting cutoff and stored score must instead be indexed at the window level.
This is the predictability condition used in Appendix~\ref{app:v7-certification}.
Cross-fitting alone does not establish response transport, correct specification, or target-law overlap.

\paragraph{Target response versus source response.}
Let $m_{b,r}(\widetilde a,c^-)$ denote the conditional response mean under the historical source regime.
The response nuisance for candidate $z$ targets $m_z(\widetilde a,c^-)$, not automatically $m_{b,r}(\widetilde a,c^-)$.

When $z_r\ne z$, fitting a regression to historical responses does not by itself identify the candidate-specific response function.
A sufficient special case is conditional response invariance on the relevant support:
\[
m_{b,r}(\widetilde a,c^-)=m_z(\widetilde a,c^-).
\]
When this equality is not assumed, the source-to-target response discrepancy must be bounded separately.
It may be included in the calibrated exposure/response budget $\epsilon_{\mathrm{exp}}(z)$ only if that budget covers the same source regimes, target support, and historical averaging scheme.
It must not be silently absorbed into a nuisance-fitting error that measures estimation under the source response law only.

\paragraph{Nuisance errors and the meaning of the budgets.}
The response-nuisance error $\epsilon_{m,t}(z)$ must control $\|\widehat m_{z,r}-m_z\|_{L_2(\nu_r,g_{b,r})}$ under the score-law measures, uniformly over the relevant source rounds or through a separately justified historical-average bound.
The quantity $\epsilon_{\pi,t}(z)$ must control $\|\widehat w_{z,r}-\bar w_{z,r}\|_{L_2(\nu_r,g_{b,r})}$, where $\widehat w_{z,r}$ is the fitted clipped joint ratio and $\bar w_{z,r}=\min\{g_{z,0}/g_{b,r},w_{\max}\}$ is its clipped oracle counterpart.
A bound on the raw fitted behavior propensity alone is not sufficient unless it is shown to imply this induced-ratio bound.
The target-joint-law error is first order and is accounted for by $\bar\epsilon_{\mathrm{target},t}(z)$.

The context-shift term $\bar\epsilon_{\mathrm{shift},t}(z)$ controls score averaging across source and target context laws.
The configuration term $\bar\epsilon_{\mathrm{cfg},t}(z)$ controls only the calibrated context-marginal nuisance transfer described in Appendix~\ref{app:context-shift-detailed}; it does not, by itself, control joint-treatment-law mismatch or conditional-response transport.
The nuisance-product, target-joint-law, clipping, context, and response terms therefore have distinct roles.
A response discrepancy already included in $\epsilon_{\mathrm{exp}}(z)$ must not be counted again as a configuration mismatch.

\paragraph{Scope of the fitting guarantees.}
Configuration-aware fitting is a valid certificate component only when target-law, response-nuisance, induced-ratio, context-transport, and response-transport bounds are independently justified.
Fit quality or small observed discrepancies alone do not guarantee target-law accuracy; the result remains conditional on Appendix~\ref{app:dr-identity-detailed}.

%% file: appendix/15_finite_sample_conditions_for_causal_value_certification.tex
\subsection{Finite-Sample Conditions for Causal Value Certification}
\label{app:v7-certification}

This appendix supplies auditable mathematical sufficient conditions for the sampling component of Theorem~\ref{thm:main} and for its use in Theorem~\ref{thm:regret}.
Its statements are conditional on the identification, overlap, nuisance, and transport requirements in Appendices
\ref{app:nuisance-detailed}--\ref{app:thm1proof-detailed}.
The target treatment unit is the joint action/exposure $\widetilde A=(A,E)$, and the context used for causal scoring is the pre-action context $C^-$.
The resulting confidence certificate combines the concentration results below with the independently calibrated target-transport bounds in Appendix~\ref{app:context-shift-detailed}.

\paragraph{Network-window unit and predictable scoring.}
Index the available network windows in observation order by $j=1,\ldots,n$.
For candidate configuration $z$, define
\[
Y_{j,z} = \frac1N\sum_{i=1}^N \widehat\psi_{z,j}(\widetilde O_{i,j}),
\qquad
\widetilde O_{i,j}
= (\widetilde A_{i,j},C^-_{i,j},R_{i,j}).
\]
Here $\widetilde A_{i,j}$ is the observed joint action/exposure treatment unit and $C^-_{i,j}$ contains only information available strictly before the focal action.

The nuisance fits, fitted behavior joint law, and candidate scoring function used at window $j$ are measurable with respect to a pre-window sigma field $\mathcal F_{j-1}$.
A strictly forward, block-separated nuisance-training protocol satisfies this predictable-fit condition and preserves the temporal ordering required by the concentration argument.
The within-window dependence among nodes is otherwise unrestricted by the martingale statement below.
By the clipped-score range, $|Y_{j,z}|\le B_\psi$.
The network-window count $n$, rather than the total number $Nn$ of dependent node-level records, is the unit in the following concentration arguments.

\begin{lemma}[Finite-sample martingale route]
\label{lem:v7-azuma}
Fix a candidate $z$ and assume the predictable-fit condition above.
For every $\alpha\in(0,1)$,
\begin{equation}
\mathbb P\!\left(
\left|
\frac1n\sum_{j=1}^n
\left[
Y_{j,z}
-\mathbb E(Y_{j,z}\mid\mathcal F_{j-1})
\right]
\right|
>
2B_\psi\sqrt{\frac{2\log(2/\alpha)}{n}}
\right)
\le\alpha.
\label{eq:v7-azuma}
\end{equation}
No independence or mixing assumption across windows is needed for this centered martingale statement.
It does not control the predictable-mean drift.
\end{lemma}

\begin{proof}
The summands are martingale differences bounded in absolute value by $2B_\psi$.
Azuma--Hoeffding and a two-sided union bound give
\[
2\exp\!\left( -\frac{x^2}{8nB_\psi^2} \right)
\]
as an upper bound on the probability that the absolute centered sum exceeds $x$.
Setting
\[
x=2B_\psi\sqrt{2n\log(2/\alpha)}
\]
and dividing by $n$ yields Eq.~\eqref{eq:v7-azuma}.
\end{proof}

\paragraph{Calibrating predictable bias and stochastic fluctuation.}
For each window, condition on $\mathcal F_{j-1}$ and use the conditional behavior and context laws in the exact nuisance and transport decomposition of Appendix~\ref{app:nuisance-detailed}.
Let
\[
\Delta^{\mathrm{pred}}_{j,z} =
\mathbb E(Y_{j,z}\mid\mathcal F_{j-1})
-W_\infty(z;0)
\]
denote the total signed predictable-mean discrepancy, distinct from the predictable-window drift term in the second line of Eq.~\eqref{eq:repaired-proof-telescope}.
The following is a sufficient pathwise budget requirement on an event $\mathcal G$:
\begin{equation}
\left|
\frac1n\sum_{j=1}^n \Delta^{\mathrm{pred}}_{j,z}
\right|
\le
\epsilon_{m,t}(z)\epsilon_{\pi,t}(z)
+\bar\epsilon_{\mathrm{target},t}(z)
+\epsilon_{\mathrm{clip},t}(z)
+\epsilon_{\mathrm{exp}}(z)
+\bar\epsilon_{\mathrm{shift},t}(z)
+\bar\epsilon_{\mathrm{cfg},t}(z)
+\widehat\kappa_t(z).
\label{eq:v7-predictable-bias}
\end{equation}
For this inequality to hold, each budget on the right-hand side must be calibrated for the relevant joint action/exposure laws, source regimes, and historical-window averaging scheme.

In particular, $\epsilon_{m,t}(z)$ and $\epsilon_{\pi,t}(z)$ must control, respectively, $\|\widehat m_{z,r}-m_z\|_{L_2(\nu_r,g_{b,r})}$ and $\|\widehat w_{z,r}-\bar w_{z,r}\|_{L_2(\nu_r,g_{b,r})}$ for the source-window score laws, uniformly over the source rounds in the prefix or through a separately justified bound on the averaged products.
The latter is an induced clipped joint-ratio error; a raw behavior-propensity estimation error is sufficient only if a valid inequality transfers it to this ratio norm.
The clipping allowance must cover the clipping contributions appearing in the actual proof decomposition, and the target-joint-law term must control the first-order error in the fitted target joint distribution.

The main-text exposure/response budget $\epsilon_{\mathrm{exp}}(z)$ must dominate the historical average of both the score-level response residual in Eq.~\eqref{eq:score-level-response-transport} and the target-context response discrepancy in Eq.~\eqref{eq:exposure-majorant}, as well as any exposure-representation error not already included in those quantities.
This is a calibration requirement for the actual stored scoring rules and admissible source regimes; the target-context bound in Eq.~\eqref{eq:exposure-majorant} alone does not establish it.
A uniform bound over source regimes and prefixes is sufficient, while a direct bound on the combined contribution in the exact telescope may be tighter.
These component bounds are folded into the single $\epsilon_{\mathrm{exp}}(z)$ term and must not be added a second time as separate terms in $\bar\epsilon_{\mathrm{tot},t}(z)$.

The context-shift allowance must apply to the source and target context laws and the source-response functional actually used in the historical average.
The configuration-mismatch allowance is the calibrated context-marginal nuisance-transfer term defined in Appendix~\ref{app:context-shift-detailed}.
It does not, by itself, control joint action/exposure input-law mismatch or conditional response transport.
The direct nuisance bounds must therefore be calibrated under the actual score-law measures, or any further transfer must be separately justified.
A response discrepancy already included in $\epsilon_{\mathrm{exp}}(z)$ must not be counted again as a configuration-mismatch term.

Equation~\eqref{eq:v7-predictable-bias} is a sufficient calibration requirement, not a bound inferred from stationarity or an empirical block-drift statistic.
It must be justified using the target-law, clipping, nuisance, exposure/response-transport, and context-law conditions in Appendices~\ref{app:nuisance-detailed}--\ref{app:thm1proof-detailed}.
One sufficient route is to bound the window-averaged contributions of each of the four non-statistical terms in Eq.~\eqref{eq:repaired-proof-telescope}.

\begin{proposition}[Concrete conditional simultaneous certificate]
\label{prop:v7-finite}
Suppose $\mathcal Z$ is finite with $M=|\mathcal Z|$, and $\mathbb P(\mathcal G)\ge1-\delta/2$, where on $\mathcal G$ Eq.~\eqref{eq:v7-predictable-bias} holds simultaneously for every $z\in\mathcal Z$.
Suppose the estimators use $n$ network windows and the predictable-fit condition holds for every candidate.
Define
\[
\operatorname{comp}(\mathcal Z)=\log M
\]
and suppose the certified statistical effective information count satisfies
\begin{equation}
0<\underline n_{\mathrm{eff},t}(z)\le n/4,
\qquad z\in\mathcal Z.
\label{eq:v7-martingale-neff}
\end{equation}
Then the bound in Theorem~\ref{thm:main}, with the same stated $2B_\psi$ statistical coefficient, holds simultaneously for every $z\in\mathcal Z$ with probability at least $1-\delta$.
\end{proposition}

\begin{proof}
Apply Lemma~\ref{lem:v7-azuma} with $\alpha=\delta/(2M)$ and take a union bound over the $M$ candidates.
The resulting sampling failure probability is at most $\delta/2$, and the stochastic radius for each candidate is
\[
2B_\psi
\sqrt{\frac{2\log(4M/\delta)}{n}}.
\]
Write
\[
L=\log M+\log(2/\delta).
\]
Since $L\ge\log 2$,
\[
\log(4M/\delta)=L+\log 2\le 2L.
\]
Consequently,
\[
2B_\psi
\sqrt{\frac{2\log(4M/\delta)}{n}}
\le
2B_\psi\sqrt{\frac{L}{n/4}}
\le
2B_\psi
\sqrt{\frac{L}{\underline n_{\mathrm{eff},t}(z)}}.
\]
The last inequality follows from $\underline n_{\mathrm{eff},t}(z)\le n/4$.
Intersect this simultaneous sampling event with $\mathcal G$ and apply Eq.~\eqref{eq:v7-predictable-bias}.
The total failure probability is at most $\delta$.
\end{proof}

\paragraph{Alternative mixing/block-coupling route.}
When nuisance scoring is fixed separately from a stationary evaluation stream and a conditional absolute-regularity bound is supplied, independent-block coupling provides an alternative finite-sample certificate.
This route requires its own dependence assumptions and does not follow merely from the martingale result above.

\begin{lemma}[Block-coupled sampling route]
\label{lem:v7-block}
Let $Y_1,\ldots,Y_n$ be a bounded sequence with $|Y_j|\le B_\psi$, stationary conditional on a separate fitting sigma field, and with absolute-regularity coefficient $\beta(k)$.
Choose integers $q,m\ge1$ such that $n=2mq+r$, $0\le r<2q$, and suppose $2m\beta(q)\le\alpha/2$.
Then, conditional on the fitting field, with probability at least $1-\alpha$,
\begin{equation}
\left|
\frac1n\sum_{j=1}^nY_j-\mathbb EY_1
\right|
\le
B_\psi\sqrt{\frac{2\log(8/\alpha)}{m}}
+\frac{4B_\psi q}{n}.
\label{eq:v7-block}
\end{equation}
In particular, if $\beta(q)\le C_\beta\exp(-q/\tau)$ with known constants and $q$ is chosen to satisfy the coupling condition, the right-hand side is fully specified.
One sufficient choice is
\[
q\ge\tau\log(4C_\beta n/\alpha),
\]
provided $n\ge2q$.
\end{lemma}

\begin{proof}
Partition the first $2mq$ observations into $2m$ successive blocks of length $q$.
Form the $m$ odd-indexed and $m$ even-indexed block averages.
Blocks within each parity group are separated by at least $q$ observations.
The absolute-regularity coupling bound controls the probability that either group differs from its independent block coupling by at most $2m\beta(q)\le\alpha/2$.

Conditional on successful coupling, each parity group consists of $m$ independent block averages bounded in absolute value by $B_\psi$.
Hoeffding's inequality and a union bound over the two groups give the stated stochastic radius with additional failure probability at most $\alpha/2$.
The $r$ omitted observations change the full sample mean relative to the mean of the first $2mq$ observations by at most
\[
\frac{2B_\psi r}{n}
\le
\frac{4B_\psi q}{n}.
\]
Combining these bounds proves Eq.~\eqref{eq:v7-block}.
\end{proof}

\paragraph{Configuration covers and complete failure accounting.}
For infinite compact $\mathcal Z$, a finite $\varepsilon$-net can be used only if the evaluated estimator and the target value are uniformly $L_Z$-Lipschitz on the relevant confidence event.
The lift from net points requires a $2L_Z\varepsilon$ additive remainder, which must be included in the stated approximation/error budget.
The local $1-\delta$ conclusion above is for one evaluation time.
For the adaptive online algorithm, choose positive $\delta_t$ satisfying
\[
\sum_{t=2}^{T}\delta_t\le\delta
\]
and allocate each $\delta_t$ among sampling, nuisance/transport calibration, and any covering event.
A uniform-in-round guarantee does not follow from repeating a fixed-round confidence interval without this allocation.

\paragraph{Effective information and drift calibration.}
The operational shrinkage proxy
\eqref{eq:app-neff-shrinkage} is diagnostic only. The theorem
uses a validated lower-information bound, such as Eq.~\eqref{eq:v7-martingale-neff} or a calibrated block-coupling counterpart, and not the proxy by itself.
The adjacent-block statistic \eqref{eq:app-kappa} is similarly usable as a drift allowance only when validated against the predictable-bias condition Eq.~\eqref{eq:v7-predictable-bias}.
These distinct inputs capture complementary features of post-burn-in data: effective information and residual predictable-mean drift.
Independent calibration, model-dependent deviation bounds, or conservative deterministic bounds provide sufficient conditions.
The slow-mixing family in Appendix~\ref{app:frontier-detailed} clarifies their role.

\paragraph{Limited-overlap local-action calibration example.}
Under a correct known local-action mixture
\[
b_A=(1-\eta)p_A+\eta\xi,
\qquad \eta<1,
\]
the target local-action law can be recovered as
\[
p_A=\frac{b_A-\eta\xi}{1-\eta},
\]
provided the equality defines a valid probability vector.
A fitted version obeys
\[
\|\widehat p_A-p_A\|_1 =
\frac{1}{1-\eta}
\|\widehat b_A-b_A\|_1
\]
when the fitted quantities satisfy the same mixture relation.
Thus an independently validated confidence bound on the local-action probabilities yields a corresponding local-action error bound.

This inversion does not, by itself, recover the target joint action/exposure law $g_{z,0}$ from the marginal local-action law.
A joint-law reconstruction requires the actual joint action-generation mechanism, including dependencies among contemporaneous actions and the exposure mapping.
At $\eta=1$, the mixture inversion does not identify the target local-action law $p_A$.
Moreover, even a known target local-action law does not identify the counterfactual context distribution $\nu_{z,0}$ from arbitrary exploratory logs.
Calibrated joint-treatment, exposure, and context- transport conditions therefore remain separate requirements.

%% file: appendix/16_calibrated_error_budgets_and_information_growth.tex
\subsection{Calibrated Error Budgets and Information Growth}
\label{app:v8-bias-growth}
This appendix develops two components of the causal decision guarantee: finite-sample calibration of target-value bias terms and growth of certified information along the selected configuration sequence.
Independent audit observations provide explicit upper bounds for the identification and transport budgets; per-visit information conditions then yield the square-root summability used by Theorem~\ref{thm:regret}.
The construction gives operational interpretations to the sufficient conditions in Theorems~\ref{thm:main} and~\ref{thm:regret}.

\subsubsection{Target identification and observational equivalence}
\begin{proposition}[Observational indistinguishability]
\label{prop:v8-nonid}
Without identifying restrictions or independent information about the no-exploration action law, the target-policy bias cannot have a uniformly vanishing, data-dependent high-probability bound based only on exploratory logs.
The same conclusion applies to the full target certificate.
\end{proposition}
\begin{proof}
Take one context, two actions $A\in\{0,1\}$, and deterministic reward $R=A$.
Suppose the only logged regime is $\eta=1$ with known exploration law $\xi(1)=1/2$.
Consider two admissible target laws $p_+(1)=1$ and $p_-(1)=0$.
In both environments the observed law is $A\sim\mathrm{Ber}(1/2)$ and $R=A$, independently of the target law.
Yet the no-exploration values are $W_+=1$ and $W_-=0$.
Under the common law of the observed data, any interval $I(D)$ having coverage at least $1-\delta$ for both targets must contain both $0$ and $1$ simultaneously with probability at least $1-2\delta$; on that event its width is at least $1$.
Thus no shrinking uniformly valid target certificate can be deduced from these logs.
The construction is also an exact failure case for inversion of the mixture $b=(1-\eta)p+\eta\xi$ at $\eta=1$.
\end{proof}
No statement here rules out identification when $\eta<1$, the known mixture model is correct, and sufficiently informative observations are available.
It rules out treating fitted target policies, enriched-exposure sensitivity, or empirical IPMs as automatically certified upper bounds.

A separate context-shift obstruction can persist even for a fixed $0<\eta<1$ if no structural relation between the context laws $\nu_{z,\eta}$ and $\nu_{z,0}$ is assumed: take a logging context $C\equiv0$ and reward $R\equiv0$ in two systems at that positive exploration rate, but let the unobserved no-exploration regime have $C\equiv1$ and deterministic target reward $\theta\in\{0,1\}$.
The two logging laws coincide and the two target values differ by one.
This counterexample is deliberately excluded when a verified transport condition links the regimes; its purpose is to show why that condition cannot be inferred from logged outcomes alone.

\subsubsection{Explicit componentwise calibration under a finite-state audit}
\label{app:v8-finite-audit}
We give one constructive, deliberately strong calibration protocol.
Let $\mathcal Z$, $\mathcal C^-$, and the joint treatment/exposure space $\widetilde{\mathcal A}$ be finite, with cardinalities $M$, $J$, and $K$, respectively.
Each $\widetilde a\in\widetilde{\mathcal A}$ is a full joint action/exposure cell, not merely the focal local action.
Let $R_{\mathrm{src}}$ be the number of source-regime types covered by the audit.
Assume $|R|\le B_u$.
Work conditionally on the data used to fit $\widehat g_{z,0,r}$, $\widehat g_{b,r}$, and $\widehat m_{z,r}$; all audit data below are independent of those fitting data.
For every candidate $z$, source regime $r$, context $c^-$, and joint cell $\widetilde a$, assume audit sampling can identify the target interventional response $m_z^0(\widetilde a,c^-)$ and the source-regime response $m_{b,r}(\widetilde a,c^-)$.
It also identifies the target joint law $g_{z,0}(\cdot\mid c^-)$ and source behavior joint law $g_{b,r}(\cdot\mid c^-)$.
These are substantive identification assumptions; observing an exposure feature alone does not supply them.
For the displayed nuisance-error formula, assume that the target response $m_z$ used by the score equals the identified audit response $m_z^0$ on the relevant support.
If it does not, the response-nuisance error must still be calibrated against the actual $m_z$, and the representation discrepancy must be separately included in the exposure/response budget.
Independent context-only audit samples identify the target context law $\nu_{z,0}$, each source context law $\nu_{z_r,\eta_r}$ in the audited source set, and the weighted nuisance-training context law $\nu_{z,t}^{\mathrm{tr}}$.

Assume every audited source joint cell on the target support satisfies $g_{b,r}(\widetilde a\mid c^-)\ge\beta>0$.
All bounds below are uniform over the audited source-regime set; for a pooled estimator, that set must include every source regime used in the corresponding historical prefix.
Let $n_{\min}\ge1$ be the minimum independent sample count per audited scalar stratum.
With
\begin{equation}
D=M R_{\mathrm{src}}(4JK+3J),\qquad
r_p=\sqrt{\frac{\log(2D/\alpha)}{2n_{\min}}},\qquad
r_m=B_u\sqrt{\frac{2\log(2D/\alpha)}{n_{\min}}},
\label{eq:v8-audit-radii}
\end{equation}
each audited joint-cell probability or context probability is within $r_p$ of its true value and each conditional response mean is within $r_m$, jointly with probability at least $1-\alpha$, by Hoeffding and a union bound over at most $D$ scalar estimates.
If audit observations are dependent, these radii must be replaced by independently justified block or dependence-aware bounds.

On this audit event, define the computable majorants
\begin{align}
e_g(z)&=\max_{r,c^-}\sum_{\widetilde a}
\left(|\widehat g_{z,0,r}(\widetilde a\mid c^-)
-\widetilde g_{z,0}(\widetilde a\mid c^-)|+r_p\right),\notag\\
e_b&=\max_{r,c^-,\widetilde a}
\left(|\widehat g_{b,r}(\widetilde a\mid c^-)
-\widetilde g_{b,r}(\widetilde a\mid c^-)|+r_p\right),\notag\\
e_m(z)&=\max_{r,c^-,\widetilde a}
\left(|\widehat m_{z,r}(\widetilde a,c^-)
-\widetilde m_z^0(\widetilde a,c^-)|+r_m\right),\notag\\
e_\nu(z;v,w)&=\|\widetilde\nu_{z,v}-\widetilde\nu_{z,w}\|_1+2Jr_p.
\label{eq:v8-majorants}
\end{align}
Here $\widetilde g$, $\widetilde m$, and $\widetilde\nu$ are independent audit estimates.
For context shift across source configurations, also set
\[
e_\nu^{\mathrm{shift}}(z) =\max_r\left(\|\widetilde\nu_{z_r,\eta_r} -\widetilde\nu_{z,0}\|_1+2Jr_p\right),
\]
where the maximum is over the audited source regimes in the prefix.
Clip fitted response regressions to $[-B_u,B_u]$.
If $e_b\le\beta/2$ and $\pi_{\min}\le\beta/2$, the fitted denominator is at least $\beta/2$, and the clipped-ratio error obeys
\begin{equation}
\|\widehat w_{z,r}-\bar w_{z,r}\|_\infty
\le\frac{2e_g(z)}{\beta}+\frac{2e_b}{\beta^2}
=:e_w(z),
\label{eq:v8-weight-error}
\end{equation}
where $\bar w_{z,r}=\min\{g_{z,0}/g_{b,r},w_{\max}\}$.
This follows by adding and subtracting $g_{z,0}/\widehat g_{b,r}$, using the lower bounds on the true and fitted denominators, and applying the 1-Lipschitz clipping map.

The following bounds then hold simultaneously on the audit event:
\begin{align}
\bar\epsilon_{\mathrm{target}}(z)&=B_u e_g(z),\notag\\
\bar\epsilon_{m\pi}(z)&=e_m(z)e_w(z),\notag\\
\bar\epsilon_{\mathrm{clip}}(z)&=2B_u\max_{r,c^-,\widetilde a}
\left[\frac{\min\{1,\widehat g_{z,0,r}(\widetilde a\mid c^-)+e_g(z)\}}
{\beta}-w_{\max}\right]_+,\notag\\
\bar\epsilon_{\mathrm{shift}}(z)&=B_u e_\nu^{\mathrm{shift}}(z),\notag\\
\bar\epsilon_{\mathrm{cfg}}(z)&=B_\psi e_\nu(z;\mathrm{tr},0).
\label{eq:v8-explicit-components}
\end{align}
Here $[x]_+=\max\{x,0\}$, and $\mathrm{src}(r)$ denotes the context law of source regime $r$.
The clipping bound is conservative because $g_{z,0}/g_{b,r}\le(\widehat g_{z,0}+e_g)/\beta$ on the audited cells.
The shift and configuration bounds use $|\mathbb E_\nu f-\mathbb E_{\nu'}f| \le\|f\|_\infty\|\nu-\nu'\|_1$ for the respective bounded context-level function classes.
The configuration term is a context-marginal nuisance-transfer allowance; it does not replace the joint-law or response-nuisance calibrations above.

\paragraph{Joint response-transport audit.}
To cover the exact proof telescope, audit both the target-context response discrepancy and the score-level response residual.
Define
\begin{align}
\widetilde E_{\mathrm{target}}(z)
&=\max_{r,c^-}\sum_{\widetilde a}
\widehat g_{z,0,r}(\widetilde a\mid c^-)
\left|\widetilde m_z^0(\widetilde a,c^-)
-\widetilde m_{b,r}(\widetilde a,c^-)\right|,\notag\\
\widetilde E_{\mathrm{score}}(z)
&=\max_{r,c^-}\left|
\sum_{\widetilde a}
\left[\widetilde g_{b,r}(\widetilde a\mid c^-)
\widehat w_{z,r}(\widetilde a,c^-)
-\widetilde g_{z,0}(\widetilde a\mid c^-)\right]
\left[\widetilde m_{b,r}(\widetilde a,c^-)
-\widetilde m_z^0(\widetilde a,c^-)\right]
\right|.
\label{eq:v8-response-audit-estimates}
\end{align}
The target-law discrepancy is bounded by $\widetilde E_{\mathrm{target}}(z)+2B_u e_g(z)+2r_m$.
The score-level residual is bounded by
\begin{align}
\bar E_{\mathrm{score}}(z)
={}&\widetilde E_{\mathrm{score}}(z)
+2B_u\{w_{\max}K r_p+K r_p\}
+2r_m(w_{\max}+1).
\label{eq:v8-score-response-bound}
\end{align}
Indeed, the audited behavior and target joint probabilities have coordinatewise error at most $r_p$, so the $\ell_1$ coefficient error in $g_{b,r}\widehat w-g_{z,0}$ is at most $w_{\max}K r_p+K r_p$.
The conditional response-difference error is at most $2r_m$, and the $\ell_1$ norm of the true coefficient vector is at most $w_{\max}+1$.

Consequently, a sufficient main-text exposure/response budget for this finite-state audit is
\begin{equation}
\bar\epsilon_{\mathrm{exp}}(z)
=
\widetilde E_{\mathrm{target}}(z)+2B_u e_g(z)+2r_m
+\bar E_{\mathrm{score}}(z).
\label{eq:v8-exposure-calibrated}
\end{equation}
This budget covers both response terms required by Appendix~\ref{app:thm1proof-detailed}; it is not merely a bound on the target-context response difference.
If the exposure representation itself has an additional approximation error not captured by the interventional response audit, that error must also be added to $\bar\epsilon_{\mathrm{exp}}(z)$.
The target-response audit requires appropriate interventional identification; comparing two coarsened observational exposure regressors does not produce this bound.
When a required audit cell or source regime is unavailable, the corresponding component is uncertified, not zero.

Finally, assume the expected bounded score in each evaluation window is at total variation distance at most $d_{j,z}$ from the corresponding reference score law, where $d_{j,z}$ has a known simultaneous upper bound.
Then
\begin{equation}
\bar\kappa_t(z)=\frac{2B_\psi}{n}\sum_{j=1}^n d_{j,z}
\label{eq:v8-drift-calibration}
\end{equation}
is a valid drift allowance.
For example, a proved uniform post-burn-in total-variation contraction $d_{j,z}\le C_{\mathrm{mix}}e^{-(T_{\mathrm{burn}}+s_j)/\tau}$ provides explicit $d_{j,z}$.
An adjacent-block empirical difference alone does not imply this contraction or certify its constants.

\begin{proposition}[Finite-audit value certificate]
\label{prop:v8-calibration}
Suppose all finite-state audit and conditional-response identification assumptions above hold, the predictable network-window scoring condition of Lemma~\ref{lem:v7-azuma} holds, and the target value has the decomposition in Eq.~\eqref{eq:repaired-proof-telescope}.
Replace the bias budgets in Theorem~\ref{thm:main} by the componentwise upper bounds in Eqs.~\eqref{eq:v8-explicit-components},
\eqref{eq:v8-exposure-calibrated}, and
\eqref{eq:v8-drift-calibration}. Set the audit level to $\alpha=\delta/2$
and allocate the remaining $\delta/2$ to the centered-score event.
Then the resulting simultaneous value certificate holds with probability at least $1-\delta$ for all $M$ configurations at that evaluation time.
For an adaptive run, the failure allocation must be repeated with roundwise levels whose sum is at most the global $\delta$.
\end{proposition}
\begin{proof}
The audit concentration event and the majorants in Eqs.~\eqref{eq:v8-weight-error}--\eqref{eq:v8-exposure-calibrated} control the target joint-law, response nuisance, induced ratio, clipping, context-shift, configuration, and score-level response-transport contributions simultaneously.
The stated total-variation envelope bounds the remaining predictable post-burn-in drift by Eq.~\eqref{eq:v8-drift-calibration}.
Substituting these bounds into the exact telescope in Eq.~\eqref{eq:repaired-proof-telescope} verifies the pathwise bias condition in Eq.~\eqref{eq:v7-predictable-bias}.
The centered-score concentration result then supplies the sampling part of the certificate.
For a full adaptive run, allocate the failure budget across rounds so that the total remains at most $\delta$.
\end{proof}

\paragraph{Calibration interfaces and implementation.}
This finite-state construction is deliberately strong: it requires independent audits of the full joint action/exposure cells, source-regime responses, target responses, and context laws.
It does not derive a joint propensity from marginal local-action propensities.
If these audits or structural restrictions are unavailable, the corresponding error budget is uncertified rather than zero.

\subsubsection{Selected-configuration information growth}
\label{app:v8-ess-growth}
Let $N_z(t-1)$ count qualifying post-initialization visits to $z$ strictly before decision $t$.
Merely defining an empirical or conservative-looking $\underline n_{\rm eff,t-1}(z)$ does not force it to grow on such visits.
\begin{lemma}[Per-visit lower information bound]
\label{lem:v8-selected-count}
Suppose there are $M<\infty$ configurations.
After at most $H$ initialization decisions, every selected configuration receives a predictable, valid score, and its certified predecision information count satisfies
\begin{equation}
\underline n_{\rm eff,t-1}(z_t)\ge
\lambda\bigl(1+N_{z_t}(t-1)\bigr),
\qquad \lambda>0,
\label{eq:v8-info-per-visit}
\end{equation}
for each noninitialization decision $t$.
Then for any adaptive sequence of configuration selections,
\begin{equation}
\sum_{t\text{ noninitialization}}
\frac1{\sqrt{\underline n_{\rm eff,t-1}(z_t)}}
\le 2\sqrt{\frac{MT}{\lambda}}.
\label{eq:v8-counting}
\end{equation}
In particular, when each eligible visit contributes at least $s\ge1$ new nonreused network windows and the martingale information certificate is constructed as $\underline n_{\rm eff}=n/4$, the condition follows with $\lambda=s/4$ after providing one certified pilot score per configuration and preserving all subsequent evaluated windows.
\end{lemma}
\begin{proof}
For a given configuration, index its qualifying selections by $k=1,\dots,N_z(T)$.
By Eq.~\eqref{eq:v8-info-per-visit}, its $k$th contribution is at most $1/\sqrt{\lambda k}$.
Thus
\[
\sum_{k=1}^{N_z(T)}\frac1{\sqrt{\lambda k}} \le\frac{2}{\sqrt\lambda}\sqrt{N_z(T)}.
\]
Summing over configurations and applying Cauchy--Schwarz yields $\sum_z\sqrt{N_z(T)}\le\sqrt{M\sum_zN_z(T)}\le\sqrt{MT}$.
This proof permits fully adaptive configuration selection; it does not require a uniform lower bound on the number of times each arm is chosen.
\end{proof}

\begin{corollary}[An explicit conditional regret bound]
\label{cor:v8-regret}
Under the simultaneous certified-value event of Theorem~\ref{thm:main} and the information-growth assumption
\eqref{eq:v8-info-per-visit}, write
$L_T=\operatorname{comp}(\mathcal Z)+\log(2T/\delta)$ and $E_t(z)=\bar\epsilon_{\rm tot,t}(z)+\widehat\kappa_t(z)$.
Then
\begin{equation}
\operatorname{Reg}_T\le
2B_u H+8B_\psi\sqrt{\frac{MTL_T}{\lambda}}
+2\sum_{t\text{ noninitialization}}E_{t-1}(z_t).
\label{eq:v8-regret-explicit}
\end{equation}
For fixed $M$ and $\lambda$, the stochastic term is $\widetilde O(\sqrt T)$.
When $M$ varies, the dependence $\sqrt M$ cannot be dropped from this counting argument.
\end{corollary}
\begin{proof}
Every initial decision loses at most $2B_u$.
On the joint confidence event, optimism as in Appendix~\ref{app:regret-detailed} gives instantaneous regret at most twice the predecision radius.
For every noninitialization decision this is at most $4B_\psi\sqrt{L_T/\underline n_{\rm eff,t-1}(z_t)} +2E_{t-1}(z_t)$.
Apply Lemma~\ref{lem:v8-selected-count} and add the initialization regret.
\end{proof}

\begin{proposition}[Growth does not follow from OFU alone]
\label{prop:v8-no-universal-growth}
Algorithm~\ref{alg:causalucb}, without a condition ensuring at least one newly usable, certified scoring window per eligible deployment, does not imply Eq.~\eqref{eq:v8-info-per-visit} or its summability conclusion.
\end{proposition}
\begin{proof}
Consider a configuration repeatedly deployed under a law that never provides the support required to fit its target-policy or response nuisance.
Its post-burn-in window count grows, but its number of \emph{certifiable} evaluation windows may remain zero.
More generally, $\underline n_{\rm eff,t}(z)\equiv1$ is a conservative lower-information number for an increasing amount of independent data, and nothing in the original OFU selection rule rules out retaining that constant certificate.
If such an arm is repeatedly chosen, the predecision reciprocal-square-root sum is $\Theta(T)$, not $\widetilde O(\sqrt T)$.
OFU controls selection given confidence radii; it does not create valid statistical information by itself.
\end{proof}
\paragraph{Implementation of information-growth checks.}
The implementation records scored network windows and predecision information counts.
The quantity $n/4$ serves as a sampling-information proxy; to apply the counting lemma, each eligible visit is checked for positive score production and the certified lower-information condition.
This diagnostic connects the observed evaluation stream with the information-growth condition used in the regret analysis.

%% file: appendix/29_identification_calibration_scope_and_growth.tex
\subsection{Identification, Calibration Sources, and Information-Growth Scope}
\label{app:certificate-scope-clarification}

This appendix collects the interpretation of assumptions already used in Theorems~\ref{thm:main}--\ref{thm:regret}.
It is a scope and implementation clarification, not an additional general certification theorem.
The finite-state independent-audit construction and the per-visit counting argument already appear in the existing technical appendices; neither should be confused with a guarantee based solely on uncalibrated exploratory telemetry.

\subsubsection{Why correcting the local-action propensity is insufficient}
Fix $z$ and an exploratory regime $\eta>0$.
Write $b_{z,\eta}(a\mid c^-)$ for the observed conditional law of the focal local action and $p_z(a\mid c^-)=\pi_{z,0}(a\mid c^-)$ for its target marginal law.
Under the stated known local-action mixture,
\[
b_{z,\eta}(a\mid c^-) =(1-\eta)p_z(a\mid c^-)+\eta\xi(a\mid c^-).
\]
When $\eta<1$, $\xi$ is known, and the mixture identity is valid at the conditioned contexts, the marginal action law can be recovered by
\[
p_z(a\mid c^-) =\frac{b_{z,\eta}(a\mid c^-)-\eta\xi(a\mid c^-)}{1-\eta}.
\]
Estimation error in $b_{z,\eta}$ is amplified by $1/(1-\eta)$, and the identity becomes nonidentifying at $\eta=1$.
This inversion does not recover the no-exploration steady-state context law $\nu_{z,0}$.
More importantly for the main theorem, it does not in general recover the target joint action/exposure law $g_{z,0}(\widetilde a\mid c^-)$.
The latter depends on the actual joint action-generation mechanism, including dependencies among contemporaneous neighboring actions and the exposure mapping.
It must be identified or calibrated from that mechanism or from an independent joint-cell audit; it cannot be inferred from the marginal local-action mixture alone.

To display the two transport issues in the general notation, let $m_{b,r}(\widetilde a,c^-)$ denote the conditional response under a historical source regime and $m_z(\widetilde a,c^-)$ the target-regime response.
Define the joint-action-corrected source value
\[
V_{z,r}^{\mathrm{joint\ corrected}} :=
\mathbb E_{C^-\sim\nu_{z_r,\eta_r}}
\int_{\widetilde{\mathcal A}} g_{z,0}(d\widetilde a\mid C^-) m_{b,r}(\widetilde a,C^-).
\]
Even if joint-treatment weighting is exact, the source and target responses may differ.
For bounded $|m_{b,r}|,|m_z|\le B_u$, a useful upper bound is
\begin{align}
\left|
V_{z,r}^{\mathrm{joint\ corrected}}-W_\infty(z;0)
\right|
\le{}&
B_u\left\|\nu_{z_r,\eta_r}-\nu_{z,0}\right\|_1
\notag\\
&+
\mathbb E_{C^-\sim\nu_{z,0}}
\int_{\widetilde{\mathcal A}}
 g_{z,0}(d\widetilde a\mid C^-)
 |m_{b,r}(\widetilde a,C^-)-m_z(\widetilde a,C^-)|,
\label{eq:supp-action-vs-context}
\end{align}
where $\|\cdot\|_1$ denotes the $\ell_1$ distance between context probability vectors.
The first term is context transport, and the second is conditional-response transport.
Both require an identifying restriction or an independently supported bound; neither follows from improved marginal action overlap.

\subsubsection{Data sources and audit cost for Theorem~\ref{thm:main}}
A valid application of the general certificate must supply joint, predecision error bounds, rather than substitute an empirical proxy for a high-probability upper bound.
Examples of sufficient sources are:
\begin{center}
\small
\begin{tabularx}{\linewidth}{@{}p{0.24\linewidth}X@{}}
\toprule
Certificate component & Sufficient source or restriction \\
\midrule
Target joint-treatment law $g_{z,0}$ &
Independent joint action/exposure-cell audit, or a verified joint
action-generation model. A marginal local-action mixture is sufficient
only when the induced joint-law identity is separately proved. \\
No-exploration context $\nu_{z,0}$ &
Target-regime context audit, or a proven structural transport bound.
Exploratory propensity overlap alone is insufficient. \\
Target-regime conditional response $m_z(\widetilde a,c^-)$ &
Identified interventional audit of the target response, or an explicit
regime-invariance/exposure-sufficiency assumption with a certified
approximation allowance. \\
Clipping and nuisance error &
Joint-support bounds and fitted-model validation on independent or
appropriately buffered evaluation data, including a bound on the induced
clipped joint-ratio error. \\
Network dependence and drift &
Predictable-score martingale conditions, independently validated mixing
or block-coupling constants, and a justified drift envelope; an empirical
autocorrelation is not by itself a certificate. \\
\bottomrule
\end{tabularx}
\end{center}
The finite-state audit constructions provide examples of these sources under strong sampling and identification assumptions.
When a required joint-treatment audit cell or structural restriction is unavailable, the corresponding budget is \emph{uncertified}, not zero; Theorem~\ref{thm:main} then does not supply an operational finite-sample coverage claim.

Any operator-collected pilot or audit windows used to learn $\nu_{z,0}$ or the target-regime response must be counted in the operational measurement budget.
For audit deployments $\mathcal A$, one can record separately their measurement-induced loss and any configuration-selection loss as
\begin{align}
D_{\mathrm{audit}}
&=\sum_{r\in\mathcal A}
 \{W_\infty(z_r;0)-W_\infty(z_r;\eta_r)\},\\
R_{\mathrm{audit}}
&=\sum_{r\in\mathcal A}
 \{W_\infty(z^\star;0)-W_\infty(z_r;0)\}.
\label{eq:supp-audit-cost}
\end{align}
These expressions are accounting definitions, not additional lower bounds.
If independent target-regime audits are supplied externally, their data-collection cost must instead be disclosed separately.
In the paper's restricted matched-reward construction, such target-regime audits are not required because the relevant target value has a known identifiable form.

\subsubsection{Anytime validity and a concrete sufficient growth condition}
For a finite evaluation horizon, allocate failure probabilities $\delta_t>0$ such that $\sum_t\delta_t\le\delta$ and divide each $\delta_t$ across the centered-score event, calibration, coupling, and drift conditions actually invoked.
A fixed-round certificate at level $1-\delta$ cannot simply be reused without adjustment after data-dependent online selection.

For the finite candidate set, write $N_z(t-1)$ for the number of qualifying past deployments of configuration $z$.
One \emph{sufficient} condition, already established through the counting argument in the existing information-growth appendix, is that outside at most $H$ initialization rounds there exists a fixed $\lambda>0$ such that
\begin{equation}
\underline n_{\mathrm{eff},t-1}(z_t)
\ge\lambda\{1+N_{z_t}(t-1)\}
\quad\text{for each eligible }t.
\label{eq:supp-selected-growth}
\end{equation}
Then a pathwise counting argument gives
\[
\sum_{t\ \mathrm{eligible}} \frac1{\sqrt{\underline n_{\mathrm{eff},t-1}(z_t)}}
\le 2\sqrt{\frac{|\mathcal Z|T}{\lambda}}.
\]
Combined with a valid simultaneous confidence event and certified bias budgets $e_{t-1}(z_t)$, the standard OFU argument yields a conditional bound of the form
\[
\operatorname{Reg}_T
\le 2B_u H
+O\!\left(B_\psi\sqrt{\frac{|\mathcal Z|T
[\operatorname{comp}(\mathcal Z)+\log(T/\delta)]}{\lambda}}\right)
+2\sum_{t\ \mathrm{eligible}}e_{t-1}(z_t).
\]
This repeats the logical implication already present in the information-growth appendix.
OFU \emph{does not imply}
\eqref{eq:supp-selected-growth}: it must be supported by certified
scoring-window production, overlap, and the dependence model.
A measurement schedule can enforce overlap only when its action and context-support assumptions have been verified.
Without such a route, regret convergence remains a conditional guarantee.

\paragraph{Relation to the identification--deployment lower bound.}
Theorem~\ref{thm:frontier} is a minimax lower-bound construction with potentially horizon-dependent mixing time, whereas Theorem~\ref{thm:regret} is an upper-bound implication under calibration and selected-information growth.
Their assumptions and loss criteria do not currently supply a matched Pareto lower/upper bound on one common model class.
Such a matching result would require additional algorithmic and theoretical analysis; it is not claimed here.

%% file: appendix/12_fixed_configuration_concentration.tex
\subsection{Fixed-Configuration Concentration}
\label{app:concentration-detailed}

Fix a candidate configuration $z$ and an evaluation round $t$.
Use the network-window score sequence $Y_{j,z}$ and count $J_t=T_0t$ defined in Appendix~\ref{app:ss-blocks}.
Each $Y_{j,z}$ is the within-window average of the node-level clipped DR scores, so $|Y_{j,z}|\le B_\psi$ without assuming independence among nodes in the same window.
The stored score rule is predictable with respect to the pre-window filtration $\mathcal F_{j-1}$.

Define the predictable-score average
\[
\bar Y_t^{\mathrm{pred}}(z)
= \frac1{J_t}\sum_{j=1}^{J_t}
\mathbb E[Y_{j,z}\mid\mathcal F_{j-1}].
\]
The martingale-difference route gives, for every $\alpha\in(0,1)$,
\begin{equation}
\mathbb P\!\left(
\left|
\widehat W_t(z;0)-\bar Y_t^{\mathrm{pred}}(z)
\right|
>
2B_\psi\sqrt{\frac{2\log(2/\alpha)}{J_t}}
\right)
\le\alpha.
\label{eq:fixed-conc-martingale}
\end{equation}
No independence or mixing assumption across network windows is needed for this centered martingale statement.
It does not, however, bound the predictable-mean discrepancy $|\bar Y_t^{\mathrm{pred}}(z)-W_\infty(z;0)|$.
That discrepancy is controlled separately by the calibrated identification, nuisance, transport, and drift budgets in Appendices~\ref{app:thm1proof-detailed} and~\ref{app:v7-certification}.

For compatibility with the main-text radius, suppose the certified information count satisfies
\[
0<\underline n_{\mathrm{eff},t}(z)\le J_t/4.
\]
Then Eq.~\eqref{eq:fixed-conc-martingale} implies the conservative bound
\begin{equation}
\mathbb P\!\left(
\left|
\widehat W_t(z;0)-\bar Y_t^{\mathrm{pred}}(z)
\right|
>
2B_\psi\sqrt{\frac{\log(2/\alpha)}
{\underline n_{\mathrm{eff},t}(z)}}
\right)
\le\alpha.
\label{eq:fixed-conc}
\end{equation}
The inequality follows because $\underline n_{\mathrm{eff},t}(z)\le J_t/4$ makes the displayed radius at least as large as the Azuma--Hoeffding radius.
A valid bound on the predictable-mean discrepancy must then be added to obtain a value certificate.
The operational ESS shrinkage proxy in Appendix~\ref{app:ss-blocks} is diagnostic only; it cannot replace the certified lower-information count.

An alternative block-coupling route is available when the score sequence is stationary conditional on a separate fitting field and a valid absolute-regularity envelope is supplied.
In that case, use the explicit block-coupled inequality in Lemma~\ref{lem:v7-block} of Appendix~\ref{app:v7-certification}.
A block-coupling radius is not implied by an empirical autocorrelation estimate alone.

%% file: appendix/13_uniformization_over_the_configuration_space.tex
\subsection{Uniformization over the Configuration Space}
\label{app:uniform-detailed}

Let $\mathcal C_\varepsilon$ be a finite $\varepsilon$-cover of the configuration space $\mathcal Z$, with cardinality $\mathcal N(\varepsilon,\mathcal Z)$.
At every net point, apply the fixed-configuration certificate with failure probability $\delta/\mathcal N(\varepsilon,\mathcal Z)$, allocating that probability among the centered-score and calibration events actually used.
A union bound gives a simultaneous event on which
\[
\max_{z'\in\mathcal C_\varepsilon} |\widehat W_t(z';0)-W_\infty(z';0)|
\le
\max_{z'\in\mathcal C_\varepsilon}
\left\{
2B_\psi
\sqrt{\frac{\log(2\mathcal N(\varepsilon,\mathcal Z)/\delta)}
{\underline n_{\mathrm{eff},t}(z')}}
+ \bar\epsilon_{\mathrm{tot},t}(z') + \widehat\kappa_t(z') \right\},
\]
provided the componentwise calibration conditions of Theorem~\ref{thm:main} hold at all net points.

If both $\widehat W_t(\cdot;0)$ and $W_\infty(\cdot;0)$ are uniformly $L_Z$-Lipschitz on the relevant confidence event, every $z\in\mathcal Z$ has a net point $z'$ with $d(z,z')\le\varepsilon$, and
\[
|\widehat W_t(z;0)-W_\infty(z;0)|
\le
|\widehat W_t(z';0)-W_\infty(z';0)|+2L_Z\varepsilon.
\]
Thus the cover adds the explicit remainder $2L_Z\varepsilon$, which must be included in the approximation/error budget.
Neither the union bound nor a finite cover establishes the Lipschitz condition or the identification and transport budgets.
For this argument, $\operatorname{comp}(\mathcal Z)$ may be taken as $\log\mathcal N(\varepsilon,\mathcal Z)$, with the covering remainder reported separately.

%% file: appendix/14_proof_of_theorem_1_conditional_certificate.tex
\subsection{Proof of Theorem 1: Conditional Certificate}
\label{app:thm1proof-detailed}

Fix a predecision round $t\ge2$ and a candidate configuration $z$.
Flatten the historical post-burn-in network windows available before round $t$ into a time-ordered sequence $j=1,\ldots,J_{t-1}$, where
\[
J_{t-1}=(t-1)T_0.
\]
Let $(r(j),s(j))$ denote the deployment round and within-round window corresponding to index $j$, and write
\[
\widetilde O_{i,j}
= \bigl(\widetilde A_{i,j},C^-_{i,j},R_{i,j}\bigr),
\qquad
\widetilde A_{i,j}=(A_{i,j},E_{i,j}).
\]
Here $C^-_{i,j}$ contains only information available strictly before the focal action, while $\widetilde A_{i,j}$ denotes the observed joint action/exposure treatment unit.

Let $\mathcal F_{j-1}$ denote the information available before scoring window $j$.
The nuisance predictions, fitted behavior joint law, and scoring rule used for that window are $\mathcal F_{j-1}$-measurable.
This predictable, prequential convention is required for the filtration-based concentration argument below.

Define the network-window score
\[
Y_{j,z} = \frac1N\sum_{i=1}^N \widehat\psi_{j,z}(\widetilde O_{i,j}),
\qquad
\widehat W_{t-1}(z;0)
= \frac1{J_{t-1}}\sum_{j=1}^{J_{t-1}}Y_{j,z}.
\]
The score $\widehat\psi_{j,z}$ is the stored prequential score rule for this window.
The target-law and response fits used inside it are frozen before the window is scored, and the behavior joint law is the fit for source round $r(j)$.
The fitting-cutoff indices are suppressed inside the main-text notation $\widehat\psi_{z,r}$.
This is exactly the main-text normalization:
\[
\widehat W_{t-1}(z;0)
= \frac1{NT_0(t-1)} \sum_{r=1}^{t-1}\sum_{s=1}^{T_0}\sum_{i=1}^N \widehat\psi_{z,r}(\widetilde O_{i,s}^{(r)}).
\]
Thus $J_{t-1}$ counts network windows, whereas $NJ_{t-1}$ counts node-level records.
The concentration argument treats the aggregated network-window scores as its time-ordered units and does not assume independence among nodes within a window.

\paragraph{Reference laws and transported responses.}
For each historical window $j$, let $P_{z,j}^{\mathrm{eval}}$ denote the corresponding post-burn-in reference evaluation law for the stored score of candidate $z$.
Its source context marginal is $\nu_{z_{r(j)},\eta_{r(j)}}$, and its conditional joint action/exposure law is denoted by $g_{b,r(j)}(\cdot\mid C^-)$.
Expectations under this law hold the fitted scoring rule fixed.
The difference between the actual conditional score mean and this reference-law mean is handled separately as predictable-window drift.

Let $m_{b,r(j)}(\widetilde a,c^-)$ be the source-regime conditional response mean, and define
\[
V_{j,z}^{\mathrm{log}}(c^-) = \int_{\widetilde{\mathcal A}} m_{b,r(j)}(\widetilde a,c^-) g_{z,0}(d\widetilde a\mid c^-),
\]
where $g_{z,0}$ is the target joint action/exposure law without measurement exploration.
This definition presumes that the source response is meaningful on the relevant target support and that the corresponding transport conditions are satisfied.

The deviation admits the following exact decomposition:
\begin{align}
\widehat W_{t-1}(z;0)-W_\infty(z;0)
={}&
\frac1{J_{t-1}}\sum_{j=1}^{J_{t-1}}
\left[
Y_{j,z}
-\mathbb E[Y_{j,z}\mid\mathcal F_{j-1}]
\right]
\notag\\
&+
\frac1{J_{t-1}}\sum_{j=1}^{J_{t-1}}
\left[
\mathbb E[Y_{j,z}\mid\mathcal F_{j-1}]
-\mathbb E_{P_{z,j}^{\mathrm{eval}}}
\widehat\psi_{j,z}
\right]
\notag\\
&+
\frac1{J_{t-1}}\sum_{j=1}^{J_{t-1}}
\left[
\mathbb E_{P_{z,j}^{\mathrm{eval}}}
\widehat\psi_{j,z}
-\mathbb E_{P_{z,j}^{\mathrm{eval}}}
V_{j,z}^{\mathrm{log}}(C^-)
\right]
\notag\\
&+
\frac1{J_{t-1}}\sum_{j=1}^{J_{t-1}}
\left[
\mathbb E_{P_{z,j}^{\mathrm{eval}}}
V_{j,z}^{\mathrm{log}}(C^-)
-\mathbb E_{C^-\sim\nu_{z,0}}
V_{j,z}^{\mathrm{log}}(C^-)
\right]
\notag\\
&+
\frac1{J_{t-1}}\sum_{j=1}^{J_{t-1}}
\left[
\mathbb E_{C^-\sim\nu_{z,0}}
V_{j,z}^{\mathrm{log}}(C^-)
-W_\infty(z;0)
\right].
\label{eq:repaired-proof-telescope}
\end{align}

\paragraph{First term: stochastic fluctuation.}
The first term is the centered fluctuation of the historical network-window scores.
It is controlled by the validated conditional concentration inequality under the dependence assumptions of Appendix~\ref{app:concentration-detailed}, using the effective sample size calibrated for the sequence $\{Y_{j,z}\}$.
The bound must apply to the actual predictably fitted score sequence; empirical autocorrelation alone does not establish the required concentration event.

\paragraph{Second term: predictable-window drift.}
The second term accounts for discrepancies between the conditional score means and their reference evaluation-law counterparts.
Define the signed predictable discrepancy
\[
d_{j,z} =
\mathbb E[Y_{j,z}\mid\mathcal F_{j-1}]
-
\mathbb E_{P_{z,j}^{\mathrm{eval}}}
\widehat\psi_{j,z}.
\]
Its contribution must be bounded by the independently justified drift allowance.
In particular, the required pathwise calibration is
\[
\left| \frac1{J_{t-1}}\sum_{j=1}^{J_{t-1}}d_{j,z} \right|
\le \widehat\kappa_{t-1}(z)
\]
on the stated calibration event.
An empirical inter-block mean difference is a diagnostic unless its relation to this predictable discrepancy has been validated.

\paragraph{Third term: nuisance error and score-level response transport.}
For a given source round $r=r(j)$, write
\[
p_z(\widetilde a\mid c^-) = g_{z,0}(\widetilde a\mid c^-),
\qquad
b_r(\widetilde a\mid c^-) = g_{b,r}(\widetilde a\mid c^-),
\]
and let
\[
\Delta_{b,r,z}(\widetilde a,c^-) = m_{b,r}(\widetilde a,c^-) - m_z(\widetilde a,c^-).
\]
For the stored score in source round $r$, write $\widehat p_{z,r}=\widehat g_{z,0,r}$ and $\widehat m_{z,r}$ for the fitted target joint law and response regression; the main text suppresses these fitting-cutoff indices.
Let $w_{z,r}=p_z/b_r$ be the oracle joint ratio, $\widehat w_{z,r}$ the fitted clipped joint ratio, and $\bar w_{z,r}=\min\{w_{z,r},w_{\max}\}$ the clipped oracle ratio.
Ratios are interpreted relative to a common dominating measure.

Under the exact nuisance decomposition in Appendix~\ref{app:nuisance-detailed}, the third term has the following conditional algebra.
Let $\nu_j=\nu_{z_{r(j)},\eta_{r(j)}}$, $p_z=g_{z,0}$, $b_{r(j)}=g_{b,r(j)}$, $\delta_{m,j}=\widehat m_{z,r(j)}-m_z$, $\delta_{w,j}=\widehat w_{z,r(j)}-\bar w_{z,r(j)}$, and $\Delta_{b,r(j),z}=m_{b,r(j)}-m_z$.
With expectations taken under the source-window reference law and fitted quantities held fixed,
\begin{align}
&\mathbb E_{P_{z,j}^{\mathrm{eval}}}\widehat\psi_{j,z}
-\mathbb E_{P_{z,j}^{\mathrm{eval}}}V_{j,z}^{\mathrm{log}}(C^-)
\notag\\
={}&\mathbb E_{C^-\sim\nu_j}\Bigg[
\int_{\widetilde{\mathcal A}}
(\widehat p_{z,r(j)}-p_z)(\widetilde a\mid C^-)
\widehat m_{z,r(j)}(\widetilde a,C^-)
\,d\widetilde a
\notag\\
&\quad+
\int_{\widetilde{\mathcal A}}
(p_z-b_{r(j)}\bar w_{z,r(j)})(\widetilde a\mid C^-)
\delta_{m,j}(\widetilde a,C^-)
\,d\widetilde a
\notag\\
&\quad-
\int_{\widetilde{\mathcal A}}
b_{r(j)}(\widetilde a\mid C^-)
\delta_{w,j}(\widetilde a,C^-)
\delta_{m,j}(\widetilde a,C^-)
\,d\widetilde a
\notag\\
&\quad+
\int_{\widetilde{\mathcal A}}
\big[b_{r(j)}\widehat w_{z,r(j)}-p_z\big]
(\widetilde a\mid C^-)
\Delta_{b,r(j),z}(\widetilde a,C^-)
\,d\widetilde a
\Bigg].
\label{eq:proof-third-term-exact}
\end{align}
For discrete joint treatments, the integrals are sums.
The first three terms are bounded by the target-joint-law, clipping, and nuisance-product budgets.
The last term is the score-level response-transport residual
\begin{align}
\epsilon_{\mathrm{score-resp},j}(z)
={}&
\mathbb E_{C^-\sim\nu_j}
\left|
\int_{\widetilde{\mathcal A}}
\big[b_{r(j)}(\widetilde a\mid C^-)
\widehat w_{z,r(j)}(\widetilde a,C^-)
-p_z(\widetilde a\mid C^-)\big]
\Delta_{b,r(j),z}(\widetilde a,C^-)
\,d\widetilde a
\right|.
\label{eq:score-level-response-transport}
\end{align}
This term is distinct from the target-context response discrepancy $\epsilon_{\mathrm{exp},r(j)}^{\mathrm{target}}(z)$ in Appendix~\ref{app:exposure-majorant}: the former is evaluated under the source context law and uses the fitted clipped ratio, whereas the latter is evaluated under the target context law and the target joint law.
If $|\Delta_{b,r,z}|\le2B_u$, a conservative bound is
\begin{equation}
\epsilon_{\mathrm{score-resp},j}(z)
\le
2B_u\,
\|\widehat w_{z,r(j)}-\bar w_{z,r(j)}\|_{L_1(\nu_j,b_{r(j)})}
+
\epsilon_{\mathrm{clip},j}(z),
\label{eq:score-level-response-transport-bound}
\end{equation}
where
\[
\epsilon_{\mathrm{clip},j}(z) =2B_u\,\mathbb E_{C^-\sim\nu_j}
\mathbb E_{\widetilde A\sim b_{r(j)}(\cdot\mid C^-)}
[(w_{z,r(j)}-w_{\max})_+].
\]
This follows from $p_z-b_{r(j)}\bar w_{z,r(j)} =b_{r(j)}(w_{z,r(j)}-w_{\max})_+$ and does not assume that correcting the joint treatment law eliminates response transport.

Accordingly, the third term is bounded by the calibrated target-joint-law, clipping, and nuisance-product errors, plus $\epsilon_{\mathrm{score-resp},j}(z)$ and any nuisance-distribution transfer allowance required to move the nuisance bounds to the reference evaluation law.
The response discrepancy vanishes only under the relevant conditional response-invariance condition or an equivalent justified transport guarantee.
Correcting the joint action/exposure distribution alone does not establish that condition.

\paragraph{Fourth term: context shift.}
The fourth term compares expectations of the same source-response functional under the source context law and the target context law.
It is controlled by the calibrated context-shift allowance in Appendix~\ref{app:context-shift-detailed}.
The calibration must apply to the relevant source round and historical averaging scheme; a context-distribution discrepancy without a justified function-class bound does not establish the required expectation bound.

\paragraph{Fifth term: target-law response transport.}
The fifth term is controlled by the target-law exposure/response discrepancy in Appendix~\ref{app:exposure-majorant}.
In particular,
\[
\left|
\mathbb E_{C^-\sim\nu_{z,0}}
V_{j,z}^{\mathrm{log}}(C^-) -W_\infty(z;0) \right|
\le
\epsilon_{\mathrm{exp},r(j)}^{\mathrm{target}}(z)
\]
when the target value and source-response functional satisfy the definitions and identification conditions in that appendix.

The score-level discrepancy in Eq.~\eqref{eq:score-level-response-transport} and the target-law discrepancy above are distinct contributions in this particular telescope.
The single main-text budget $\epsilon_{\mathrm{exp}}(z)$ must be calibrated to dominate their historical average, as well as any exposure-representation error:
\[
\frac1{J_{t-1}}\sum_{j=1}^{J_{t-1}}
\left[
\epsilon_{\mathrm{score-resp},j}(z) + \epsilon_{\mathrm{exp},r(j)}^{\mathrm{target}}(z) \right].
\]
Alternatively, an independently justified tighter bound on the combined contribution may be used.
If the existing $\epsilon_{\mathrm{exp}}(z)$ calibration controls only the target-law discrepancy, it must be extended to cover the score-level term as well.
The component bounds are folded into the single $\epsilon_{\mathrm{exp}}(z)$ allowance; they are not then added again as separate response terms.
The clipping allowance must also cover every clipping contribution actually retained in the exact nuisance decomposition.
This is a clarification of the existing error budget, not a change to the main theorem's core expression.

\paragraph{Combining the bounds.}
Combining the stochastic fluctuation bound with the predictable-drift, nuisance, context-shift, and exposure/response-transport bounds yields the deviation inequality in Theorem~\ref{thm:main}, provided that the averaged non-statistical contributions are bounded by $\bar\epsilon_{\mathrm{tot},t-1}(z)$ and the predictable drift is bounded by $\widehat\kappa_{t-1}(z)$ under the stated calibration.

The target-joint-law, clipping, and nuisance-product contributions must be calibrated for the actual source rounds and reference laws used in the estimator.
The configuration-mismatch allowance is included only when needed to transfer nuisance guarantees across training and evaluation distributions; a response discrepancy already covered by the exposure/response budget must not be counted again.
All component bounds must apply to the corresponding historical windows and their averaged contributions.
A bound for a single evaluation round does not by itself bound the historical average.

A simultaneous guarantee over configurations requires a valid union or covering event, with the allocated failure probabilities included in the overall confidence budget.
For a continuous configuration space, the covering argument must additionally account for an explicit Lipschitz covering remainder in the approximation/error budget; it does not follow from the union bound alone.

Empirical autocorrelation estimates and empirical inter-block mean differences are diagnostics only.
They do not independently establish the required high-probability concentration, drift, identification, or transport events.
The certificate therefore holds only when each component bound and the corresponding dependence and calibration conditions are justified.

%% file: appendix/17_identification_deployment_frontier_a_slow_mixing_steady.tex
\subsection{Identification--Deployment Frontier: A Slow-Mixing Steady-State Hard Family}\label{app:frontier-detailed}

We construct a two-configuration, two-action, two-node steady-state family in which the parameter changes the no-exploration configuration value itself, normal deployment of either configuration is permitted, and the full adaptively collected log has information of order only $1+\sum_{t=1}^T\eta_t$.
The crucial mechanism is that ordinary deployment can repeatedly reveal the \emph{same} latent state, whereas measurement exploration can refresh that state.
This construction is valid as a horizon-indexed hard family when the model class does not impose a \emph{horizon-uniform} bound on the mixing time.
The latter qualification is mathematically essential and is discussed below.

\paragraph{Instance and steady-state target.}
Fix a horizon $T\ge1$ and a predetermined exploration schedule $\eta_1,\ldots,\eta_T\in[0,1]$, and set $E_T=\sum_{t=1}^T\eta_t$.
Let $\theta\in\{-1,+1\}$ index the two environments, and define
\[
q_\theta=\tfrac12+\theta\Delta,\qquad \Delta=\frac{1}{32\sqrt{1+E_T}},\qquad \varepsilon_T=T^{-2}.
\]
There are two configurations $\mathcal Z=\{z_0,z_1\}$, two actions $\mathcal A=\{a_0,a_1\}$, and two network nodes sharing a latent state $U_t\in\{0,1\}$.
The graph contains an edge between the nodes; the exposure of either node may be taken to be the realized action of its neighbor.
This exposure and the static observed context are functions of the logged actions and do not reveal additional latent randomness.
The latent state starts in its stationary law $U_0\sim\operatorname{Ber}(q_\theta)$.
We take one post-burn-in observation window per slow round ($T_0=1$).
Any prescribed fixed number of burn-in steps can be inserted as identity updates before this window: the controller is already in stationarity, and these steps reveal no extra observations.
This is a legitimate (trivial) burn-in subcase of the black-box adaptation model.

Under either configuration, the no-exploration controller always chooses $a_0$.
At slow round $t$, each node independently selects $a_1$ with probability $\eta_t/2$ and $a_0$ otherwise.
Thus its action law is exactly
\[
\pi^{\mathrm{obs}}_{z,\eta_t}=(1-\eta_t)\delta_{a_0} +\eta_t\xi,\qquad \xi(a_0)=\xi(a_1)=\tfrac12>0,
\]
independently of which configuration $z_t$ was adaptively selected.
Draw an unobserved spontaneous refresh marker $B_t\sim\operatorname{Ber}(\varepsilon_T)$ independently of all prior randomness.
Define $J_t=\mathbf1\{B_t=1\text{ or }A_{1,t}=a_1\text{ or }A_{2,t}=a_1\}$.
If $J_t=1$, refresh $U_t$ as an independent $\operatorname{Ber}(q_\theta)$ draw; otherwise set $U_t=U_{t-1}$.
The observed normalized per-node reward is
\[
R_{i,t}=g_{z_t}(U_t)-\tfrac14\mathbf1\{A_{i,t}=a_1\},\qquad g_{z_0}(u)=\tfrac12,\quad g_{z_1}(u)=u.
\]
In particular, $|R_{i,t}|\le1$ and the observed logs contain the realized configuration, actions, graph/exposure context, and these rewards.
The latent-state transition is not supplied to the operator.
Each node's exploratory action can refresh a shared state, and therefore affects future rewards of both nodes.

For any fixed configuration and fixed $\eta$, the hidden-state chain is aperiodic and has the unique stationary distribution $\operatorname{Ber}(q_\theta)$, since a spontaneous refresh occurs with positive probability $\varepsilon_T$.
A refresh draws from this same law, so the stationary distribution does not change with $\eta$.
Writing network utility as the average of the two node rewards, the no-exploration target values are exactly
\begin{equation}
W^\theta_\infty(z_0;0)=\tfrac12,\qquad
W^\theta_\infty(z_1;0)=\tfrac12+\theta\Delta.
\label{eq:slow-frontier-value}
\end{equation}
The better configuration therefore depends on $\theta$, and the target value separation at $z_1$ is $2\Delta$.
This separation concerns the paper's actual $W_\infty(z;0)$, not the reward of an action absent from the no-exploration target.

\paragraph{KL bound for every adaptive configuration-selection rule.}
Let $P_\theta^T$ be the complete observed history under environment $\theta$, including any randomized configuration decisions.
It suffices, by data processing, to upper-bound the KL of a more informative augmented experiment exposing $U_0$ and every action, spontaneous refresh marker, and refreshed latent draw.
Conditional on any augmented history, the algorithm's configuration-selection kernel is the same function under the two environments, the action and spontaneous-marker kernels are identical, and only a newly sampled $\operatorname{Ber}(q_\theta)$ state contributes a likelihood difference.
The chain rule hence gives
\begin{equation}
D_{\mathrm{KL}}(P_+^T\Vert P_-^T)
\le d(\Delta)\biggl(1+\sum_{t=1}^T\mathbb E_+[J_t]\biggr),
\quad d(\Delta)=\mathrm{kl}\!\left(\operatorname{Ber}(\tfrac12+\Delta),
\operatorname{Ber}(\tfrac12-\Delta)\right).
\label{eq:slow-frontier-kl}
\end{equation}
Here
\[
\mathbb P(J_t=1)=1-(1-\varepsilon_T)(1-\eta_t/2)^2
\le \varepsilon_T+\eta_t
\]
by the union bound.
For $0<\Delta\le1/32$, $ d(\Delta)=2\Delta\log\frac{1/2+\Delta}{1/2-\Delta} \le16\Delta^2$.
Because $T\varepsilon_T=1/T\le1$,
\[
D_{\mathrm{KL}}(P_+^T\Vert P_-^T)
\le 16\Delta^2(2+E_T)
\le32\Delta^2(1+E_T)\le\tfrac1{32}.
\]
This upper bound accounts for all ordinary deployment observations, including repeated deployments of $z_1$ with $\eta_t=0$; they are potentially numerous but mostly repeat the same hidden Bernoulli state.

\paragraph{Value-identification lower bound.}
By Pinsker's inequality, $\operatorname{TV}(P_+^T,P_-^T)\le\sqrt{(1/32)/2}=1/8$.
For any estimate $\widehat W$ of $W_\infty(z_1;0)$, choose the sign corresponding to the closer target value in Eq.~\eqref{eq:slow-frontier-value}.
A sign error forces $|\widehat W-W_\infty^\theta(z_1;0)|\ge\Delta$.
The two-point testing inequality therefore yields
\begin{align}
\inf_{\widehat W}\max_{\theta\in\{-1,+1\}}
\mathbb E_\theta|\widehat W-W_\infty^\theta(z_1;0)|
&\ge\frac\Delta2\bigl(1-\operatorname{TV}(P_+^T,P_-^T)\bigr)\\
&\ge\frac{7\Delta}{16}
\ge\frac{1}{100\sqrt{1+E_T}}.
\label{eq:slow-frontier-minimax}
\end{align}
The infimum includes estimators using all logs and all adaptive configuration deployments.
Thus the original cumulative-exploration rate holds for this horizon-indexed family.

\paragraph{Deployment loss on the same instance.}
For every fixed configuration $z$ and exploration level $\eta$, the stationary latent distribution is unchanged.
Each node selects $a_1$ with probability $\eta/2$ and pays utility cost $1/4$ on that action.
Therefore
\[
W_\infty^\theta(z;0)-W_\infty^\theta(z;\eta)=\eta/8.
\]
For arbitrary adaptive configuration decisions and the fixed schedule,
\begin{equation}
\mathrm{DeploymentGap}_T
=\sum_{t=1}^T\bigl[W_\infty^\theta(z_t;0)-
W_\infty^\theta(z_t;\eta_t)\bigr]
=\frac18\sum_{t=1}^T\eta_t.
\label{eq:slow-frontier-gap}
\end{equation}
Consequently the two original-rate lower bounds hold simultaneously on one constructed family with universal constants $c_1=1/100$ and $c_2=1/8$.

\paragraph{Essential mixing-time qualification.}
Under $\eta=0$ the latent transition kernel is $K_\theta(u,\cdot)=(1-\varepsilon_T)\delta_u+ \varepsilon_T\operatorname{Ber}(q_\theta)$.
It is geometrically mixing for every \emph{fixed} horizon $T$, with a decay factor $(1-\varepsilon_T)^k$, but its mixing time is of order $T^2$.
The horizon-dependent mixing time is an explicit feature of the minimax class in Theorem~\ref{thm:frontier}: each fixed-$T$ process mixes geometrically, with mixing times of order $T^2$ across the family.
The bound therefore characterizes the stated predetermined exploration schedule and the associated horizon-dependent class.
A horizon-uniform mixing restriction or an adaptively chosen measurement schedule defines a different lower-bound problem, with its own information budget and mixing assumptions.

%% file: appendix/18_causal_ucb_regret_proof.tex
\subsection{CAUSAL-UCB Regret Proof}\label{app:regret-detailed}
For each round $t\ge2$, put $\widetilde W_t(z)=\widehat W_{1:t-1}(z;0)$ and let $b_t(z)$ be the corresponding radius computed from the same \emph{past} information, as specified in Algorithm~\ref{alg:causalucb}.
Let $\mathcal E$ denote the simultaneously valid calibration/concentration event on which $|\widetilde W_t(z)-W_\infty(z;0)|\le b_t(z)$ for every candidate and round.
The event requires valid failure-probability allocation across rounds, configurations, and all calibration subevents in Appendix~\ref{app:v7-certification}.

On $\mathcal E$, the optimism step $z_t\in\arg\max_z
\{\widetilde W_t(z)+b_t(z)\}$ gives
\begin{align}
W_\infty(z^\star;0)
&\le\widetilde W_t(z^\star)+b_t(z^\star)\\
&\le\widetilde W_t(z_t)+b_t(z_t),
\end{align}
and therefore
\[
W_\infty(z^\star;0)-W_\infty(z_t;0) \le\widetilde W_t(z_t)+b_t(z_t)-W_\infty(z_t;0) \le2b_t(z_t).
\]
For the first, uninitialized deployment $t=1$, bounded rewards give a constant regret contribution (at most $2B_u$); the algorithm may use an arbitrary fixed tie-breaking rule when all initial radii are infinite.
With
\[
b_t(z)=2B_\psi\sqrt{\frac{\operatorname{comp}(\mathcal Z)+\log(2T/\delta)}
 {\underline n_{\mathrm{eff},t-1}(z)}}
+\bar\epsilon_{\mathrm{tot},t-1}(z)+\widehat\kappa_{t-1}(z),
\]
valid for $t\ge2$, summing produces
\[
\operatorname{Reg}_T\le2B_u+2\sum_{t=2}^T b_t(z_t).
\]
The simplification to the stated $\widetilde O(B_\psi\sqrt{T\operatorname{comp}(\mathcal Z)})$ stochastic term requires the \emph{selected-arm, pre-deployment} summability bound $\sum_{t=2}^T\underline n_{\mathrm{eff},t-1}(z_t)^{-1/2} =\widetilde O(\sqrt T)$.
It does not follow from optimism alone and need not hold under insufficient exploration.
The additive deterministic terms then produce the conditional form in Theorem~\ref{thm:regret} (up to the harmless one-step index shift).

%% file: appendix/19_deployment_gap_corollary.tex
\subsection{Deployment-Gap Corollary}\label{app:gap-detailed}
If $\Winf(z;\eta)$ is $L_W$-Lipschitz in $\eta$ uniformly over $z$, then
\[
\Winf(z_t;0)-\Winf(z_t;\eta_t)
\le
|\Winf(z_t;0)-\Winf(z_t;\eta_t)|
\le
L_W\eta_t.
\]
Summing gives
\[
\sum_{t=1}^T\{\Winf(z_t;0)-\Winf(z_t;\eta_t)\}
\le
L_W\sum_{t=1}^T\eta_t.
\]
This is separate from target regret: regret evaluates no-exploration configuration quality, while deployment gap evaluates the cost of collecting identifiable logs.

%% file: appendix/28_restricted_case_certificate_proofs.tex
\subsection{Proofs for the Restricted Local-Interference Certificate}
\label{app:restricted-case-proofs}

This appendix proves Proposition~\ref{prop:certified-special} and Corollary~\ref{cor:restricted-regret}.
The claims are restricted to the specified homogeneous local-interference model; they do not assert that unknown target context laws or dependent black-box controllers can be certified from exploratory logs alone.

\subsubsection{A simultaneous event for matched rewards}
For each $z\in\mathcal Z$, expose the matched rewards in their order of observation, breaking ties within one network window by a fixed node order.
Denote this sequence by $G_{z,1},G_{z,2},\ldots$.
The assumptions in the restricted model allow us to couple these observations to an independent, pre-generated sequence of draws distributed as $\operatorname{clip}(\mu_z+\sigma Z,0,1)$.
Adaptive selection of $z$ only reveals a prefix of this sequence; the matched-reward distribution is unaffected by past configuration choices or by the indicator array.

The clipping map $x\mapsto\operatorname{clip}(x,0,1)$ is $1$-Lipschitz.
By Gaussian concentration, every matched reward is $\sigma$-sub-Gaussian around its mean $W_z$ (the same statement holds when $\sigma$ is a conservative sub-Gaussian scale bound).
For every fixed $m\ge1$,
\begin{equation}
\Pr\!\left(\left|\frac1m\sum_{\ell=1}^mG_{z,\ell}
              -W_z\right|>
                 \sigma\sqrt{\frac{2L}{m}}\right)
\le 2e^{-L},
\qquad L=\log\frac{2K(M+1)}{\delta}.
\label{eq:restricted-prefix-tail}
\end{equation}
Here $M=NTT_0$ is an upper bound on the total number of node-window reward slots in the entire experiment, and $K=|\mathcal Z|$.
All means of observations in $[0,1]$ differ by at most $1$, so the radius may be truncated at $1$.
A union bound over the $K$ candidates and all nonempty prefix lengths $1\le m\le M$ shows that the event
\begin{equation}
\mathcal E_{\mathrm{match}}
=\bigcap_{z\in\mathcal Z}\ \bigcap_{m=1}^{M}
\left\{\left|\frac1m\sum_{\ell=1}^{m}G_{z,\ell}-W_z\right|
\le\epsilon(m)\right\}
\label{eq:restricted-sim-event}
\end{equation}
has probability at least $1-\delta$, because $2KM e^{-L}\le \delta$.
The event is defined on pre-generated streams and therefore also holds at all adaptively selected stopping and examination times.
No sample splitting across examination times is required.
At matched count $m=0$, the prescribed predictor is $1/2$, so $|1/2-W_z|\le1/2=\epsilon(0)$ deterministically.

\subsubsection{Proof of Proposition~\ref{prop:certified-special}}
\begin{proof}
Fix any configuration $z$ and any nonempty prefix of $n$ network windows collected at its selected deployments.
Put $e_{j-}=\widehat m_{j-}(z)-W_z$ and $h_j=K_j/(Nq)$.
For every scored window, an exact algebraic rearrangement gives
\begin{align}
Y_{j,z}-W_z
&=(1-h_j)e_{j-}
  +\frac1{Nq}\sum_{i=1}^{N}I_{j,i}(R_{j,i}-W_z).
\label{eq:restricted-score-decomposition}
\end{align}
Summing over windows and taking absolute values yields
\begin{align}
|\widehat W_{z,n}-W_z|
&\le \frac1n\sum_{j=1}^{n}|1-h_j|\,|e_{j-}|
 +\frac{m_n}{nNq}
 \left|\overline G_{z,m_n}-W_z\right|,
\label{eq:restricted-bound-before-event}
\end{align}
where the second term is interpreted as zero when $m_n=0$.
Because the nuisance mean is computed from matched rewards strictly before the deployment round being scored, it is either the constant $1/2$ or the mean of a prefix of length $m_{j-}$ of the same pre-generated matched-reward stream.
On $\mathcal E_{\mathrm{match}}$, for every scored window, simultaneously,
\[
|e_{j-}|\le\epsilon(m_{j-}),\qquad |\overline G_{z,m_n}-W_z|\le\epsilon(m_n) \quad\text{when }m_n\ge1.
\]
Substitution into \eqref{eq:restricted-bound-before-event} gives
\[
|\widehat W_{z,n}-W_z| \le\frac1n\sum_{j=1}^{n}|1-h_j|\epsilon(m_{j-}) +\frac{m_n}{nNq}\epsilon(m_n)=b_{z,n}.
\]
Since $W_z\in[0,1]$, Euclidean projection onto $[0,1]$ cannot increase the distance to $W_z$, and therefore $|\widetilde W_{z,n}-W_z|\le b_{z,n}$.
The reasoning holds for every $z$ and every prefix on the single event \eqref{eq:restricted-sim-event}.
For an unvisited configuration, $|1/2-W_z|\le1/2$ supplies the stated trivial interval.
Finally, $\widehat m_{j-}$, $K_j$, $m_{j-}$, $m_n$, $q$, and $\epsilon(\cdot)$ are functions only of past observed actions, matched rewards, and pre-specified constants.
When used at decision round $t$, the interval is computed from deployments completed strictly before round $t$; it is thus a predecision certificate, not a post-hoc interval.
\end{proof}

\subsubsection{Proof of Corollary~\ref{cor:restricted-regret}}
\begin{proof}
Work on the simultaneous event of Proposition~\ref{prop:certified-special}.
Write $c_{z,t-1}=\widetilde W_{z,t-1}$, $b_{z,t-1}$ for the radius available immediately before decision $t$, and $U_{z,t-1}=\min\{1,c_{z,t-1}+b_{z,t-1}\}$.
The certificate and $W_z\in[0,1]$ imply $U_{z,t-1}\ge W_z$ for every $z$.
Hence, for every noninitialization choice $z_t\in\arg\max_z U_{z,t-1}$,
\begin{align*}
W_{z^\star}-W_{z_t}
&\le U_{z^\star,t-1}-W_{z_t}
\le U_{z_t,t-1}-W_{z_t}\\
&\le c_{z_t,t-1}+b_{z_t,t-1}-W_{z_t}
\le 2b_{z_t,t-1}.
\end{align*}
If upper bounds are used without clipping at $1$, the same argument applies with $U_{z,t-1}=c_{z,t-1}+b_{z,t-1}$.
Every initialization round has regret at most $1$ because $W_z\in[0,1]$.
Adding at most $H$ such rounds proves
\[
\operatorname{Reg}_T
\le H+2\sum_{t=H+1}^{T}b_{z_t,t-1}.
\]
\end{proof}

\paragraph{Scope of the result.}
The confidence event above uses known $q$, independent matched rewards, and a common target-law matched-reward mean.
It does not require an independent target-context audit \emph{within this restricted model}.
In contrast, the general Theorem~\ref{thm:main} requires separately valid target-action, context-transport, exposure, and dependence budgets.
In particular, this restricted model does not establish matching optimal regret--inference or identification--deployment bounds for the general slow-mixing class of Theorem~\ref{thm:frontier}.

%% file: appendix/20_diagnostic_algorithms.tex
\subsection{Diagnostic Algorithms}\label{app:diagnostic-algorithms}
\begin{algorithm}[H]
\caption{Runtime diagnostics and certificate validation for candidate configuration $z$}
\label{alg:diagnostics}
\begin{algorithmic}[1]
\Require Post-burn-in score windows; a dependence-calibration procedure; target-law reference data or independent audit bounds; and the identification, exposure, transport, and drift bounds required by Theorem~\ref{thm:main}.
\State Compute clipped scores $Y_r(z)=\widehat\psi_z(O_r)$ on post-burn-in windows.
\State Estimate autocorrelations $\widehat\rho_{\ell,t}(z)$ up to lag $L_t$.
\State Compute the plug-in effective sample size $\widehat n_{\mathrm{eff},t}(z)$ using Eq.~\eqref{eq:app-neff}.
\State Compute the operational shrinkage proxy $\widetilde n_{\mathrm{eff},t}(z)$ using Eq.~\eqref{eq:app-neff-shrinkage}; obtain $\underline n_{\mathrm{eff},t}(z)$ only from a validated lower-information calibration.
\State Partition windows into adjacent blocks and compute block means $\bar\psi_{g,t}(z)$.
\State Compute the empirical excess block-drift diagnostic $\widehat\kappa_t(z)$ using Eq.~\eqref{eq:app-kappa}; validate it as an upper bound on residual mean drift before using it in the radius.
\State Estimate context discrepancies only against an independently calibrated target-law reference, when such a reference is available.
\State Compute the exposure-enrichment sensitivity $\widehat\Delta_{\mathrm{exp},t}(z)$ and the cross-configuration discrepancy diagnostics.
\State Obtain calibrated upper bounds for target-action error, nuisance-product error, clipping, exposure, context transport, and configuration mismatch. Empirical discrepancies alone are not certified upper bounds.
\If{all components required by Theorem~\ref{thm:main} are certified}
    \State Assemble the confidence radius $b_t(z)$ from the certified quantities.
\Else
    \State Mark the candidate's radius as uncertified.
    \State Trigger the applicable refitting or independent-calibration procedure; do not report theorem-backed coverage until the missing condition is established.
\EndIf
\end{algorithmic}
\end{algorithm}

\subsection{Conceptual schematics}
\label{app:conceptual-schematics}
The schematic contrasts raw-reward and certified causal-value selection; it is not an empirical result.
\begin{figure}[!htbp]
\centering
\includegraphics[width=0.58\linewidth]{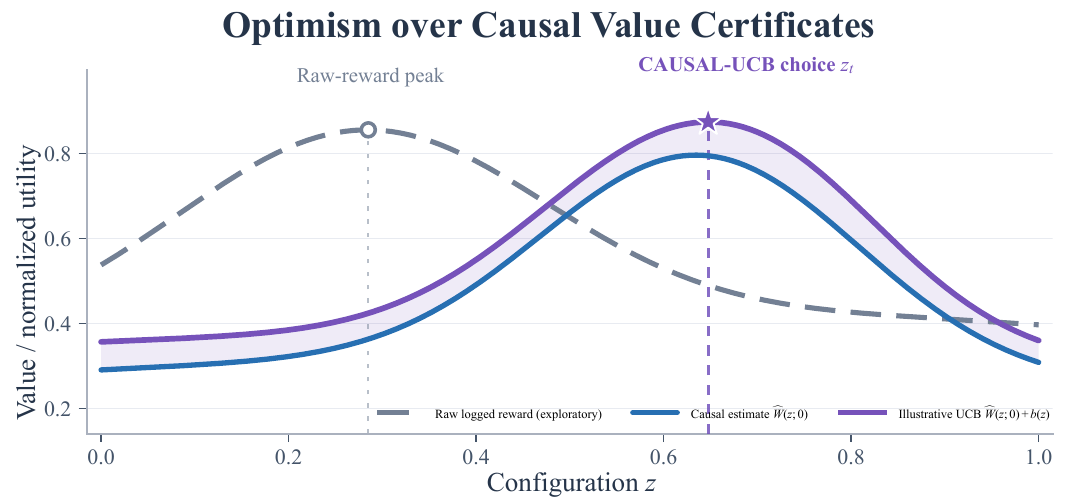}
\caption{Conceptual comparison of raw-reward and certified causal-value maximizers.}
\label{fig:causal-optimism-illustration-app}
\end{figure}

%% file: appendix/21_code_and_evaluation_protocol.tex
\subsection{Code and Evaluation Protocol}\label{app:new-exp-protocol}
\paragraph{Exact source and commands.}
The main controlled evaluation scripts are located in \texttt{experiments/dependencies/certified\_replication} in the accompanying code archive.
Install dependencies from \texttt{requirements.txt}, then execute the following commands from that directory:
\begin{verbatim}
python -m unittest discover -s tests -v
OPENBLAS_NUM_THREADS=1 python fast_compare.py --out results_final \
 --seeds 24 --T 200 --scenarios high_mismatch,low_mismatch
python audit_and_figures.py
\end{verbatim}
The archive includes individual method--seed measurements, certified decision-round logs, seed-pairing metadata, protocol JSON, and a SHA-256 manifest.
To reproduce the main tables and figures, run the commands above without modifying the declared protocol, reference files, or seed list.

\paragraph{Separation of three statistical objects.}
For a deployed configuration $z_t$, target regret is computed with the oracle \emph{no-exploration} target; deployment cost is the expected reward difference under the fixed exploration rate; uncertainty width is the worst-pair contrast diameter built from the valid \emph{predecision} intervals.
The evaluator never uses oracle values for nuisance fitting or selecting an action.
The reported decision certificate uses the predecision radius throughout the evaluation.

\paragraph{Statistical setting.}
The local-interference certificate uses known exposure neighborhoods, fixed exploration, independent matched rewards with specified noise scale, homogeneous targets, and untruncated weights.
Its confidence guarantee is simultaneous over the finite evaluation horizon.
Broader network models can be analyzed through the calibrated target-action, transport, and dependence budgets of Theorem~\ref{thm:main}.
The trace-driven study uses its own semi-synthetic counterfactual construction, as described in Section~\ref{subsec:real-validation}.

%% file: appendix/27_experimental_data_sources_and_evaluation_scope.tex
\subsection{Experimental Data Sources and Evaluation Scope}\label{app:archived-boundary}
The controlled ring benchmark evaluates the predecision finite-sample certificate under known local interference.
METR-LA and FlockLab supply trace-derived network structure and context observations, with model-generated counterfactuals and seed-level uncertainty.
Each study is presented with its corresponding outcome and calibration model.
Supplementary mechanism studies use separate sampling protocols and data sources; their results are reported separately from the primary benchmark.
\begin{table}[!htbp]
\centering
\caption{\textbf{Low-mismatch controlled comparison.} Means over 24 paired seeds and 200 rounds; smaller is better in every numeric column.}
\label{tab:certified-low-mismatch}
\small
\setlength{\tabcolsep}{4pt}
\renewcommand{\arraystretch}{1.0}
\begin{tabular}{@{}lrrrrr@{}}
\toprule
Method & Regret & Gap & ATE error & Width $W$ & Dep. gap \\
\midrule
\rowcolor{CausalBlue}
\textbf{CAUSAL-UCB (ours)} & 1.018 & \textbf{.000} & .0176 & \textbf{.323} & 4.622\\
Certified IPS-UCB & 1.052 & \textbf{.000} & .0597 & .644 & 4.615\\
Legacy DR-UCB & 11.984 & .027 & \textbf{.0057} & 1.105 & 2.545\\
Legacy IPS-UCB & 11.984 & .027 & .0197 & 1.105 & 2.545\\
Raw-UCB & 15.098 & .071 & .0235 & .968 & 1.993\\
Greedy-Raw & \textbf{1.000} & \textbf{.000} & .0291 & 2.000 & 4.626\\
Switchback-OED & 20.000 & .110 & .0231 & .775 & 
\textbf{1.344}\\
\bottomrule
\end{tabular}
\vspace{1pt}
\parbox{.96\linewidth}{\footnotesize
Bold indicates the column minimum (ties included), not statistical significance.
Width $W$ is decision-active only for certified methods; other widths are diagnostic.
Dep. gap is simulator-based.}
\end{table}

\paragraph{Third-party data provenance.}
The METR-LA input, \texttt{metr-la.h5}, is
distributed through the DCRNN project
~\cite{li2018dcrnn_traffic}
(\url{https://github.com/liyaguang/DCRNN}).
Our FlockLab input,
\texttt{2019-07\_FlockLab\_sky.csv},
comes from the July 2019 wireless-link
traces associated with Jacob et al.
~\cite{jacob2019flocklab}
(\url{https://zenodo.org/records/3727526},
Zenodo record v1.4.0).

Both sources are used only as network-derived
context and topology inputs for offline
semi-synthetic experiments; outcomes are
model-generated. Neither the source files
nor derived trace arrays are redistributed
in the code supplement.
Input filenames, archived hashes, and access
links are documented in the supplement's
\texttt{DATA\_PROVENANCE.md}.

%% file: appendix/26_metrics_paired_comparisons_and_pareto.tex
\subsection{Metric Definitions, Paired Comparisons, and Regret--Deployment Pareto Analysis}\label{app:metric-pareto}
\paragraph{Metric inventory and interpretation.}
The frozen controlled benchmark contains 336 method--seed records: two mismatch regimes, seven methods, and 24 paired seeds per regime.
Each record includes cumulative target regret, terminal target-value gap, maximum pairwise ATE error, local audit width, and cumulative deployment gap.
The field \texttt{width\_decision\_certificate} is populated as a decision-active radius only for Certified CAUSAL-UCB and Certified IPS-UCB.
Widths for other methods are local diagnostic widths and do not imply simultaneous target coverage.

\paragraph{Paired-seed comparisons.}
For a metric $M$ and a baseline $b$, define the paired difference
\[
D_s=M_s^{\mathrm{CAUSAL\text{-}UCB}}-M_s^{b}.
\]
For $n$ matched seeds, we report the mean difference and the paired Student-$t$ interval
\[
\bar D \ \pm\ t_{0.975,n-1}\frac{s_D}{\sqrt n},
\]
where $s_D$ is the sample standard deviation of the seed-level paired differences.
A paired interval is not obtained by subtracting two marginal confidence intervals.
Wilson intervals are used for binary full-path coverage outcomes.
Zero-width empirical intervals in degenerate finite samples are not interpreted as zero population uncertainty.

\paragraph{Regret--deployment Pareto order.}
We report cumulative target regret and cumulative expected deployment gap as separate objectives, both minimized.
Method $a$ weakly dominates method $b$ on a given regime's sample means if
\[
\mathrm{Regret}_a\le \mathrm{Regret}_b,\qquad
\mathrm{DeploymentGap}_a\le \mathrm{DeploymentGap}_b,
\]
with at least one strict inequality for strict dominance.
The empirical Pareto set contains methods not dominated by another method on these two sample means.
This set is descriptive: mean non-dominance is not a significance test, and the expected deployment gap is calculated by the simulator rather than directly observed in a real deployment.

\begin{figure}[!htbp]
\centering
\begin{subfigure}[t]{0.32\linewidth}
\centering
\includegraphics[width=\linewidth]{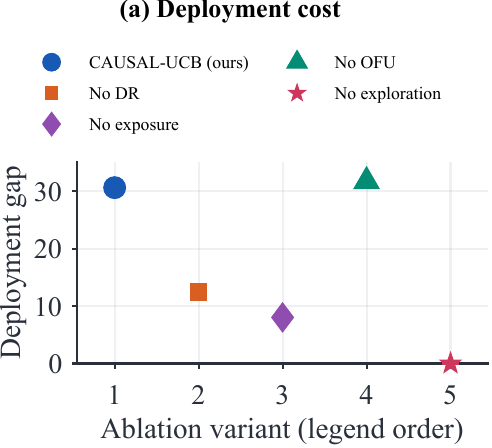}
\end{subfigure}\hfill
\begin{subfigure}[t]{0.32\linewidth}
\centering
\includegraphics[width=\linewidth]{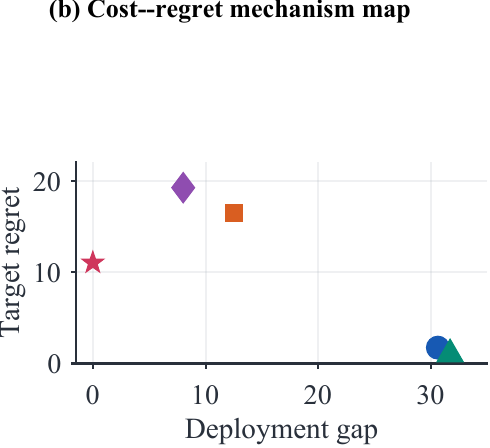}
\end{subfigure}\hfill
\begin{subfigure}[t]{0.32\linewidth}
\centering
\includegraphics[width=\linewidth]{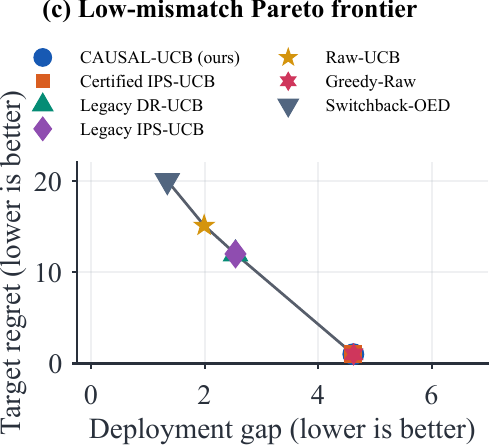}
\end{subfigure}

\caption{\textbf{Supplementary deployment-cost and Pareto diagnostics.}
(a) Deployment gap by ablation variant;
(b) cost--regret map for the same variants;
(c) low-mismatch sample-mean Pareto frontier.
Panels (a)--(b) use the controlled ablation protocol described in the main text
and are not pooled with the five-metric benchmark table.
Panel (c) reports the low-mismatch counterpart of the high-mismatch
frontier in main-text Figure~\ref{fig:exploration-regret}(b).
Regret and simulator-based deployment gap are separate minimized objectives.
The cost--regret and Pareto comparisons are descriptive and do not establish
statistically significant dominance.}
\label{fig:ablation-cost-supp}
\end{figure}

The high-mismatch sample-mean Pareto frontier is shown in
main-text Figure~\ref{fig:exploration-regret}(b);
the low-mismatch counterpart is shown in
Figure~\ref{fig:ablation-cost-supp}(c).
These frontiers compare sample means and do not constitute significance tests.

\paragraph{Interpretation of the controlled comparison.}
Under high mismatch, Certified CAUSAL-UCB, Certified IPS-UCB, and Greedy-Raw represent distinct non-dominated mean trade-offs: the certified DR method has the lowest regret but higher deployment cost, while Greedy-Raw has lower deployment cost but substantially higher regret.
The reference methods with worse regret and higher deployment cost are dominated by at least one of these points in the reported means.
Under low mismatch, the frontier is less discriminating because several methods trade small regret differences against small deployment-gap differences.
The figure therefore complements the five-metric main table without ranking methods through an arbitrary weighted score.

\paragraph{Coverage and certificate width.}
The audit-width column must be read with its provenance.
For the two certified methods, the predecision radius enters the selection rule and has the stated finite-sample interpretation under the model conditions.
For other methods, the recorded width is diagnostic only.
Full-path certificate coverage is evaluated separately in Appendix~\ref{app:seven-term-fullpath-300} and under hidden misspecification in Appendix~\ref{app:seven-term-misspec}; these experiments use different protocols and are not pooled.


%% file: appendix/22_five_level_robustness_and_parameter_sweeps.tex
\subsection{Robustness and Parameter Sweeps}\label{app:formal-sweeps}

\begin{figure}[H]
\centering
\includegraphics[width=.40\linewidth]{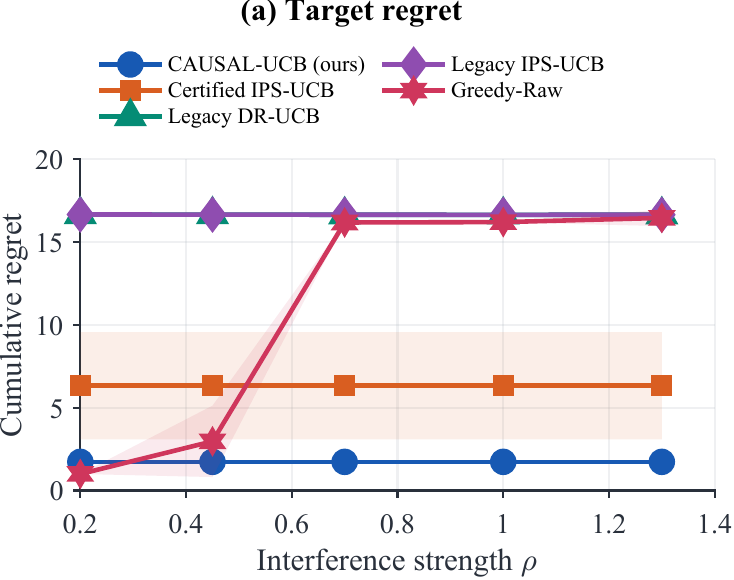}\hfill
\includegraphics[width=.40\linewidth]{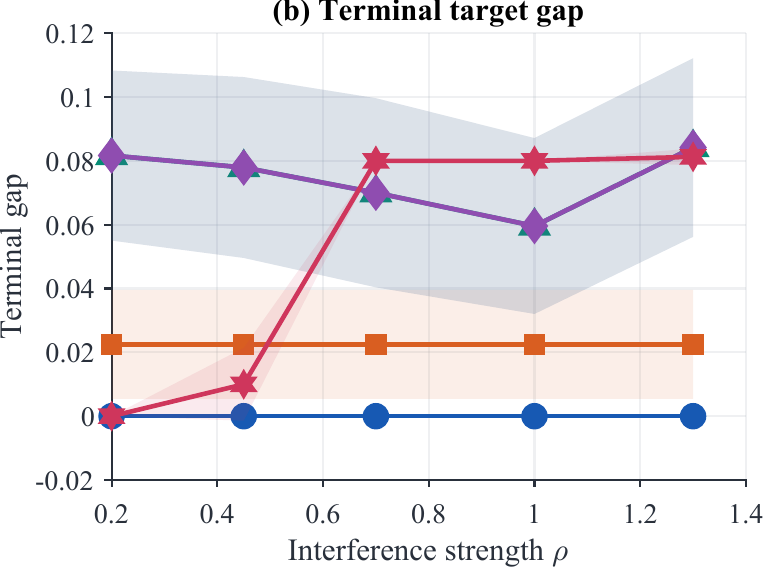}

\par\vspace{3pt}
\includegraphics[width=.40\linewidth]{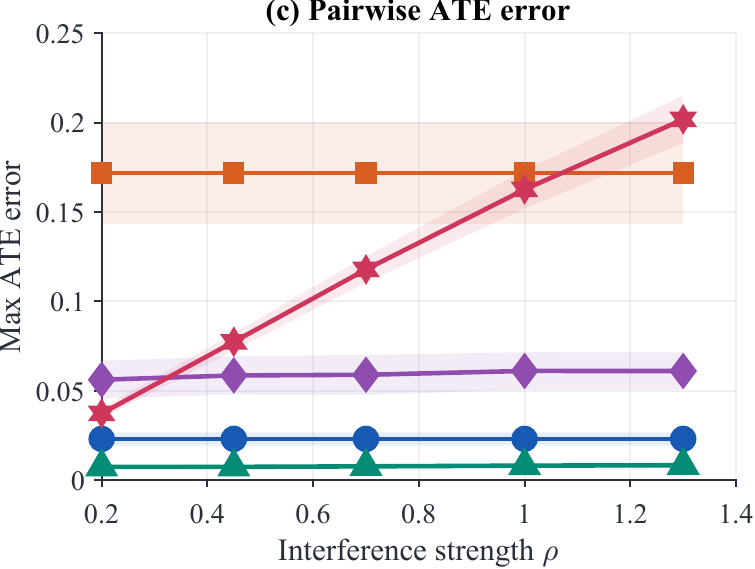}\hfill
\includegraphics[width=.40\linewidth]{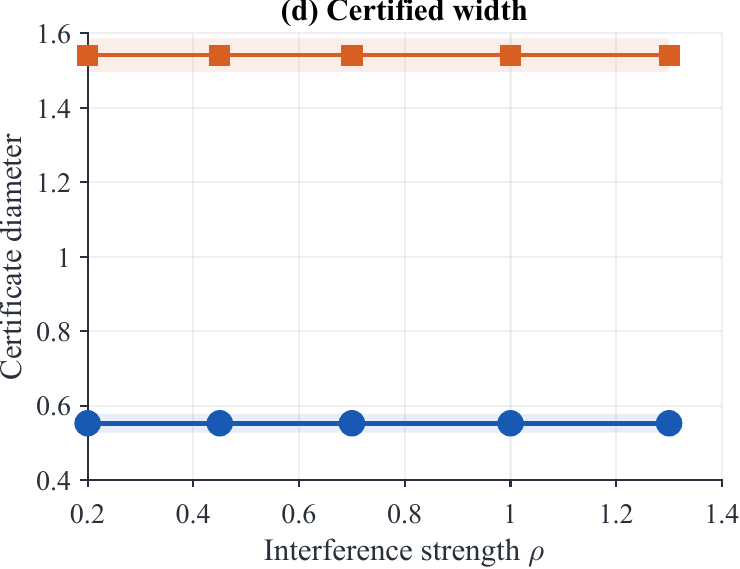}
\caption{Supplementary interference sensitivity across the seven levels shown in the panels.
The four outcomes are cumulative target regret, terminal target-value gap, maximum pairwise ATE error, and decision-active certificate width.
Widths for non-certified methods remain diagnostic and do not imply simultaneous target coverage.}
\label{fig:interference-full}
\end{figure}
\begin{figure}[H]
\centering
\includegraphics[width=.40\linewidth]{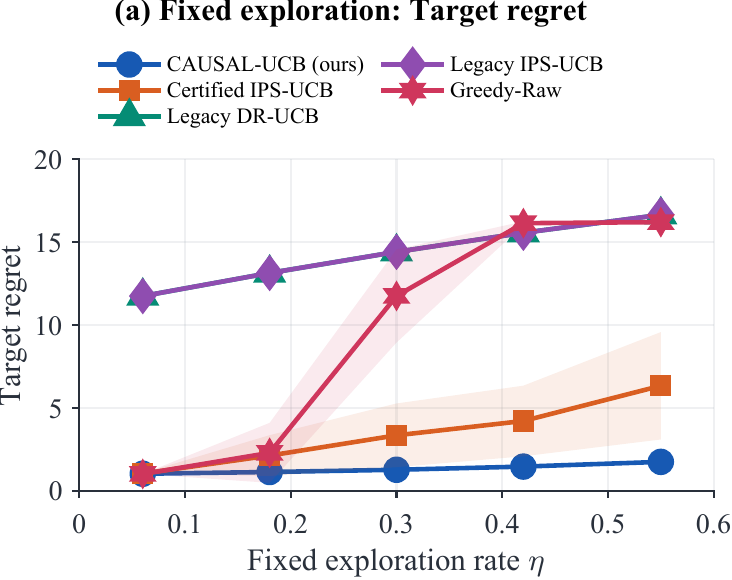}\hfill
\includegraphics[width=.40\linewidth]{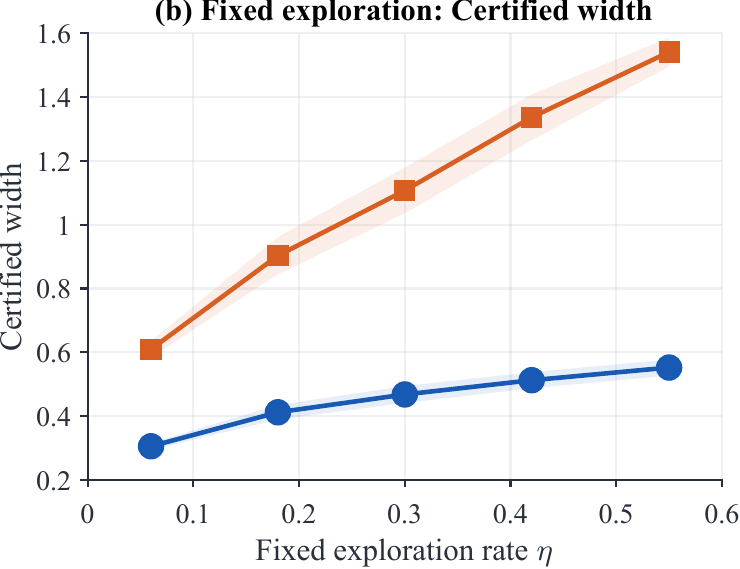}

\par\vspace{3pt}
\includegraphics[width=.40\linewidth]{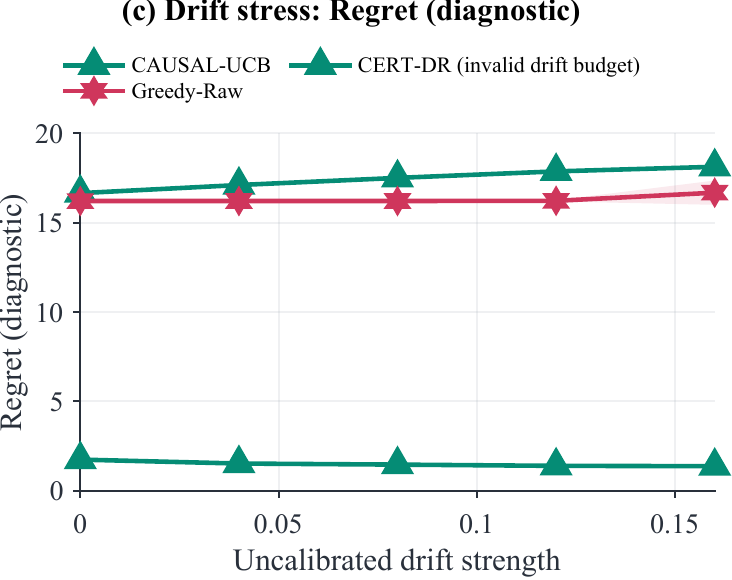}\hfill
\includegraphics[width=.40\linewidth]{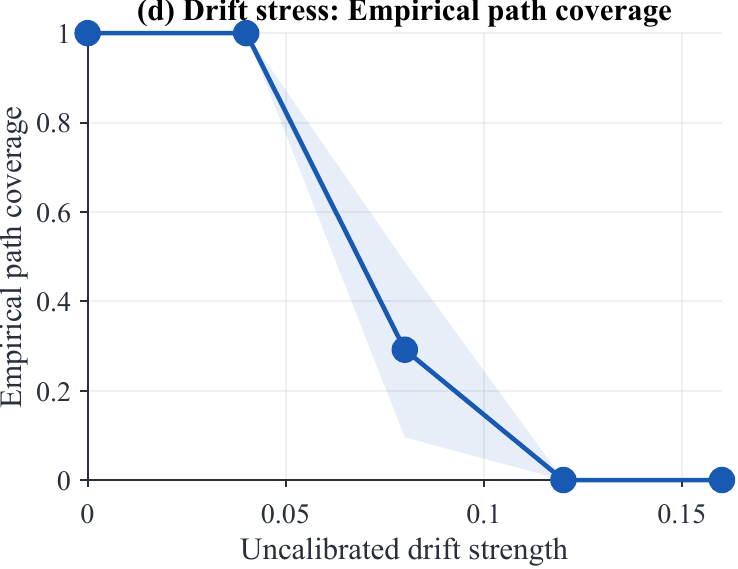}
\caption{Exploration-rate and uncalibrated-drift stress diagnostics.
Panels (a)--(b) report target regret and certificate width across the fixed exploration rates shown.
Panels (c)--(d) report regret and empirical path coverage under deliberately uncalibrated post-burn-in drift.
The drift panels violate the calibrated-drift condition and are descriptive diagnostics, not theorem-certified coverage evidence.}
\label{fig:exploration-drift-stress}
\end{figure}

\paragraph{Detailed robustness evidence.}
The main table reports regret, terminal gap, causal-estimation error, audit width, and deployment gap as separate quantities.
This appendix provides the full interference sensitivity panels and the complementary fixed-exploration and uncalibrated-drift stress diagnostics.
The latter deliberately leave the validated drift regime and are interpreted as failure-boundary evidence, not theorem-certified coverage results.

\begin{figure}[H]
\centering
\includegraphics[width=.32\linewidth]{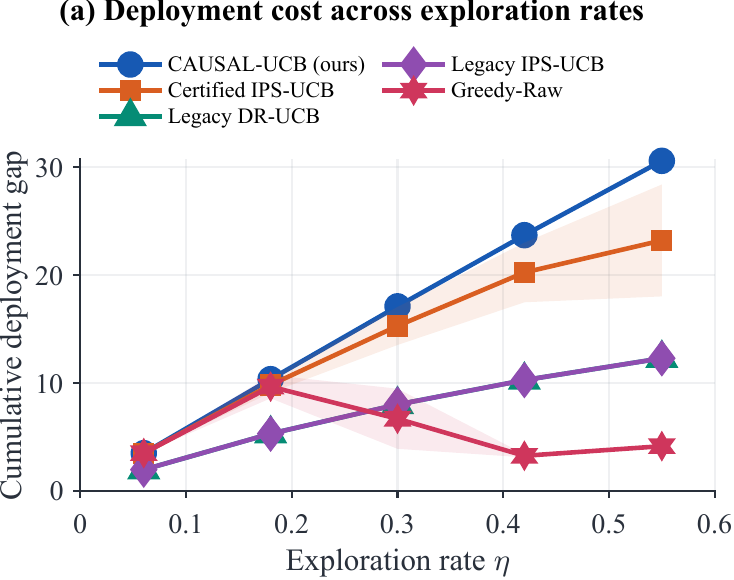}\hfill
\includegraphics[width=.32\linewidth]{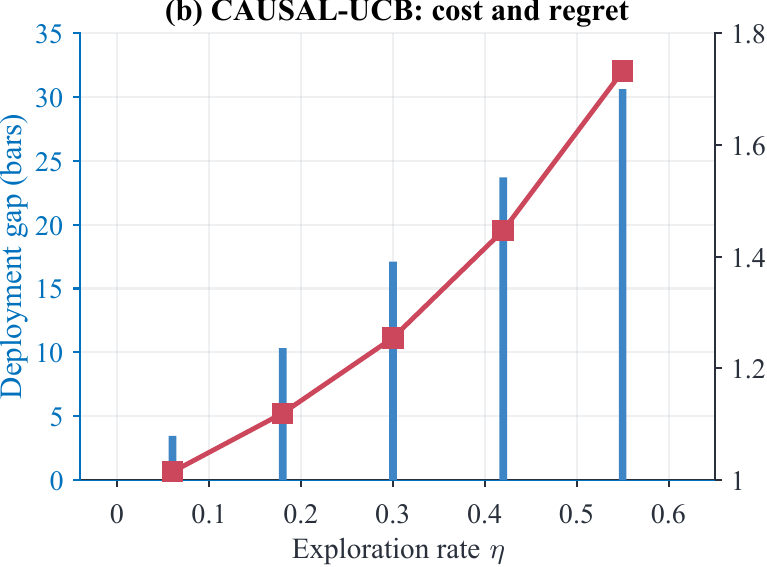}\hfill
\includegraphics[width=.32\linewidth]{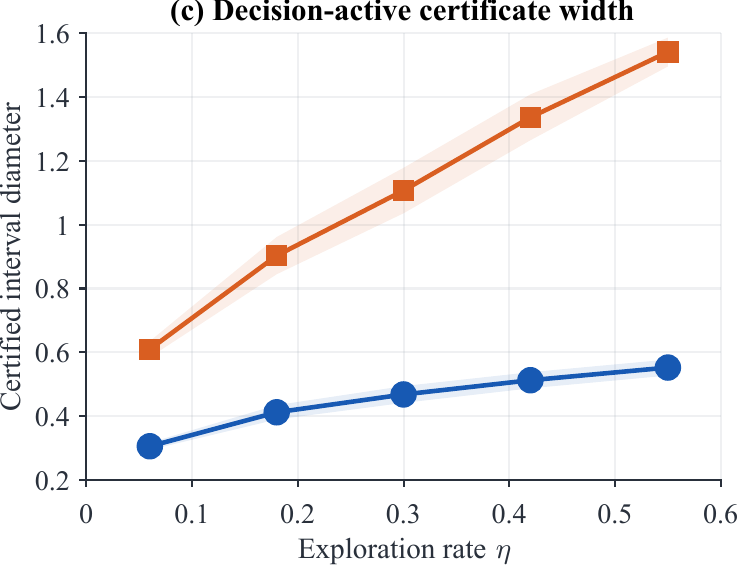}

\caption{\textbf{Measurement-exploration cost and certificate width.}
(a) Cumulative deployment gap across exploration rates and methods;
(b) Certified DR deployment gap (bars) and regret (line) on separate axes;
(c) Decision-active certificate width across fixed exploration rates.
Panels (a)--(b) correspond to the controlled fixed-rate sweep, not the minimax lower-bound construction.
Panel (c) reports a radius diagnostic for the displayed certified methods;
width alone does not establish coverage, which remains conditional on the calibration requirements of Theorem~\ref{thm:main}.}
\label{fig:exploration-cost-panels}
\end{figure}

\paragraph{Exploration-rate and deployment-cost diagnostics.}
Figure~\ref{fig:exploration-cost-panels} expands the measurement-cost comparison accompanying main-text Figure~\ref{fig:exploration-regret}(a) and Figure~\ref{fig:exploration-regret}(b).
The exploratory logging rate is varied in a separate controlled sweep; deployment gap denotes measurement-induced utility loss and is distinct from target regret.

%% file: appendix/23_trace_driven_semi_synthetic_comparisons_as_grouped_bars.tex
\subsection{Trace-Driven Semi-Synthetic Comparisons}
\label{app:trace-comparisons}

\begin{figure}[H]
\centering
\includegraphics[width=.48\linewidth]{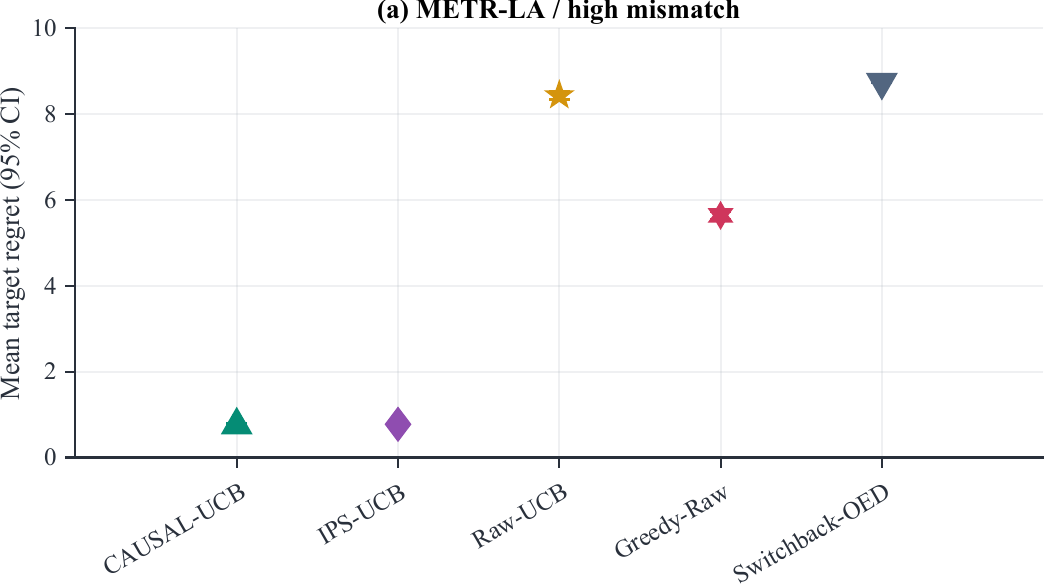}\hfill
\includegraphics[width=.48\linewidth]{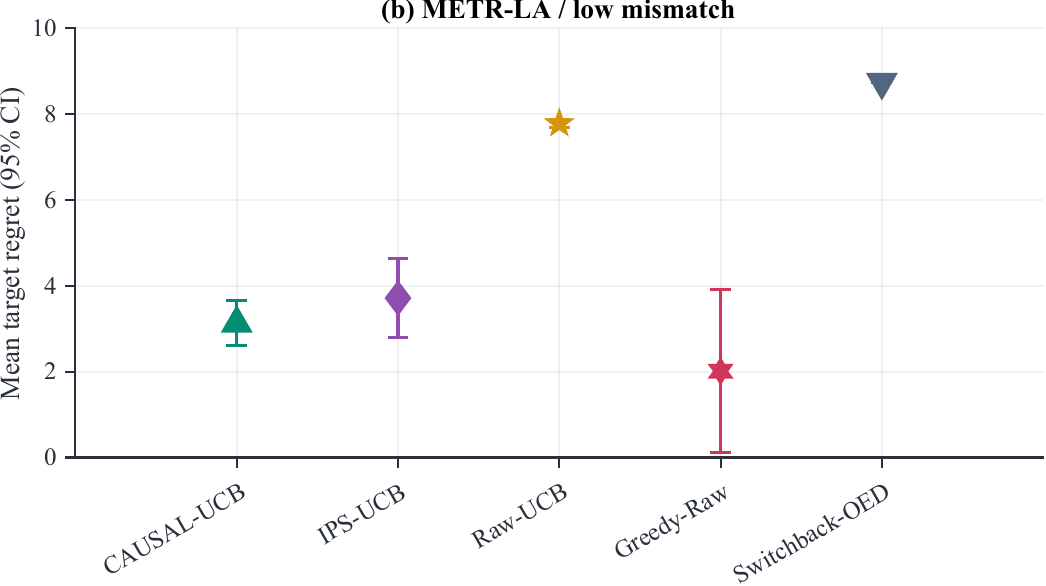}

\par\medskip
\includegraphics[width=.48\linewidth]{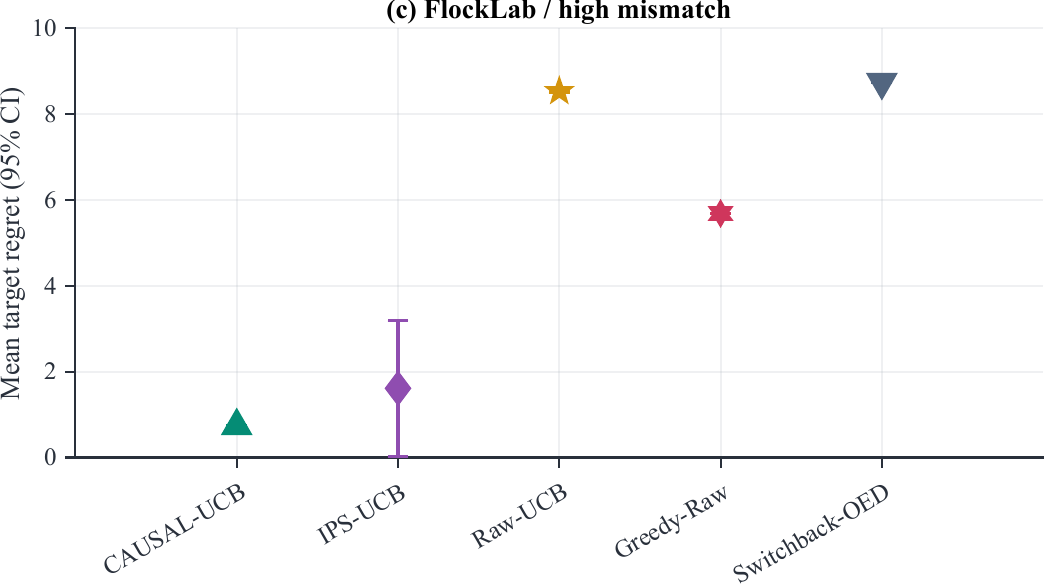}\hfill
\includegraphics[width=.48\linewidth]{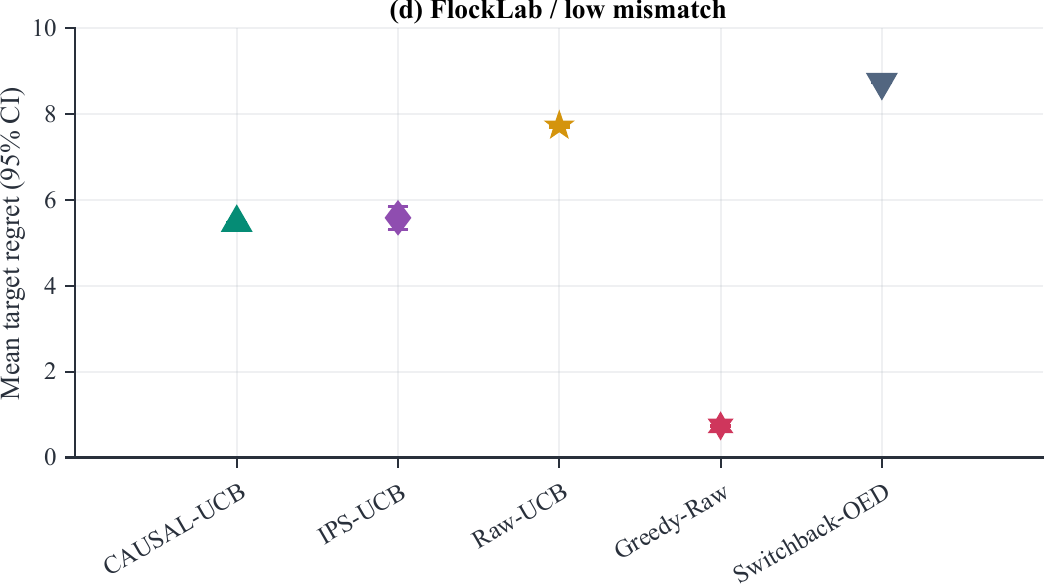}
\caption{\textbf{Trace-driven semi-synthetic comparisons
by dataset and mismatch regime.}
Panels (a)--(d) show METR-LA/high mismatch,
METR-LA/low mismatch, FlockLab/high mismatch,
and FlockLab/low mismatch, respectively.
Whiskers are 95\% Student-$t$ intervals across
eight independent seeds.
The high-mismatch cells are also summarized in
Table~\ref{tab:semi-syn-high}; this figure adds
the low-mismatch cells.
These runs are \emph{not} live deployments and are
\emph{not} certified by the homogeneous ring-model
confidence theorem.}
\label{fig:trace-bars}
\end{figure}

\paragraph{Protocol and comparison.}
Figure~\ref{fig:trace-bars} extends
Table~\ref{tab:semi-syn-high} by reporting cumulative
target regret under both high- and low-mismatch regimes.
The semi-synthetic replays use METR-LA and FlockLab topology/context traces with model-generated
counterfactual outcomes, spanning 180 decision rounds
and eight independent seeds per condition.
The four panels separate dataset-specific effects from
the influence of logging--target mismatch, while the
95\% Student-$t$ intervals quantify seed-level
variability.

\paragraph{Regime-dependent behavior.}
Under high mismatch, CAUSAL-UCB achieves lower target
regret than IPS-UCB on both METR-LA and FlockLab,
consistent with the benefit of correcting exploratory
measurements toward the deployment target.
Under low mismatch, however, Greedy-Raw achieves
lower regret on both traces, showing that the
advantage of causal correction is not uniform across
measurement regimes.
These comparisons characterize empirical transfer
and its limitations; they neither establish
identification under arbitrary network interference
nor extend the restricted ring-model certificate
to trace-driven environments.

%% file: appendix/24_seed_level_distributions_and_pareto_summary.tex
\subsection{Seed-Level Outcomes and Sensitivity Summaries}
\label{app:distributions-summary}
\begin{figure}[!htbp]
\centering
\includegraphics[width=.455\linewidth]{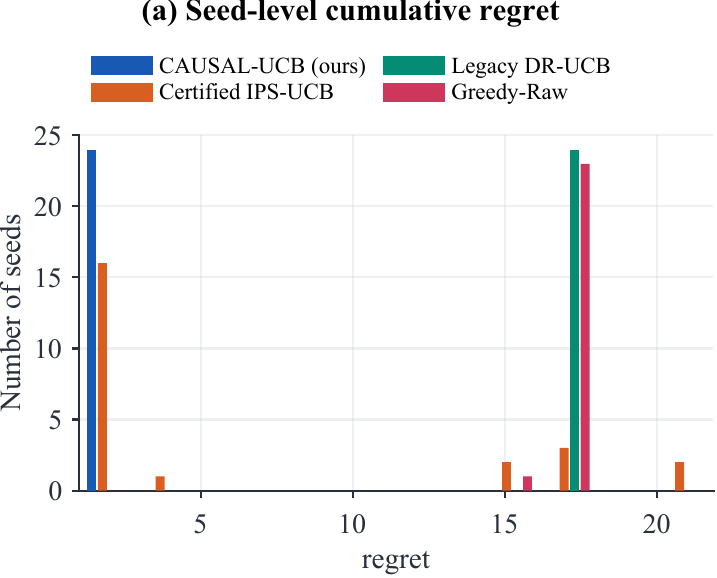}\hfill
\includegraphics[width=.455\linewidth]{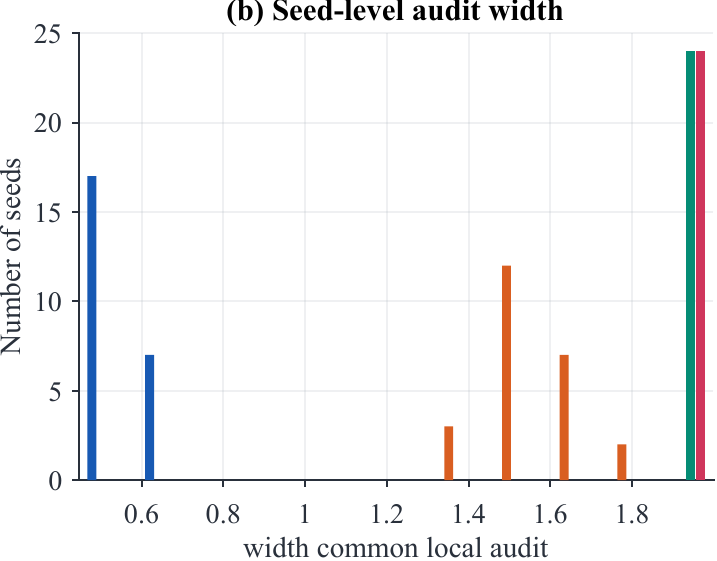}

\par\vspace{3pt}
\includegraphics[width=.455\linewidth]{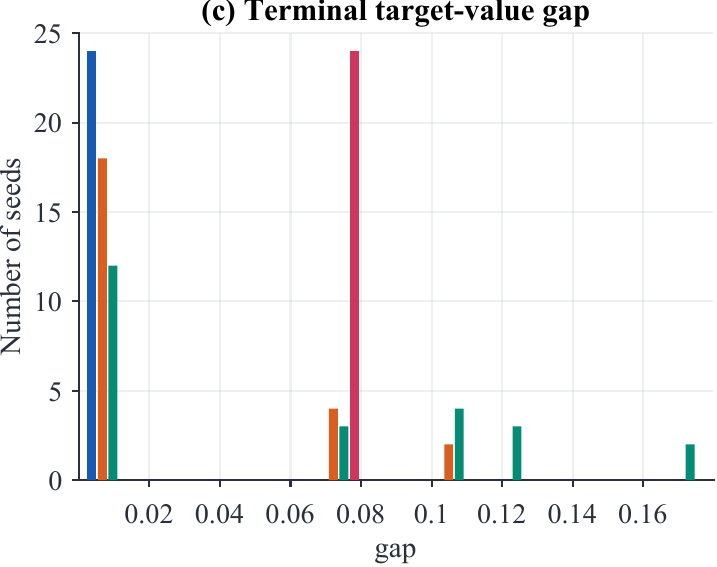}\hfill
\includegraphics[width=.455\linewidth]{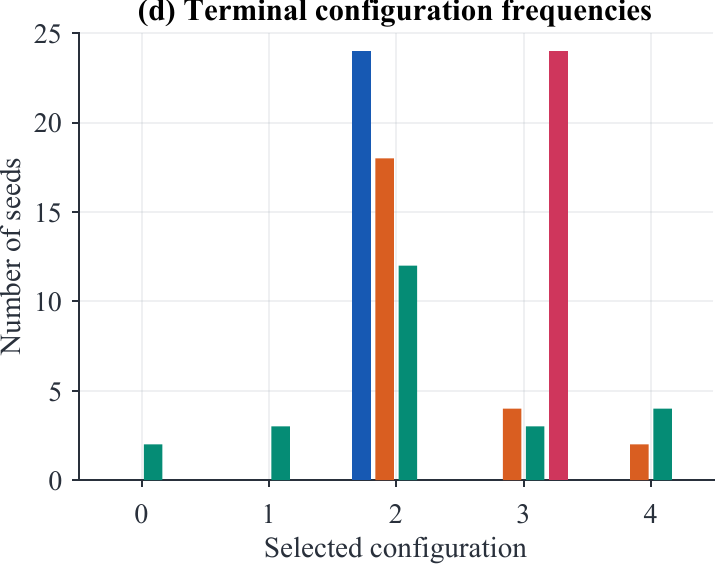}
\caption{\textbf{Seed-level outcomes under high mismatch.} Panels (a)--(d) report cumulative target regret, local audit width, terminal target-value gap, and final selected-configuration frequency.
All panels use the frozen 24 paired seeds.
Widths for non-certified methods are diagnostic only, and the final panel reports configuration IDs rather than unobserved true-value ranks.}
\label{fig:seed-distributions}
\end{figure}
The high-mismatch sample-mean Pareto frontier is shown in main-text Figure~\ref{fig:exploration-regret}(b).
Its low-mismatch counterpart is reported in Figure~\ref{fig:ablation-cost-supp}(c) in Appendix~\ref{app:metric-pareto}; neither sample-mean frontier is a significance test.
Only the two certified methods have a decision-active audit radius.

These known-local-exposure sensitivity studies use 24 independent paired seeds per setting and $T=200$.
The number and values of sweep levels are metric-specific and are shown in each panel; they are not forced to a common grid.
Regret, ATE error, and deployment gap are summarized separately, with seed-level Student-$t$ intervals where shown.

\begin{figure}[!htbp]
\centering
\includegraphics[width=.32\linewidth]{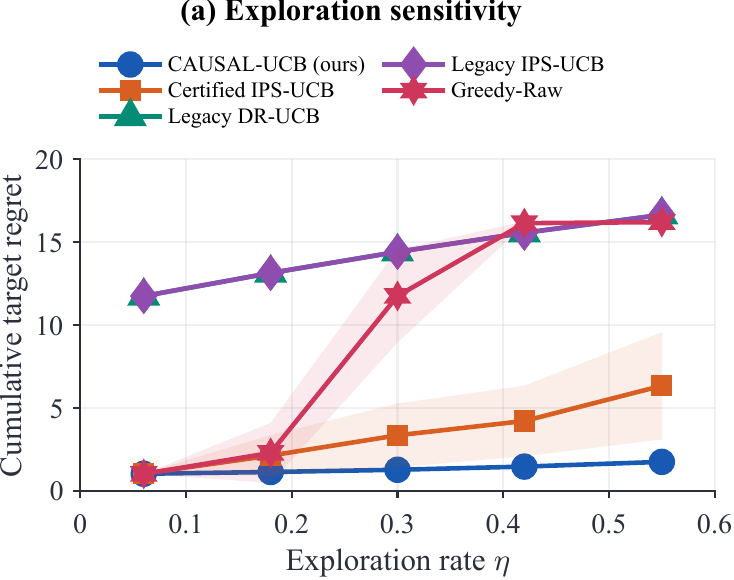}\hfill
\includegraphics[width=.32\linewidth]{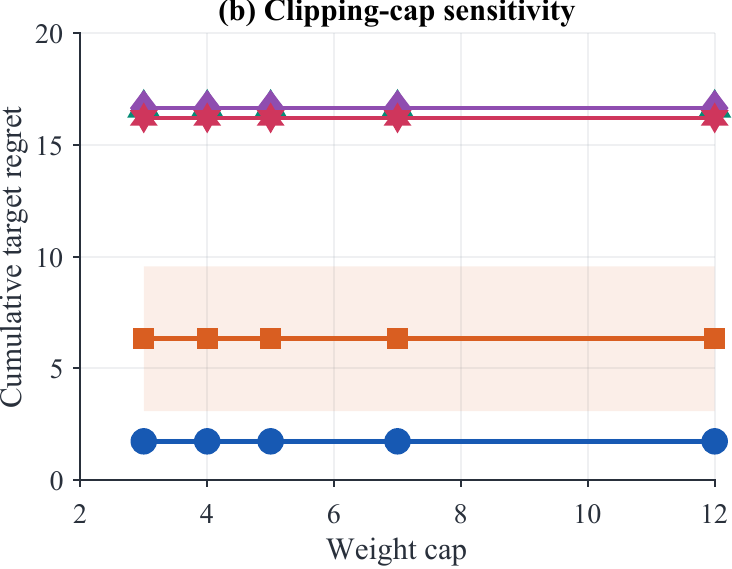}\hfill
\includegraphics[width=.32\linewidth]{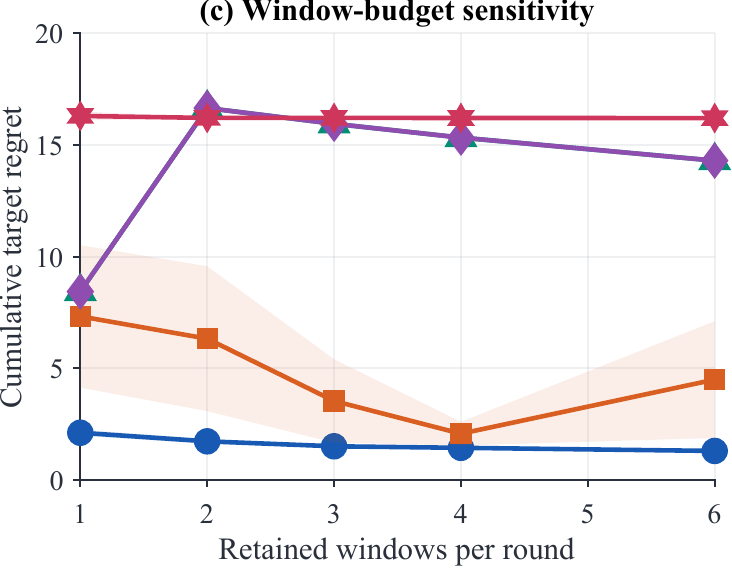}
\caption{Fixed-rate sensitivity sweeps: (a) exploration rate, (b) importance-weight clipping cap, and (c) retained-window budget. A nonzero clipping bias must be accounted for before an interval is treated as certified.}
\label{fig:fixed-rate-sensitivity}
\end{figure}

\begin{figure}[!htbp]
\centering
\includegraphics[width=.32\linewidth]{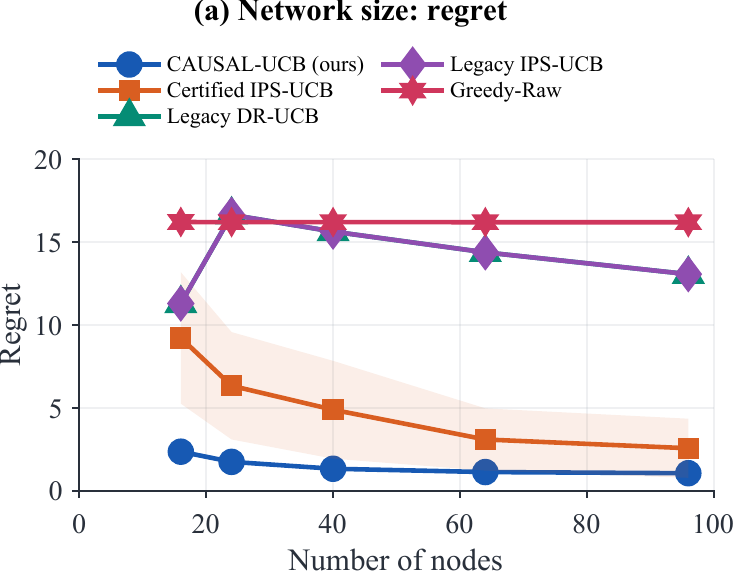}\hfill
\includegraphics[width=.32\linewidth]{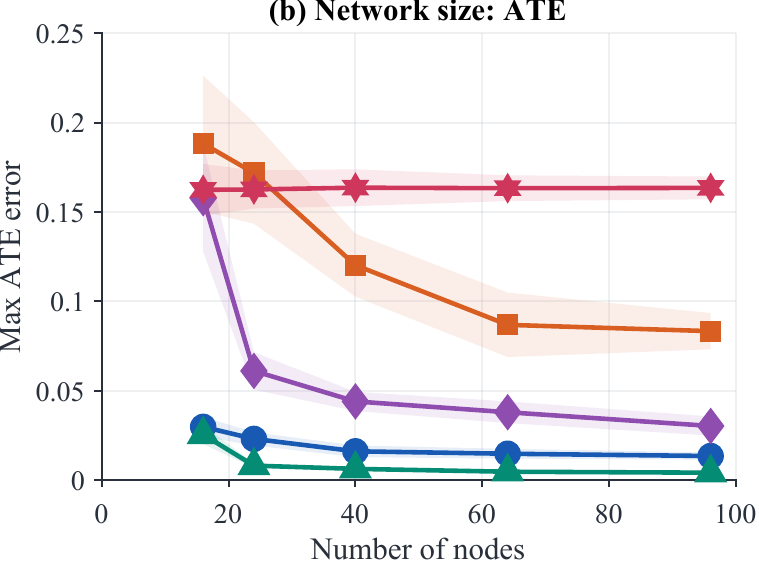}\hfill
\includegraphics[width=.32\linewidth]{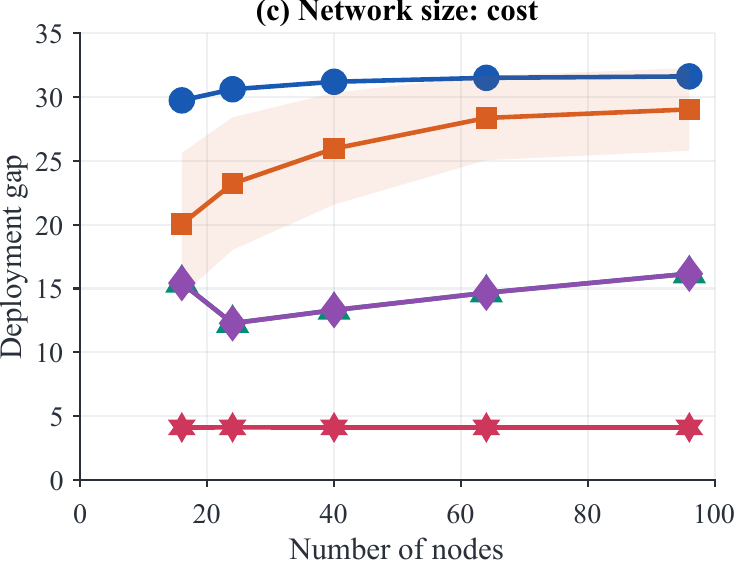}
\caption{Network-size sensitivity across the displayed node counts. Panels report target regret, maximum pairwise ATE error, and deployment gap, respectively.}
\label{fig:network-size-sensitivity}
\end{figure}

\begin{figure}[!htbp]
\centering
\includegraphics[width=.32\linewidth]{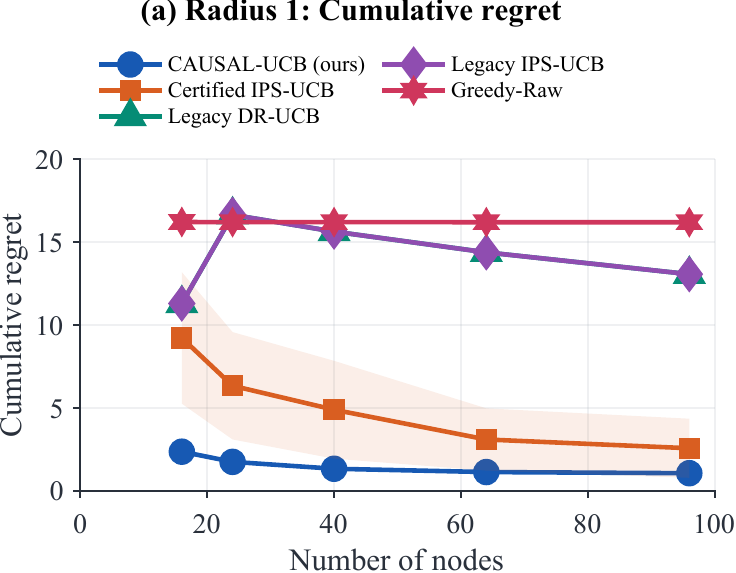}\hfill
\includegraphics[width=.32\linewidth]{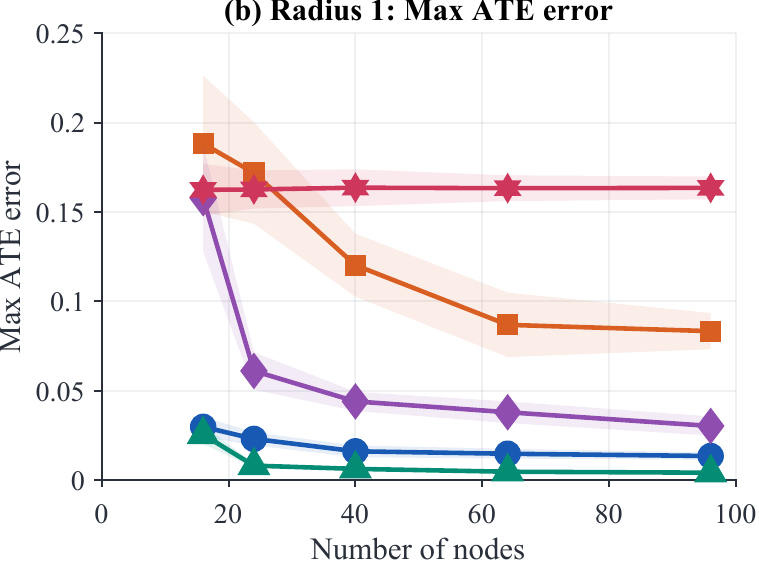}\hfill
\includegraphics[width=.32\linewidth]{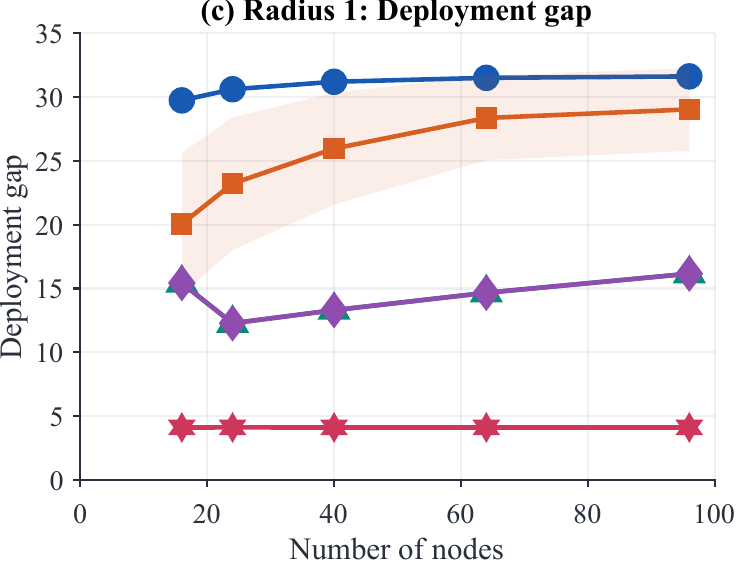}

\par\vspace{3pt}
\includegraphics[width=.32\linewidth]{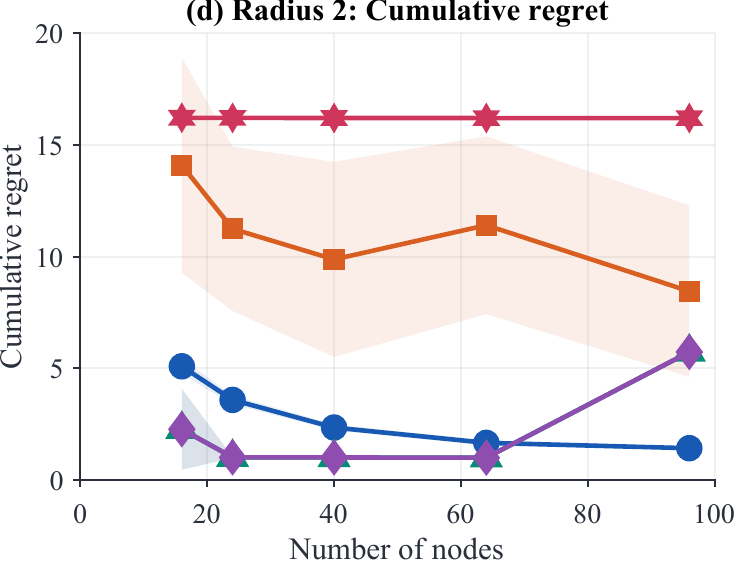}\hfill
\includegraphics[width=.32\linewidth]{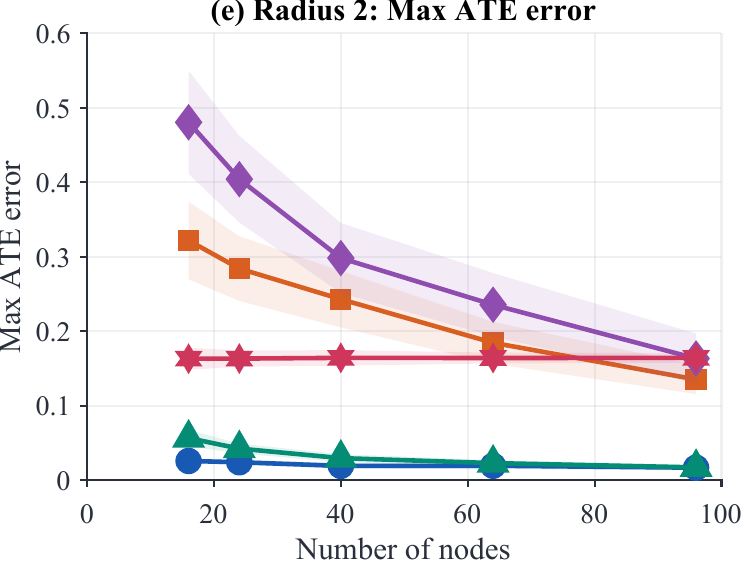}\hfill
\includegraphics[width=.32\linewidth]{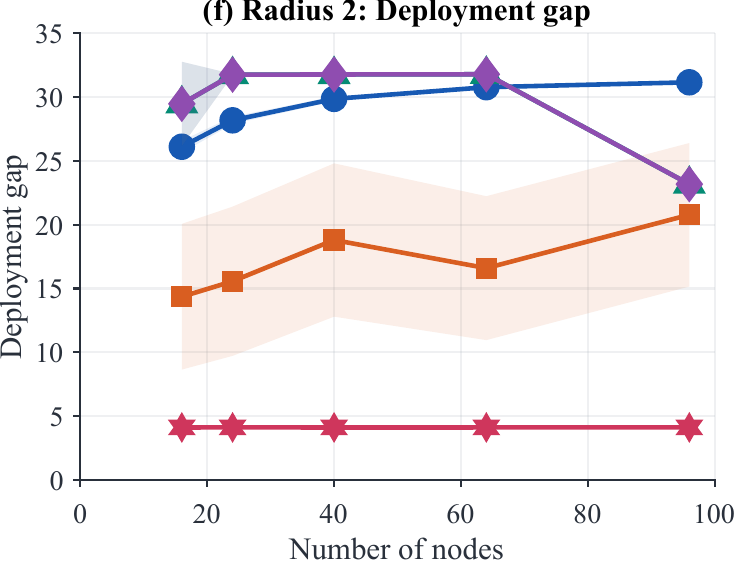}
\caption{Known-radius scalability sweep. Panels (a)--(c) report target regret, maximum pairwise ATE error, and deployment gap for radius $r=1$; panels (d)--(f) report the same metrics for radius $r=2$.}
\label{fig:radius-scalability}
\end{figure}

\begin{figure}[!htbp]
\centering
\includegraphics[width=.32\linewidth]{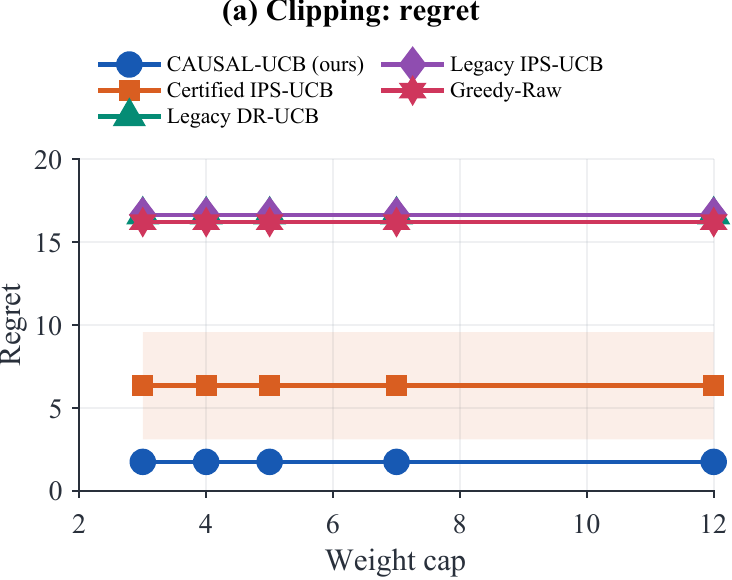}\hfill
\includegraphics[width=.32\linewidth]{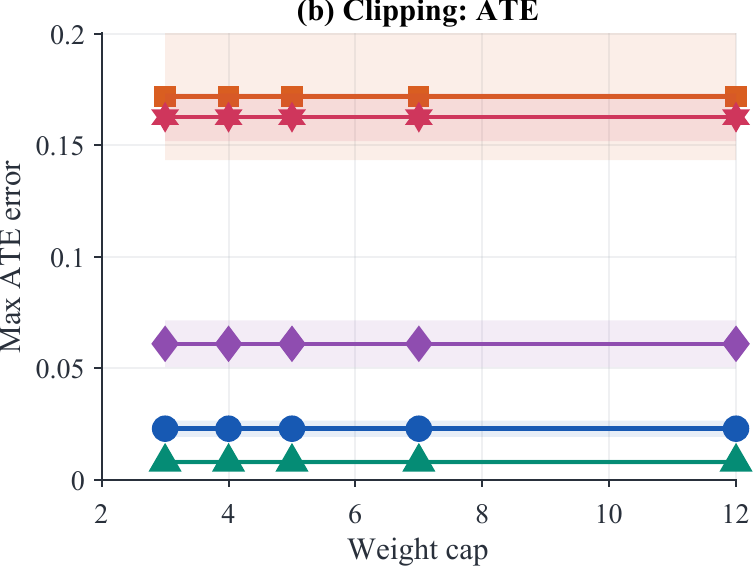}\hfill
\includegraphics[width=.32\linewidth]{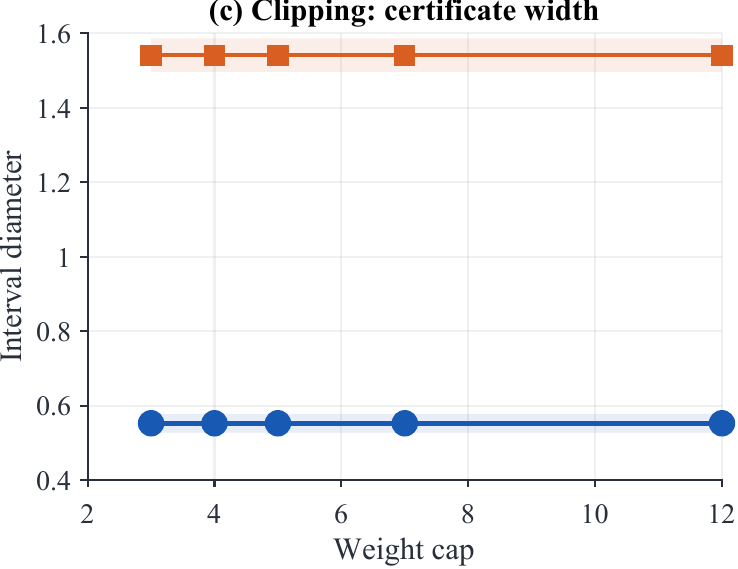}
\caption{Clipping-cap sensitivity across the displayed cap values. Panels report target regret, maximum pairwise ATE error, and certificate width. Clipping without a valid bias allowance is a stress diagnostic, not a certified guarantee.}
\label{fig:clipping-cap-sensitivity}
\end{figure}

\begin{figure}[!htbp]
\centering
\includegraphics[width=.32\linewidth]{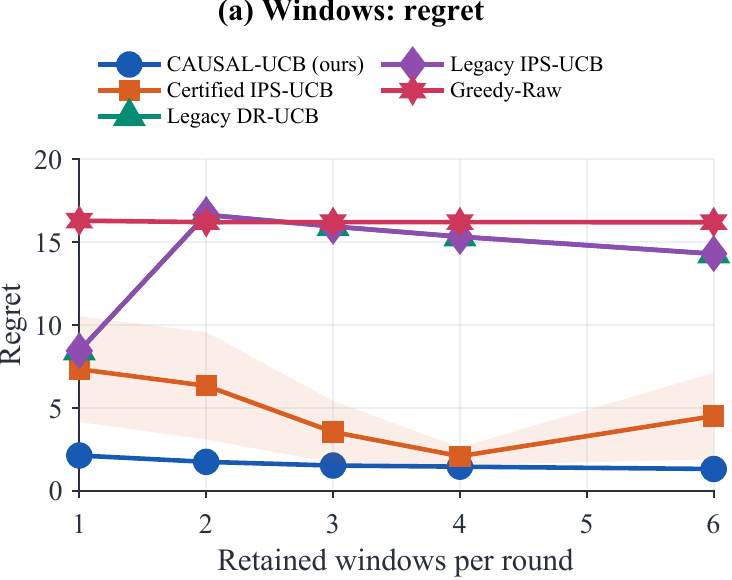}\hfill
\includegraphics[width=.32\linewidth]{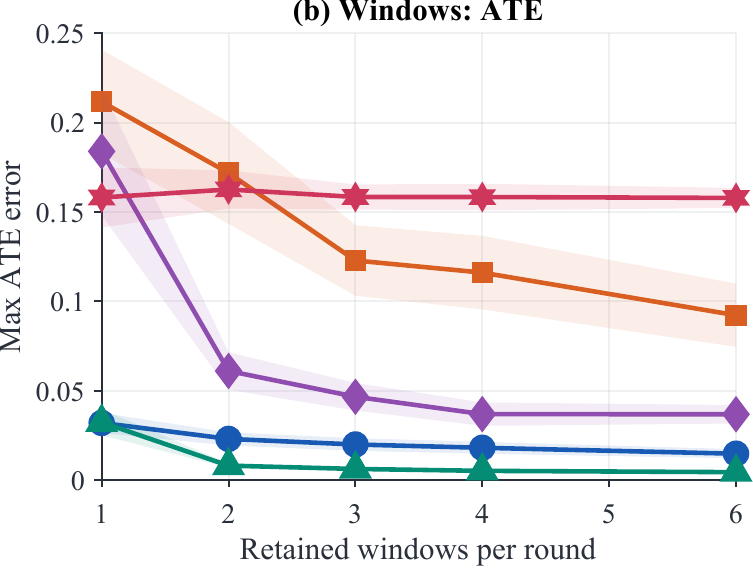}\hfill
\includegraphics[width=.32\linewidth]{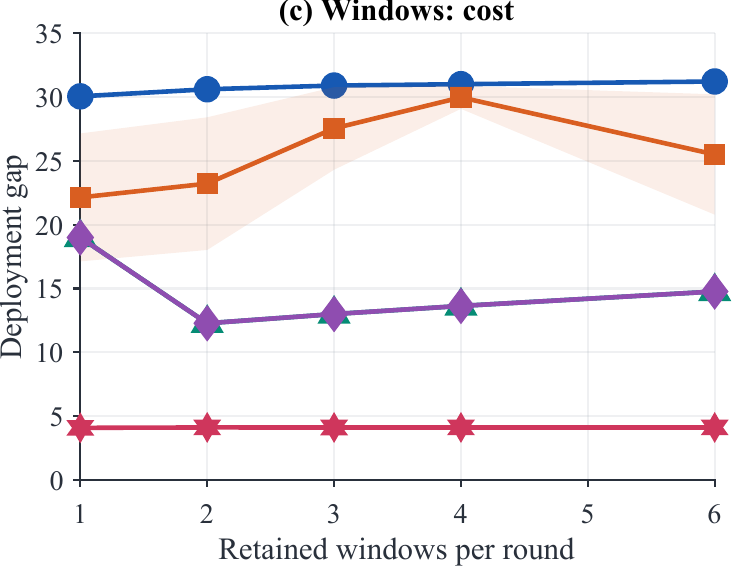}
\caption{Retained-window-budget sensitivity. Panels report target regret, maximum pairwise ATE error, and deployment gap across the window counts shown.}
\label{fig:window-budget-sensitivity}
\end{figure}
\paragraph{Scope of context-shift and annealing diagnostics.}
When the generator adds uncalibrated drift or changes $\eta$ over time, the fixed-exploration stationary proposition is inapplicable; measured coverage under these modifications is descriptive, not a valid extension of Theorem~\ref{thm:main}.
These diagnostics are reported separately from the certified main benchmark and do not alter its theorem assumptions.

\subsubsection{Auditable Contextual Support Extension}\label{app:joint-support-geometry}

\begin{figure}[!htbp]
\centering
\begin{subfigure}[t]{0.48\linewidth}
\centering
\includegraphics[width=\linewidth]{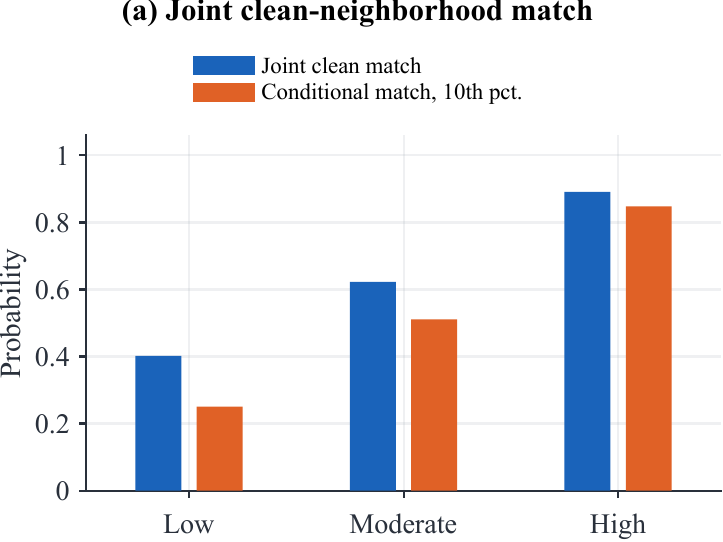}
\end{subfigure}\hfill
\begin{subfigure}[t]{0.48\linewidth}
\centering
\includegraphics[width=\linewidth]{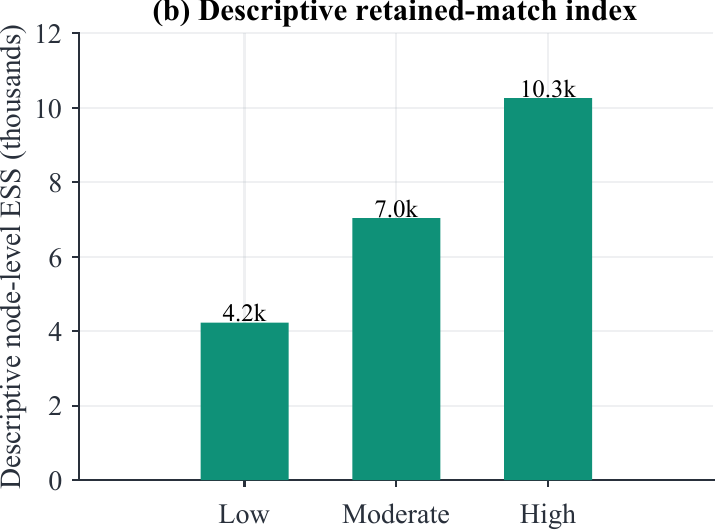}
\end{subfigure}
\caption{\textbf{Supplementary overlap quantification for Figure~\ref{fig:joint-support-geometry}.} Panel (a) reports empirical joint clean-neighborhood match probability and the conditional 10th-percentile match probability; panel (b) reports descriptive node-level IPW effective counts across overlap regimes.
These counts are not the dependence-calibrated effective sample size used in Theorem~\ref{thm:main}.}
\label{fig:joint-support-summary}
\end{figure}

Figure~\ref{fig:joint-support-geometry} is a new, separate mechanism experiment, not an unchanged instance of the certified homogeneous ring generator.
We preserve the fixed 24-node radius-one ring, the five-configuration setup, 24 seeds and paired target/logging sampling, but \emph{add} context-dependent measurement flip probabilities and an explicit logging-induced context response.
Let $X_i=(X_{1i},X_{2i})$ denote the stationary pre-exploration latent context, generated from seeded, spatially smoothed Gaussian covariates.
Conditional on $X$, measurement flips are independent:
\[
 p_i=\eta\,\sigma(0.95X_{1i}-0.65X_{2i}),\qquad A_i\sim\mathrm{Ber}(p_i),\qquad E_i=0.6A_i+0.2A_{i-1}+0.2A_{i+1}.
\]
The unperturbed target has $A_i=E_i=0$ and $C_i^0=X_i$.
Logging operates with
\[
 C_{1i}^{\eta}=X_{1i}+0.90E_i+0.16\eta X_{2,i-1},\quad C_{2i}^{\eta}=X_{2i}-0.62E_i+0.26\eta X_{1,i+1}.
\]
Rewards are generated by clipping $0.62-0.60E_i+0.075 C_{1i}-0.060 C_{2i}+\zeta_i$ to $[0,1]$, with the same independent noise realization in target and logging simulator modes.
The full-neighborhood conditional clean event probability is
\[
 q_i(X)=\prod_{j\in\{i-1,i,i+1\}}(1-p_j),\qquad I_i=\mathbb 1\{A_{i-1}=A_i=A_{i+1}=0\}.
\]
For each $\eta\in\{0.55,0.30,0.08\}$ we sample 24 seeds, 20 spaced rounds and all 24 nodes, yielding 11,520 logging and 11,520 paired target records.
The empirical clean rate is $\sum I_i/n$; the reported $q_{10\%}$ is the empirical tenth percentile of the true conditional $q_i$; the displayed IPW information index is $(\sum I_i/q_i)^2/(\sum (I_i/q_i)^2)$.
This index is \emph{descriptive across dependent ring nodes}, not the certified network-window effective sample size of Theorem~\ref{thm:main}.
The paired target mode is available solely inside the simulator; no deployment algorithm observes its counterfactual data.
Because these conditional propensities and context shifts violate assumptions of the previous certified fixed-$\eta$ submodel, \emph{no valid Theorem~\ref{thm:main} certificate or regret conclusion is asserted} for this extension.
The figure is a mechanistic visualization of why context transport must be independently controlled even with positive conditional action support.


%% file: appendix/25_matched_log_mechanism_studies.tex
\subsection{Matched-Log Mechanism Studies}\label{app:matched-log-studies}
Supplementary mechanism evidence comprises a 60-seed, 1,000-round factorial study and a matched-log DR-by-exposure study, alongside the primary controlled benchmark.
Their sampling protocols and data sources are documented separately; results from distinct protocols are not pooled.
\begin{figure}[!htbp]
\centering
\includegraphics[width=.32\linewidth]{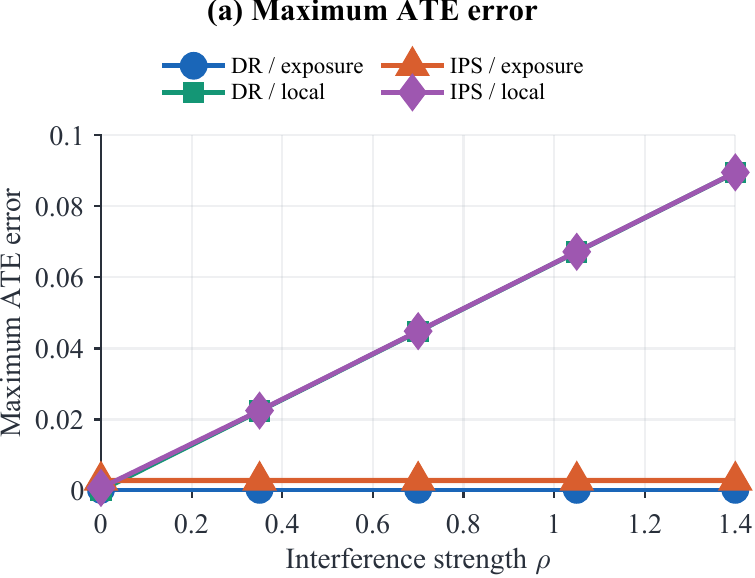}\hfill
\includegraphics[width=.32\linewidth]{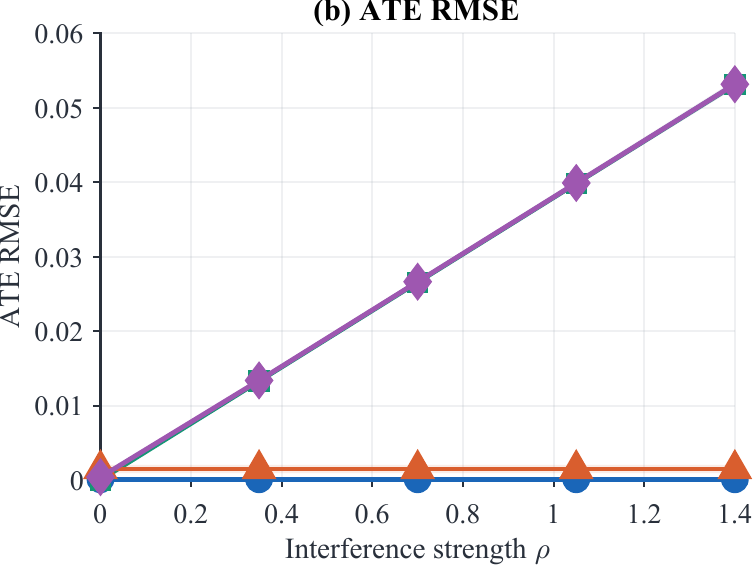}\hfill
\includegraphics[width=.32\linewidth]{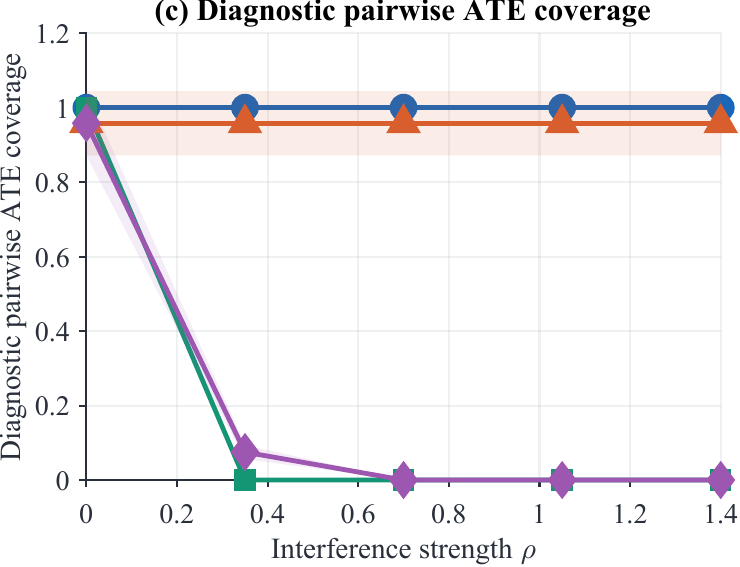}
\caption{Matched-log DR/IPS-by-exposure comparison across the interference levels shown.
Panels report maximum pairwise ATE error, ATE RMSE, and diagnostic pairwise ATE coverage.
The local representation omits neighboring actions; the displayed coverage is a diagnostic and is not the general Theorem~\ref{thm:main} guarantee.}
\label{fig:matched-log-dr-exposure}
\end{figure}
\begin{figure}[!htbp]
\centering
\includegraphics[width=.48\linewidth]{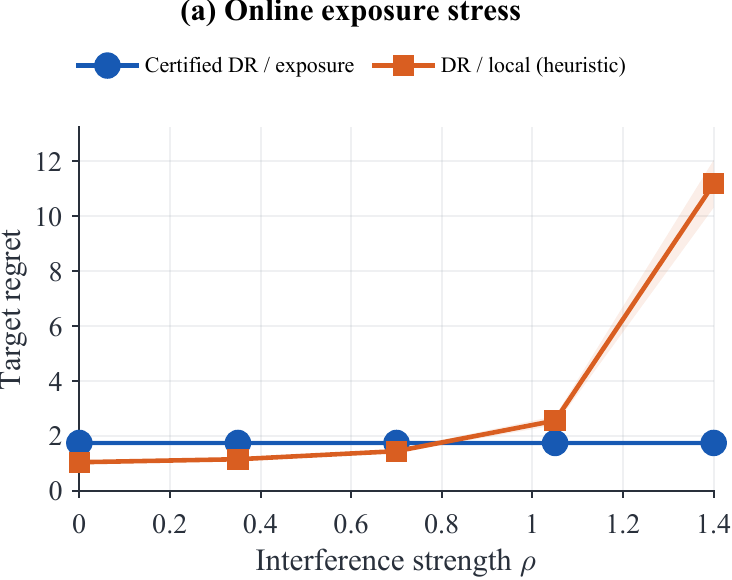}\hfill
\includegraphics[width=.48\linewidth]{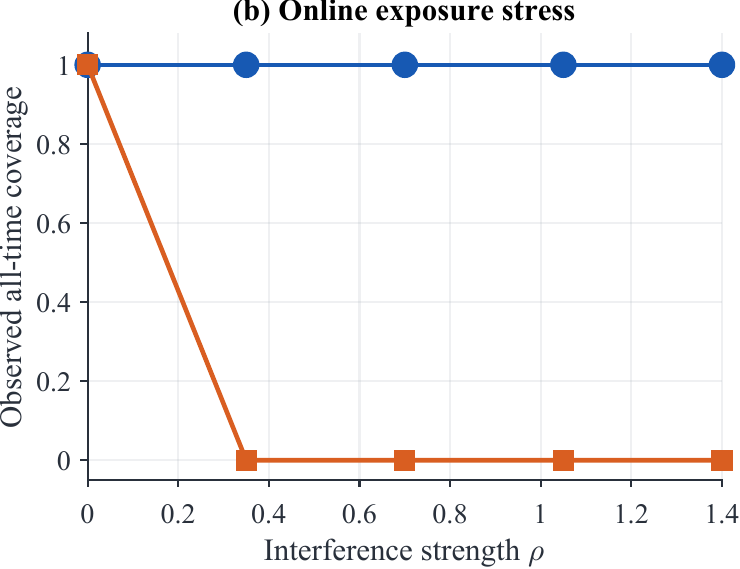}
\caption{Online exposure-versus-local stress test across the interference levels shown.
Panel (a) reports target regret; panel (b) reports observed all-time coverage.
The local estimator is heuristic and the coverage curve is diagnostic, not a certificate under the general theorem.}
\label{fig:online-exposure-stress}
\end{figure}
\begin{figure}[!htbp]
\centering
\begin{subfigure}[t]{0.32\linewidth}
\centering
\includegraphics[width=\linewidth]{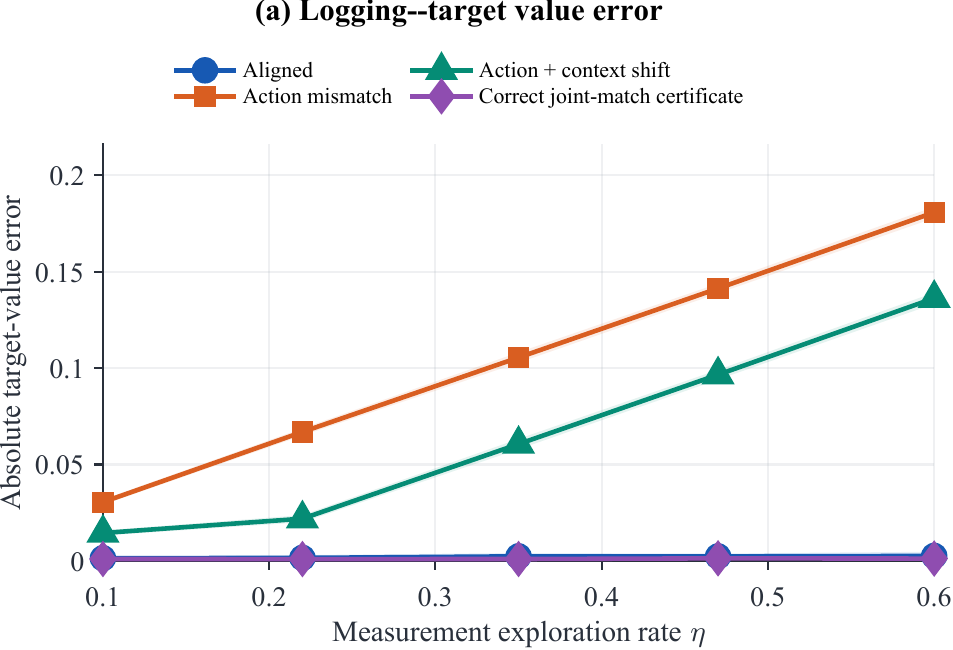}
\end{subfigure}\hfill
\begin{subfigure}[t]{0.32\linewidth}
\centering
\includegraphics[width=\linewidth]{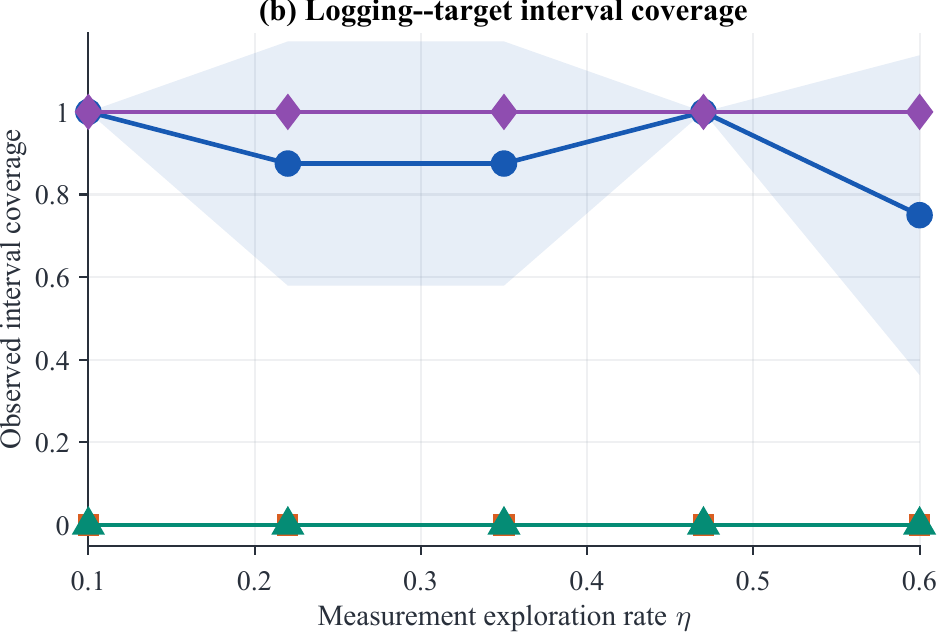}
\end{subfigure}\hfill
\begin{subfigure}[t]{0.32\linewidth}
\centering
\includegraphics[width=\linewidth]{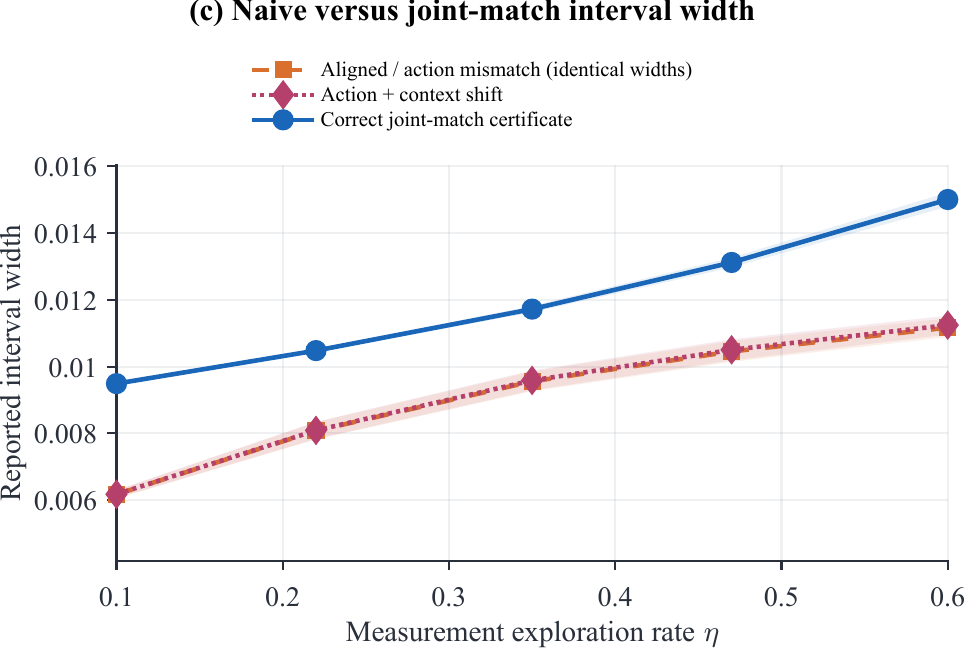}
\end{subfigure}
\caption{Logging--target alignment and mismatch sensitivity across the exploration rates shown.
The panels report absolute value error, naive interval coverage, and interval width.
The sampling-only intervals are diagnostic: high coverage or narrow width does not establish target identification under context shift.}
\label{fig:logging-target-factorial}
\end{figure}
\paragraph{Study provenance.}
The separate matched-log study reported DR/exposure ATE RMSE $0.0312$ and coverage $97.2\%$, while its online comparison reported DR-UCB/local regret $18.963$ and CAUSAL-UCB regret $21.723$.
That study used a distinct measurement design.
The primary $24\times200$ benchmark uses the controlled protocols and seed records described in Appendix~\ref{app:new-exp-protocol}; the two sets of results are interpreted within their respective experimental settings.


\subsubsection{Finite-state calibration under independent audit observations}
\label{app:finite-state-blackbox-operational}
Let $\mathcal Z,\mathcal C,\mathcal A$ be finite with cardinalities $M,J,K$.
Assume $R\in[0,1]$.
For each $z$, the logging law has context distribution $\nu_b^z(c)$, conditional action law $b_z(a\mid c)$, and response mean $m_{\rm log}^z(a,c)$.
The no-exploration target is
\[
 W_z^0=\sum_c\nu_0^z(c)\sum_a p_z(a\mid c)m_0^z(a,c).
\]
The operators $p_z, b_z,m_{\rm log}^z,m_0^z,\nu_b^z,\nu_0^z$ are not given to the learning algorithm.
Their estimation is conditional on independently identified calibration observations.
In particular, $m_0^z(a,c)$ must be identified from a valid target-regime response audit; observational exposure features alone do not ensure this.

\paragraph{Independent calibration.}
Freeze the fitted nuisances $\widehat p,\widehat b,\widehat m$ before the evaluated online trajectory.
At every indicated action/context cell, collect at least $n_{\min}$ independent conditional audit observations for $p,b,m_{\rm log},m_0$.
Independently collect context-law audit samples for $\nu_0,\nu_b,\nu_{\rm tr}$.
These audits are independent of nuisance fitting and online evaluation.
Denote audit means by tildes.
Take $D=M(4JK+3J)$, audit failure allocation $\alpha>0$, and
\[
 r=\sqrt{\frac{\log(2D/\alpha)}{2n_{\min}}}.
\]
Hoeffding and the union bound imply that every scalar audit error is at most $r$ jointly with probability at least $1-\alpha$.
Let $p^-=(\widetilde p-r)_+$, $p^+=\min(1,\widetilde p+r)$, $b^-=(\widetilde b-r)_+$, and $b^+=\min(1,\widetilde b+r)$, all cellwise.
The certificate refuses to issue if any $b^-=0$ or the propensity floor exceeds $\min b^-/2$.

For $\widehat w=\min\{\widehat p/\max(\widehat b,\pi_{\min}),w_{\max}\}$, define audit bounds (with maxima taken over $c,a$ or $c$ as indicated)
\begin{align}
 e_p(z)&=\max_c\sum_a(|\widehat p_z-\widetilde p_z|+r),\\
 e_m(z)&=\max_{c,a}(|\widehat m_z-\widetilde m_{\rm log}^z|+r),\\
 w^-&=\min(p^-/b^+,w_{\max}),\qquad
 w^+=\min(p^+/b^-,w_{\max}),\\
 e_w(z)&=\max_{c,a}\max\{|\widehat w-w^-|,|\widehat w-w^+|\},\\
 e_{\rm clip}(z)&=\max_c\sum_a[p^+-w_{\max}b^-]_+,\\
 e_\nu(z;v,0)&=\|\widetilde\nu_v^z-\widetilde\nu_0^z\|_1+2Jr,\\
 \widetilde E_{\rm exp}(z)&=\sum_c\widetilde\nu_0^z(c)
  \sum_a\widehat p_z(a\mid c)|\widetilde m_0^z(a,c)-
   \widetilde m_{\rm log}^z(a,c)|.
\end{align}
Under the scalar audit event, an explicit \emph{conservative} seven-part bias budget is
\begin{align}
 \epsilon_{\rm target}&=e_p,&
 \epsilon_{m\pi}&=e_m e_w,\\
 \epsilon_{\rm clip}&=e_{\rm clip},&
 \epsilon_{\rm exp}&=\widetilde E_{\rm exp}+Jr+e_p+2r,\\
 \epsilon_{\rm shift}&=\tfrac12 e_\nu(z;b,0),&
 \epsilon_{\rm cfg}&=e_m e_\nu(z;{\rm tr},0),\\
 \epsilon_{\rm drift}&=2(1+2w_{\max})d_z.&
\end{align}
Here $d_z$ is an \emph{externally verified} total-variation envelope for the residual deviation of a retained network-window score law from its stationary counterpart.
All components are nonnegative and may be capped by crude support bounds if desired.
The configuration term is a conservative, optional allowance: the cellwise regression audit already controls training-to-target transport in this finite-state construction.
This does not establish such transport for an arbitrary unobserved latent confounder or unmatched outcome semantics.

\paragraph{Dependent network-window evaluations.}
Define the frozen score
\[
 \psi_z(C,A,R)=\sum_a\widehat p_z(a\mid C)\widehat m_z(a,C) +\widehat w_z(A,C)\{R-\widehat m_z(A,C)\}.
\]
With $\widehat m_z\in[0,1]$, this score lies in $[-w_{\max},1+w_{\max}]$, an interval of length $S=1+2w_{\max}$.
Treat each retained \emph{network window} as one observation, regardless of the number of correlated nodes in it.
Assume the per-configuration infinite stream of retained windows is stationary, or has the above independently certified drift envelope, and obeys a \emph{verified} conditional block-coupling bound: for separation $g$, the total coupling failure over all candidate streams and at most $n_{\max}$ retained windows per stream is at most
\[
 \delta_{\rm mix}\ge M n_{\max}\overline\beta(g).
\]
For example, an independently established Markov TV contraction $\overline\beta(g)\le\rho^g$ suffices.
An empirical autocorrelation estimate by itself does not prove this assumption.
Nuisances are frozen before scoring; arbitrary adaptive cross-fitting requires an additional predictability/buffering proof.

\begin{proposition}[Operational finite-audit certificate]
\label{prop:operational-finite-audit}
Under all the independent-audit, identification, common conditional- response meaning, support, mixing, and drift assumptions above, allocate $\alpha$ to calibration, $\delta_s$ to the centered-score event, and $\delta_{\rm mix}$ to coupling, with $\alpha+\delta_s+\delta_{\rm mix}\le\delta$.
Then, simultaneously for every $z$ and every scored prefix $1\le n\le n_{\max}$, with probability at least $1-\delta$,
\[
 |\widehat W_{z,n}^{\rm DR}-W_z^0|\le
 \underbrace{S\sqrt{\frac{\log(2M(n_{\max}+1)/\delta_s)}{2n}}}_{\text{centered network-window score}}
 +\sum_{v\in\{\mathrm{target},m\pi,\mathrm{clip},\mathrm{exp},
                  \mathrm{shift},\mathrm{cfg},\mathrm{drift}\}}\epsilon_v(z).
\]
Unvisited configurations are assigned the trivial $[0,1]$ interval.
The radius depends only on frozen fitted nuisances, independent calibration data, retained scoring observations, and certified coupling/drift envelopes; no true target values enter online selection.
\end{proposition}
\begin{proof}
The calibration event follows from Hoeffding over $D$ bounded scalar quantities.
For each context, let $\mu_{\rm log}(c)=\sum_a p(a\mid c)m_{\rm log}(a,c)$.
The conditional score expectation obeys the exact identity
\[
 \mathbb E_b[\psi\mid c]-\mu_{\rm log}(c)
 =\sum_a(\widehat p-p)\widehat m +\sum_a(p-b\widehat w)(\widehat m-m_{\rm log}).
\]
The first term is bounded by $e_p$; the second by $e_m\{e_w+e_{\rm clip}\}$, and the looser displayed ledger $e_m e_w+e_{\rm clip}$ is valid because $e_m\le1$.
Transporting the context law from $\nu_b$ to $\nu_0$ costs at most the total variation bound $\epsilon_{\rm shift}$ for a function in $[0,1]$.
Transporting the conditional response from $m_{\rm log}$ to $m_0$ is bounded by $\epsilon_{\rm exp}$, using $r$ for each target/logging conditional mean and audited target weights.
$\epsilon_{\rm cfg}$ and $\epsilon_{\rm drift}$ are conservative further upper bounds on the indicated residual terms.
Conditional on successful block coupling, the centered frozen score is a bounded independent sequence; a Hoeffding union bound across all $M$ candidates and all $n_{\max}$ prefixes gives the displayed sampling term.
Adding the calibration and coupling failure budgets proves the simultaneous bound.
Adaptivity of the arm order is allowed because the event covers every prefix of every presampled arm stream.
\end{proof}
\begin{corollary}[Use of the same radius in online decisions]
If the algorithm chooses $z_t\in\arg\max_z \min\{1,\widehat W_{z,t-1}+b_{z,t-1}\}$ from the \emph{predecision} certified radii above, then outside $H$ pilot rounds the same joint event gives $W^0(z^\star)-W^0(z_t)\le2b_{z_t,t-1}$, hence $\operatorname{Reg}_T\le H+2\sum_{t>H}b_{z_t,t-1}$.
\end{corollary}
\paragraph{Information requirements and calibration cost.}
The finite-state certificate combines independent observations of target context and action laws, target-regime response calibration, and a specified conditional mixing envelope.
These quantities define the auditable black-box class in which the stated confidence event holds.
Their measurement cost is included when interpreting the identification--deployment trade-off.
The construction complements the local-interference certificate with a finite-state route to calibrating the general error decomposition.

%% file: appendix/31_certified_optimism_initial_information.tex
\subsection{Certified Optimism under Heterogeneous Initial Information}
\label{app:certified-optimism-initial-info}

\paragraph{Question and controlled design.}
We investigate when the confidence upper bounds constructed from the seven-term causal value certificate provide a useful configuration-level information acquisition mechanism, beyond the point estimates alone.
The experiment uses the independently audited, finite-state black-box response model underlying our operational certificate test.
There are three global configurations, two context states, and two possible local actions.
In addition to logged rewards, external action/intervention and context audits are available; a known retained-block transition envelope controls dependence.
These assumptions are sufficient for the audit construction, but are not implied by arbitrary black-box exploratory logs.

We compare \emph{Certified-OFU}, selecting $\arg\max_z\{\widehat W(z;0)+b_z\}$ (with target-value truncation), \emph{Certified-Greedy}, selecting $\arg\max_z\widehat W(z;0)$, and a pre-specified decaying forced-exploration Greedy policy.
All policies share identical nuisance fits, independent audits, initial scored windows, and configuration-specific potential score streams; only the choice rule differs.
For a visited configuration, the certificate radius takes the form
\begin{equation}
 b_z(n)=3.2\sqrt{\frac{\log(2KN_{\max}/\delta_s)}{2n}}
 +\sum_{j=1}^{7}\beta_{z,j},\quad
 \delta_s=0.03,\ \delta_{\mathrm{audit}}=0.02.
 \label{eq:certstarve-radius}
\end{equation}
The seven nonnegative, independently audited buckets cover target action, nuisance product, clipping, exposure/response transport, context shift, cross-configuration transfer, and retained-block dependence/drift.
The final bucket includes the explicit predictable-mean allowance $3.2(0.45)^6$ for the known retained-block kernel $P_z=\lambda I+(1-\lambda)\mathbf{1}\nu_{z,\mathrm{log}}^\top$, $\lambda=(0.45)^6$.
The finite audit and martingale-score failure budgets sum to $0.05$ for each independent run.

\paragraph{Initial-information regimes.}
We fix five regimes before evaluating outcomes.
The primary \emph{starved-optimal} regime has target values $(0.506,0.511,0.527)$ and pilot scored-window allocations $(1500,1500,120)$; thus the truly optimal third configuration is initially least observed.
Balanced-easy and balanced-ambiguous regimes use equal $(900,900,900)$ pilots; two further regimes assign the 120-window pilot to a suboptimal configuration instead.
In each regime we evaluate 96 paired seeds, 80 online rounds, and 400 new scored retained blocks per round, with the same exploration mixture $\eta=0.46$, independent $200{,}000$-sample audits per calibrated cell, and $10{,}000$ fitted observations per cell.
Exact simulator target values are used only to assess regret and coverage, never by the learner.

\begin{table}[H]
\centering
\caption{\textbf{Certified OFU versus Greedy under five fixed initial-information regimes.} Mean cumulative target regret across 96 paired seeds; $n_{\rm mis}$ is the number of seeds whose initial point estimates rank the true best configuration incorrectly.
Lower regret is better.
All regimes are reported.}
\label{tab:certstarve-five-regimes}
\small
\setlength{\tabcolsep}{5pt}
\begin{tabular}{lrrrr}
\toprule
Regime & $n_{\rm mis}$ & Certified-OFU & Certified-Greedy & Decaying-Forced \\
\midrule
Easy, balanced & 0 & 0.327 & \textbf{0.000} & 1.728 \\
Ambiguous, balanced & 17 & 0.330 & \textbf{0.051} & 0.299 \\
Starved optimal & 33 & \textbf{0.241} & 0.467 & 0.268 \\
Starved middle & 24 & 0.297 & \textbf{0.019} & 0.238 \\
Starved low & 30 & 0.314 & \textbf{0.024} & 0.241 \\
\bottomrule
\end{tabular}
\end{table}

\paragraph{Correction of an initial misranking.}
In the starved-optimal regime, 33 of 96 independent seeds exhibit an \emph{initial} point-estimate misranking.
On these 33 paths, mean cumulative regret is $0.232$ for Certified-OFU and $1.077$ for Certified-Greedy; the paired mean difference (Greedy minus OFU) is $0.845$, with a seed-level descriptive 95\% confidence interval $[0.621,1.068]$.
The initial misranking persists in the point-estimate ordering for 22/33 Greedy paths at the final round.
The full-sample difference across all 96 seeds remains positive, $0.226$ (95\% CI $[0.091,0.362]$).
These outcomes reveal a configuration-information bottleneck: a policy that never visits an underestimated candidate cannot shrink that candidate's estimation uncertainty through selected-arm feedback.

\begin{figure}[!htbp]
\centering
\includegraphics[width=.48\linewidth]{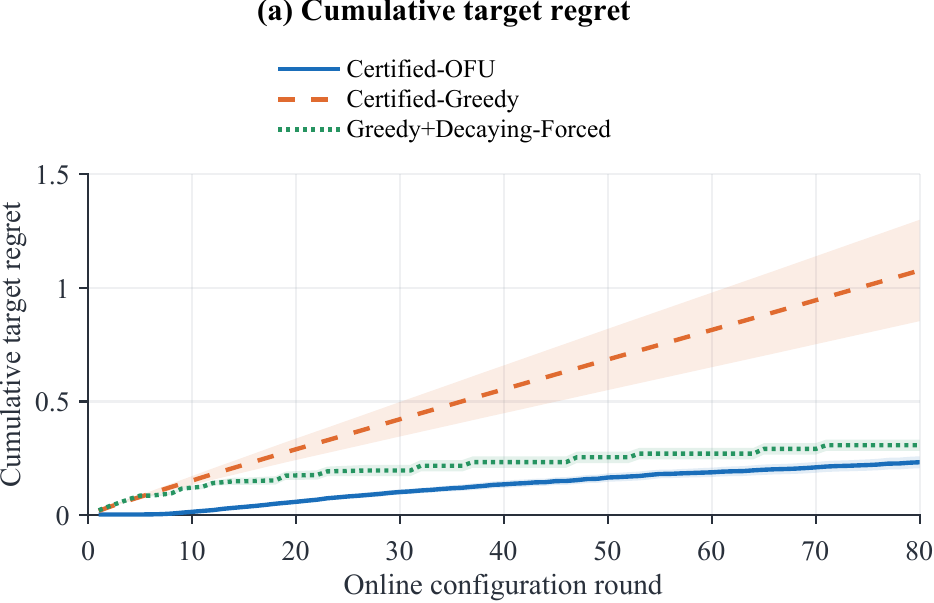}\hfill
\includegraphics[width=.48\linewidth]{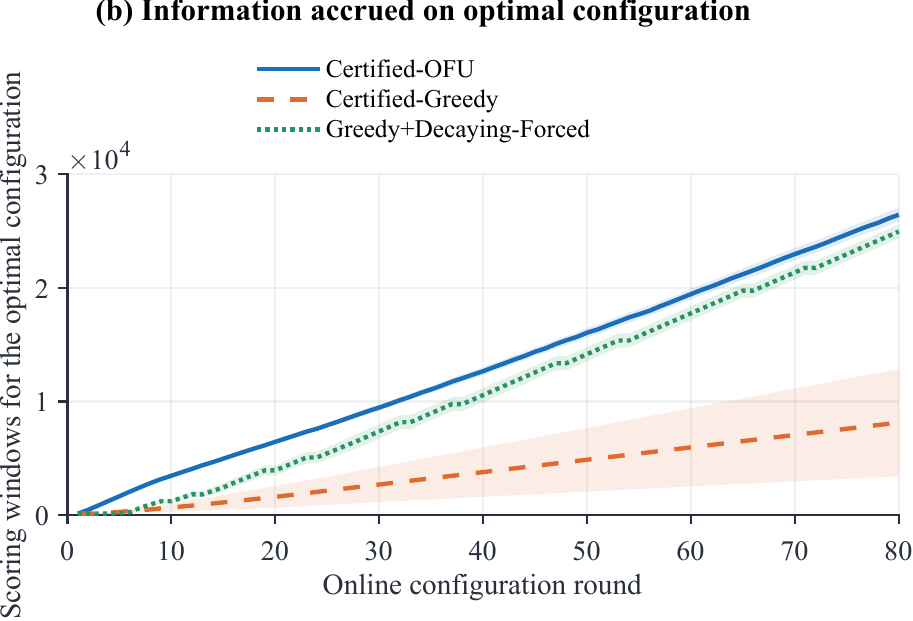}\hfill
\caption{\textbf{Correction under initially insufficient optimal-configuration information.} Mean trajectories and pointwise 95\% seed-level intervals on the 33 initial-misranking paths.
Left: cumulative target regret.
Right: number of certified scoring windows accrued on the truly optimal configuration.
All policies start from the same pilot counts and receive 400 new scored blocks per online round.}
\label{fig:certstarve-regret-info}
\end{figure}

\paragraph{The role of certificate tightness.}
To test whether the advantage arises exclusively from the optimal configuration having a larger \emph{fixed} seven-term budget, we repeat the starved-optimal regime after enlarging each of the seven arm-specific bias bounds to the maximum across all arms.
This is still a valid, more conservative certificate and makes the fixed bias terms identical.
On the 33 initial-misranking paths, equal-budget OFU obtains mean cumulative regret $0.724$ versus Greedy's $1.077$; the paired difference is $0.353$ (95\% CI $[0.130,0.576]$).
Across \emph{all} 96 paths, however, equal-budget OFU incurs $0.724$ regret versus Greedy's $0.467$: conservative uncertainty induces unnecessary exploration when the point estimates are already adequate.
The original arm-specific seven-term OFU achieves $0.241$ on these 96 paths, emphasizing the importance of certificate tightness as well as coverage validity.

\begin{figure}[!htbp]
\centering
\includegraphics[width=.48\linewidth]{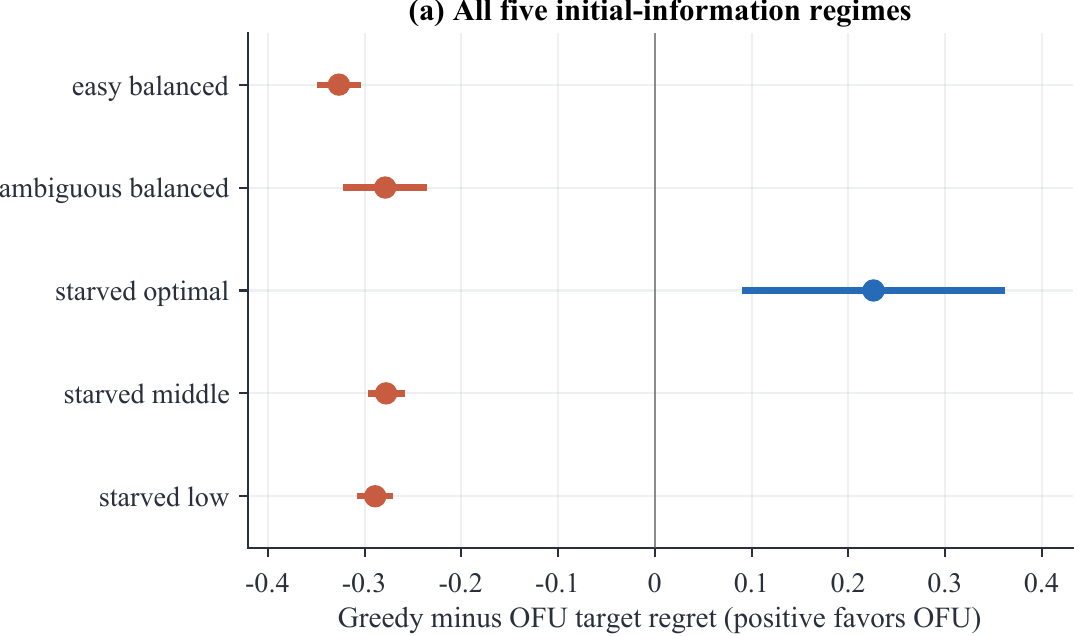}\hfill
\includegraphics[width=.48\linewidth]{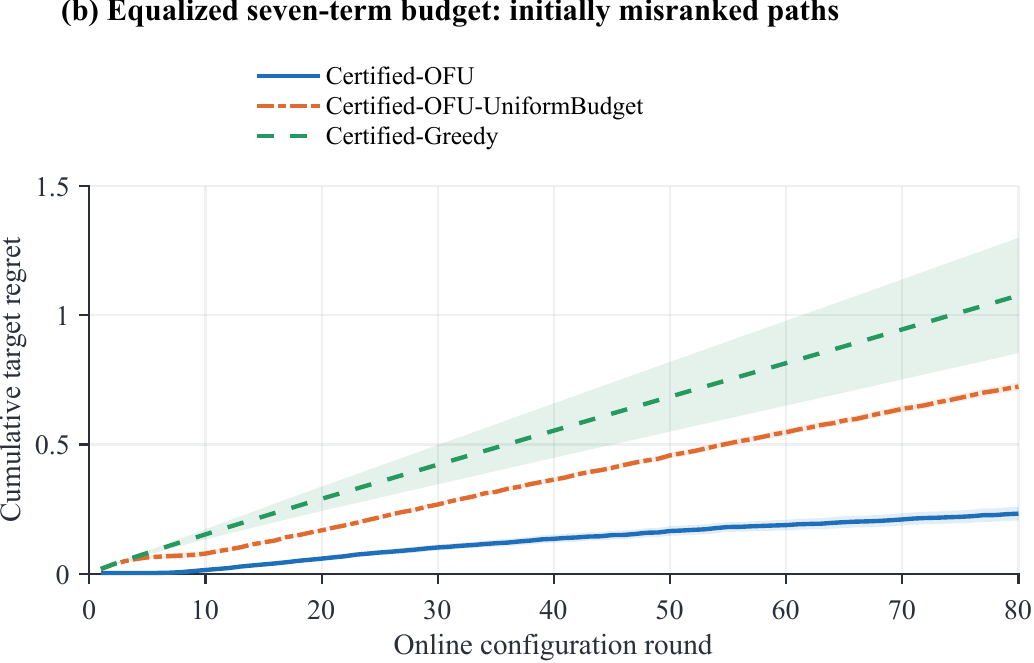}\hfill
\caption{\textbf{Decision value is conditional on the information structure.} Left: paired Greedy-minus-OFU target regret, with 95\% seed-level intervals, in all five fixed regimes (positive favors OFU).
Right: a valid conservative uniform seven-term bias envelope preserves OFU's advantage on initially misranked paths but incurs additional exploration cost.}
\label{fig:certstarve-scopes}
\end{figure}

\paragraph{Audit cost and interpretation.}
All three policies spend 32,000 additional scored blocks per seed; independent intervention audits are a separate, substantial prerequisite.
In the starved-optimal scenario, the model-based \emph{expected} measurement-induced utility displacement $\sum_t[W_{\infty}(z_t;0)-W_{\infty}(z_t;\eta)]$ is approximately $3.319$ for Certified-OFU versus $3.137$ for Greedy (oracle-computed diagnostic, never used in selection).
Thus improved target-regret control can coincide with greater measurement-induced utility displacement.
The case study identifies \emph{when} the OFU mechanism can correct point-estimation lock-in under valid audit assumptions; it does not imply OFU is uniformly optimal, nor does it extend the finite-state certificate automatically to unaudited networks.
Tight certificates and meaningful information-growth conditions are both necessary ingredients when interpreting the regret result.

%% file: appendix/appendix_xxv3_ofu_18cell.tex
\subsection{Gap--Noise--Budget Stress Test of Certified Optimism}
\label{app:ofu-18cell-stress}

To characterize when optimism helps and when it does not, we ran a predeclared $3\times3\times2$ stress grid over target-value gap, response-noise level, and online scoring budget.
The top-versus-middle target gap is $\Delta\in\{0.005,0.016,0.030\}$, the bounded-noise amplitude is $\sigma\in\{0.2,0.6,1.0\}$, and the budget is 200 or 400 scored windows per decision.
For response mean $m(c,a)$, the generated response is $R=m(c,a)+\sigma\min\{m(c,a),1-m(c,a)\}(2U-1)$ with $U\sim\mathrm{Unif}[0,1]$, preserving the conditional mean and the $[0,1]$ bound.
We use 24 paired seeds per cell, 80 online decisions, 10,000 fitting samples per cell, and 200,000 independent audit samples per cell.
The paired comparison is $D_s=\operatorname{Regret}_{s}^{\mathrm{OFU}}-\operatorname{Regret}_{s}^{\mathrm{Greedy}}$; negative values favor OFU.

Across the 18 cells, OFU has lower mean regret in 12 cells and Greedy has lower mean regret in 6.
The unweighted mean of the 18 cell-level paired differences is $-0.0426$, but this average is descriptive because the cells represent different operating regimes.
Only four cell-wise 95\% paired-$t$ intervals lie entirely below zero, while two lie entirely above zero; the remaining intervals include zero.
The results therefore support a conditional mechanism claim, not universal dominance by either policy.


\begin{figure*}[!t]
\centering

\begin{subfigure}[t]{0.49\textwidth}
    \centering
    \includegraphics[width=0.98\linewidth]
    {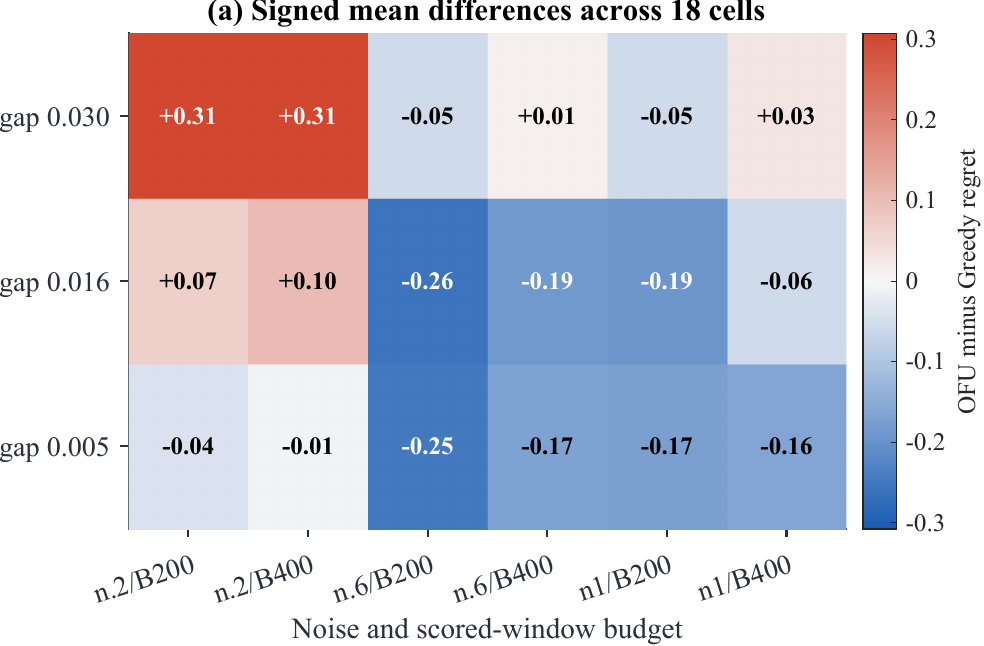}
\end{subfigure}\hfill%
\begin{subfigure}[t]{0.49\textwidth}
    \centering
    \includegraphics[width=\linewidth]
    {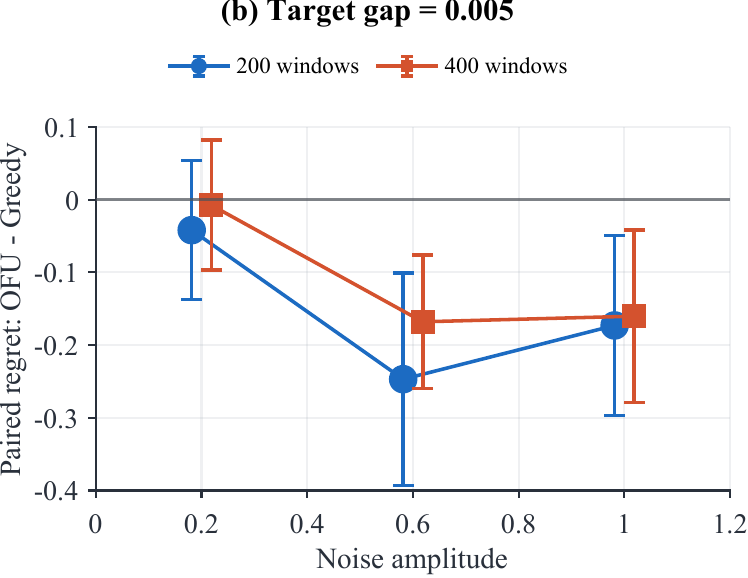}
\end{subfigure}

\vspace{3pt}

\begin{subfigure}[t]{0.49\textwidth}
    \centering
    \includegraphics[width=\linewidth]
    {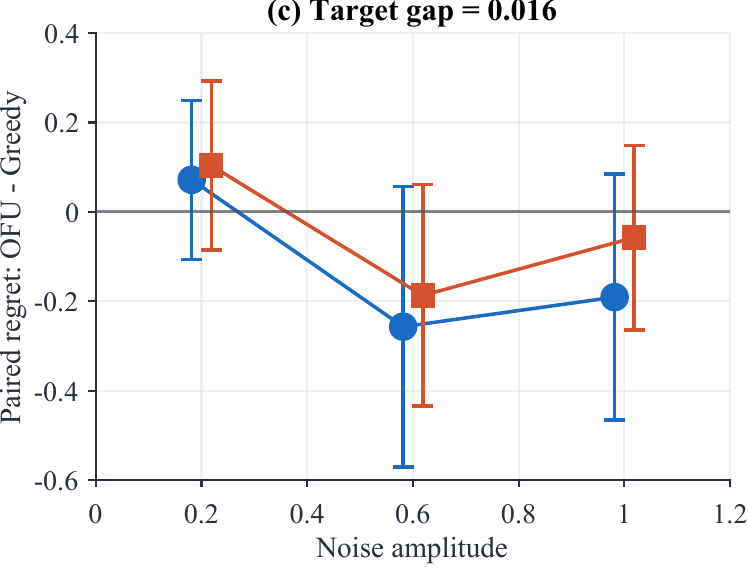}
\end{subfigure}\hfill%
\begin{subfigure}[t]{0.49\textwidth}
    \centering
    \includegraphics[width=\linewidth]
    {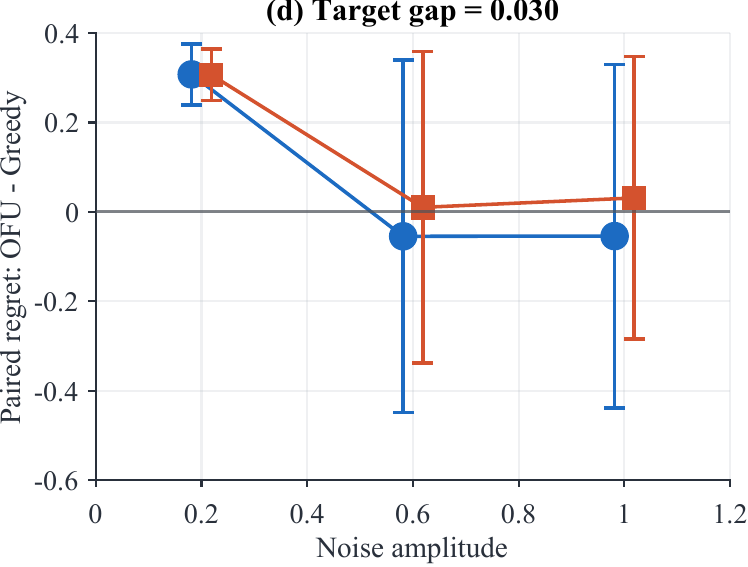}
\end{subfigure}

\captionsetup{font=footnotesize,skip=3pt}
\caption{\textbf{Gap--noise--budget sensitivity of certified optimism.}
(a) Mean paired regret differences across the
$3\times3\times2$ stress grid, n.2/B200 denotes noise amplitude
$\sigma=0.2$ and a budget of 200 scored windows
per decision; the remaining labels follow
the same convention;
(b)--(d) paired regret differences across noise amplitudes
at target gaps $0.005$, $0.016$, and $0.030$,
respectively, under 200 and 400 scored windows.
Differences are OFU minus Greedy; negative values favor OFU.
Each cell uses 24 paired seeds and 80 online rounds.
Error bars denote 95\% paired Student-$t$ intervals;
the heatmap reports cell means.}
\label{fig:ofu-18cell-overview}
\end{figure*}

\begin{figure*}[!b]
\centering

\begin{minipage}[t]{0.53\textwidth}
\vspace{0pt}
\centering
\captionsetup{font=footnotesize,skip=3pt}

\captionof{table}{Cell-wise means and paired regret differences
in the gap--noise--budget grid.
$D=\mathrm{Regret}_{\rm OFU}-\mathrm{Regret}_{\rm Greedy}$;
negative $D$ favors OFU.}
\label{tab:ofu-greedy-18cell}

\scriptsize
\setlength{\tabcolsep}{2pt}
\renewcommand{\arraystretch}{1.08}

\resizebox{\linewidth}{!}{%
\begin{tabular}{rrrrrrl}
\hline
Gap & $\sigma$ & Windows & OFU regret &
Greedy regret & $\bar D$ & Paired 95\% CI\\
\hline
0.005 & 0.20 & 200 & 0.104 & 0.146 & -0.042 & [-0.138, 0.054] \\
0.005 & 0.20 & 400 & 0.099 & 0.106 & -0.008 & [-0.097, 0.082] \\
0.005 & 0.60 & 200 & 0.104 & 0.351 & -0.247 & [-0.393, -0.101] \\
0.005 & 0.60 & 400 & 0.102 & 0.270 & -0.168 & [-0.260, -0.076] \\
0.005 & 1.00 & 200 & 0.101 & 0.274 & -0.173 & [-0.296, -0.049] \\
0.005 & 1.00 & 400 & 0.101 & 0.261 & -0.160 & [-0.279, -0.042] \\
0.016 & 0.20 & 200 & 0.238 & 0.166 & 0.072 & [-0.106, 0.250] \\
0.016 & 0.20 & 400 & 0.228 & 0.124 & 0.104 & [-0.085, 0.293] \\
0.016 & 0.60 & 200 & 0.236 & 0.493 & -0.257 & [-0.571, 0.057] \\
0.016 & 0.60 & 400 & 0.231 & 0.418 & -0.187 & [-0.435, 0.061] \\
0.016 & 1.00 & 200 & 0.227 & 0.418 & -0.191 & [-0.466, 0.085] \\
0.016 & 1.00 & 400 & 0.229 & 0.287 & -0.057 & [-0.264, 0.149] \\
0.030 & 0.20 & 200 & 0.309 & 0.001 & 0.308 & [0.240, 0.376] \\
0.030 & 0.20 & 400 & 0.307 & 0.000 & 0.307 & [0.249, 0.364] \\
0.030 & 0.60 & 200 & 0.300 & 0.355 & -0.055 & [-0.449, 0.340] \\
0.030 & 0.60 & 400 & 0.314 & 0.304 & 0.010 & [-0.338, 0.359] \\
0.030 & 1.00 & 200 & 0.296 & 0.350 & -0.054 & [-0.438, 0.330] \\
0.030 & 1.00 & 400 & 0.308 & 0.276 & 0.031 & [-0.285, 0.347] \\
\hline
\end{tabular}%
}
\end{minipage}
\hfill
\begin{minipage}[t]{0.45\textwidth}
\vspace{0pt}
\centering
\captionsetup{font=footnotesize,skip=3pt}

\includegraphics[width=\linewidth]%
{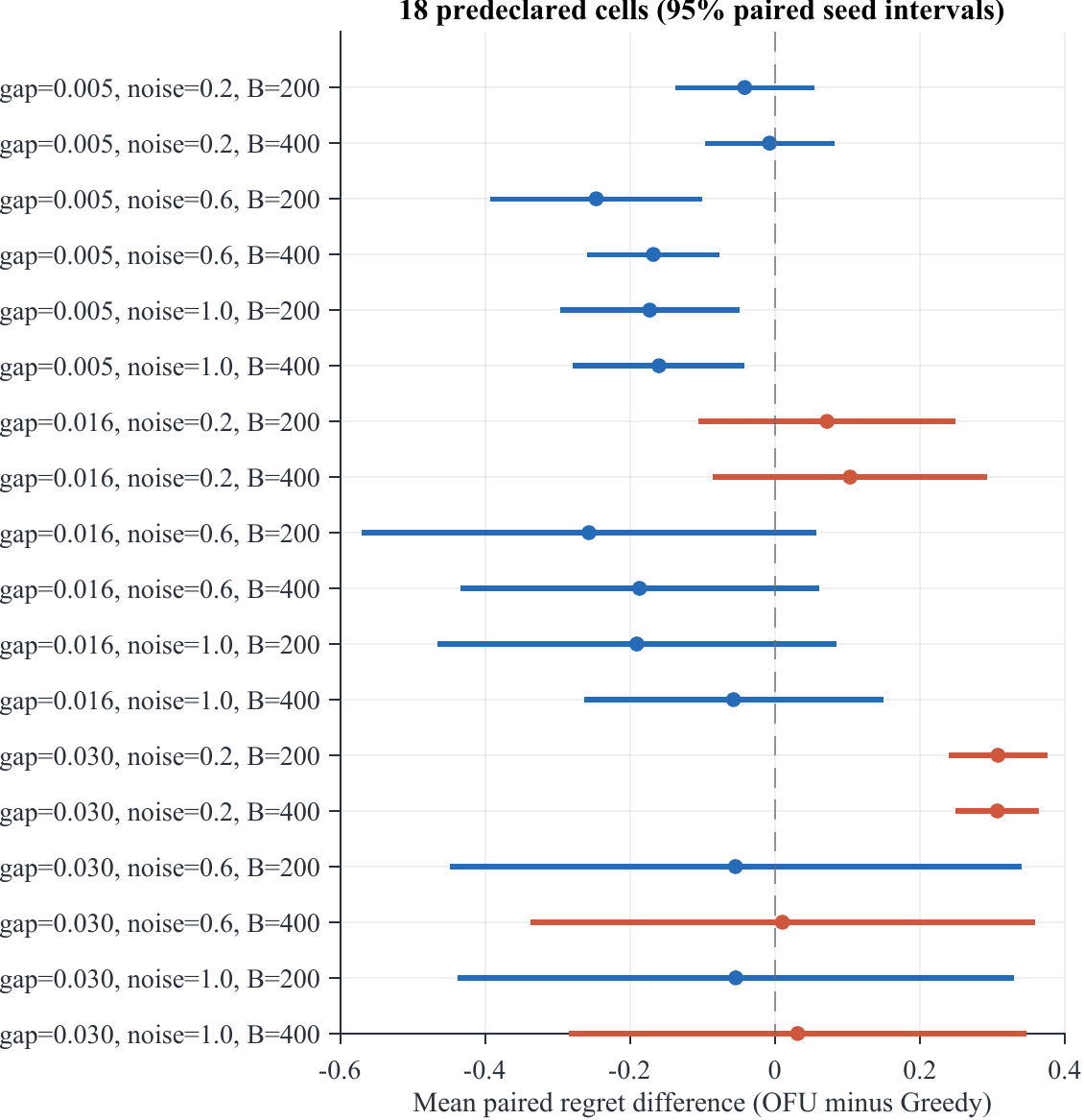}

\captionof{figure}{Paired regret differences across
the 18 gap--noise--budget cells.
The vertical zero line denotes equal mean regret;
negative differences favor certified OFU.
Error bars are 95\% paired Student-$t$ intervals
across 24 seeds per cell. $B$ denotes the observation budget.}
\label{fig:ofu-greedy-18cell}

\end{minipage}

\end{figure*}

Figure~\ref{fig:ofu-18cell-overview} summarizes
the sensitivity patterns: OFU has lower mean regret
in every smallest-gap cell, whereas Greedy performs
better in the largest-gap, lowest-noise cells.
The cell-wise values and paired intervals are given
in Table~\ref{tab:ofu-greedy-18cell} and
Figure~\ref{fig:ofu-greedy-18cell}.

The seed count per cell is 24; it is a supplementary sensitivity analysis, while the 96-seed five-scenario experiment remains the primary mechanism study.

%% file: appendix/appendix_xxv2_coverage_300x160.tex

\subsection{Empirical Full-Path Validation of the Seven-Component Certificate}
\label{app:seven-term-fullpath-300}

\begin{wrapfigure}{r}{0.43\linewidth}
\vspace{-8pt}
\centering
\includegraphics[width=\linewidth]{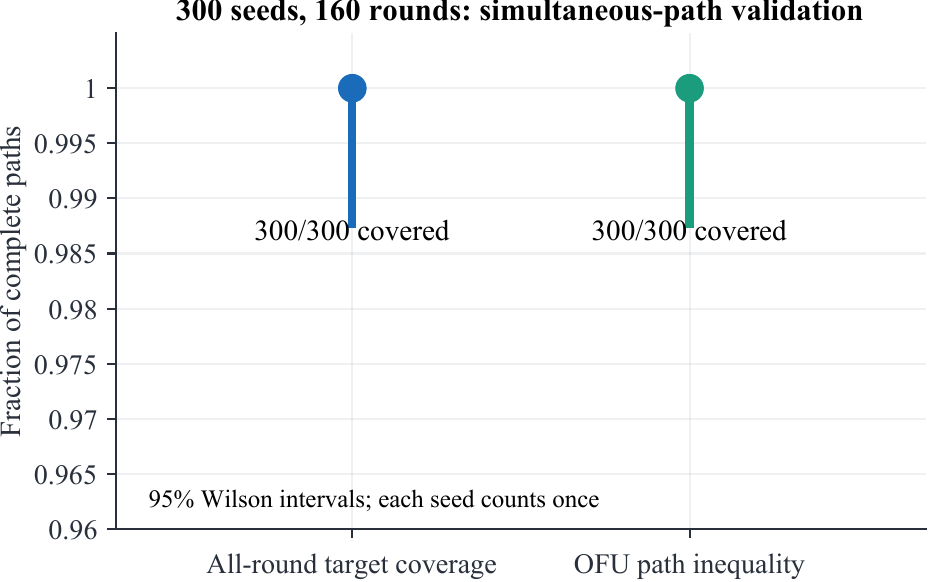}
\captionsetup{font=footnotesize,skip=3pt}
\caption{Full-path certificate coverage and the OFU path inequality over 300 independently seeded runs, each with 160 online rounds. Points show the empirical fractions and intervals show 95\% Wilson uncertainty.}
\label{fig:seven-term-coverage-300}
\vspace{-8pt}
\end{wrapfigure}

We evaluate the operational seven-component certificate under the finite-state, independently audited protocol specified in the finite-state audit construction in Appendix~\ref{app:finite-state-blackbox-operational}.
The experiment uses 300 independently seeded runs, 160 online rounds per run, and 2,000 scored windows per decision.
Each run uses 10,000 frozen fitting observations and 200,000 independent observations per calibration cell.
The seven-term attestation is constructed before online decisions; the simulator's true target values are used only by the external evaluator and never by the policy.

A run is counted as covered only if every configuration's target value lies inside its reported interval at every predecision time and at the terminal check.
All 300 runs satisfy this full-path criterion (300/300; empirical rate 100\%; Wilson 95\% interval $[0.9874,1.0000]$).
The OFU path inequality is also satisfied in all 300 runs.
Mean cumulative regret is 1.0298 and the mean total seven-term bias budget is 0.0975 under this particular generator.

These results are an implementation-level check under the stated finite-state audit, target-intervention identification, and verified dependence-envelope assumptions.
They are not a proof of 100\% population coverage, and they do not establish that these assumptions can be inferred from exploratory logs alone.
The additional audit observations are a substantive information cost and are not included in online regret.

\begin{figure}[H]
\centering
\includegraphics[width=.32\linewidth]{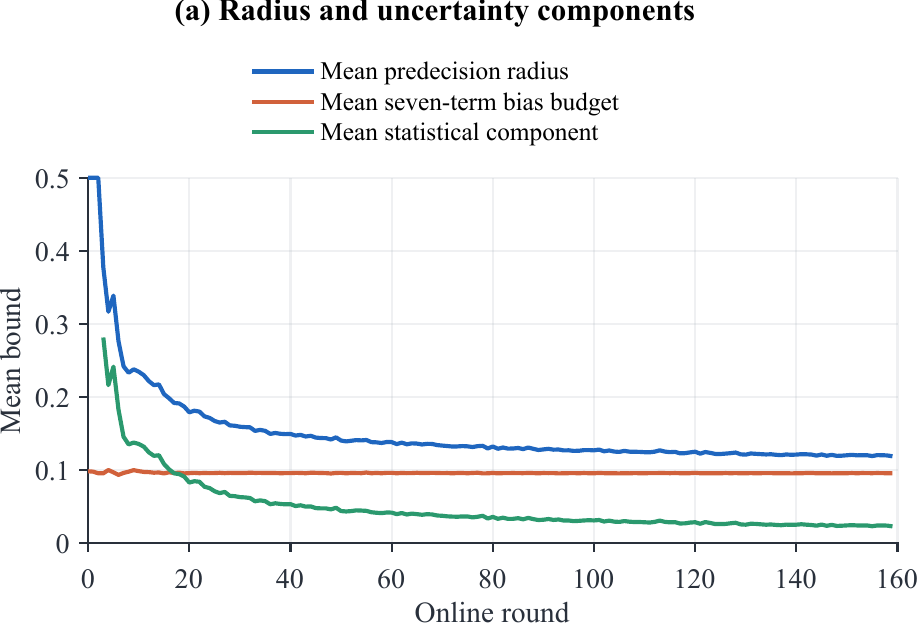}\hfill
\includegraphics[width=.32\linewidth]{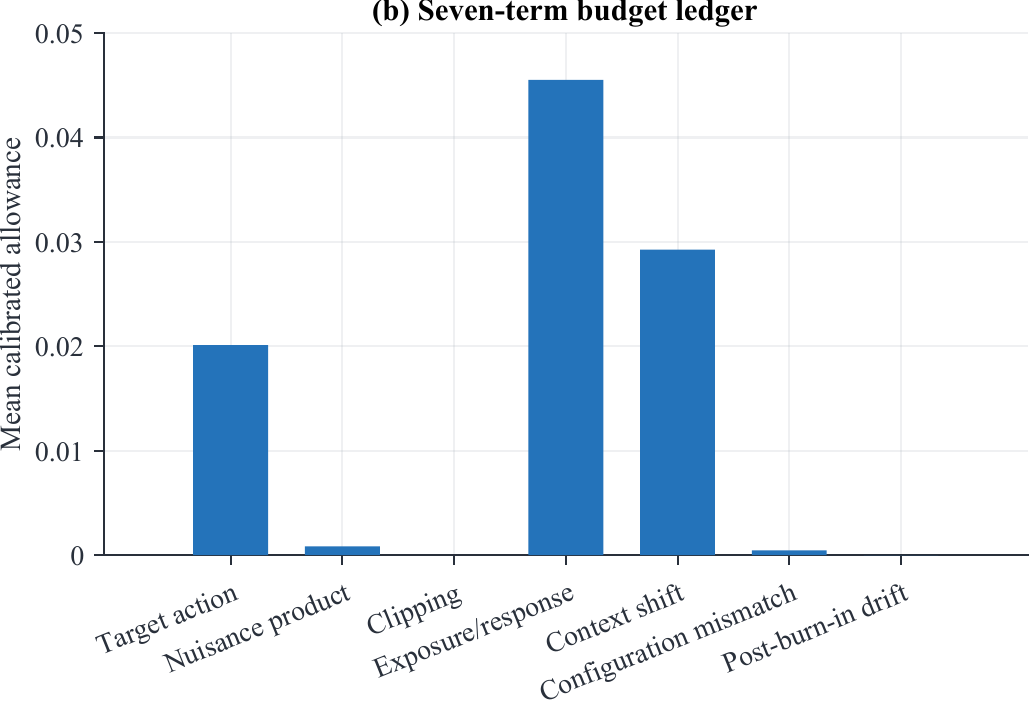}\hfill
\includegraphics[width=.32\linewidth]{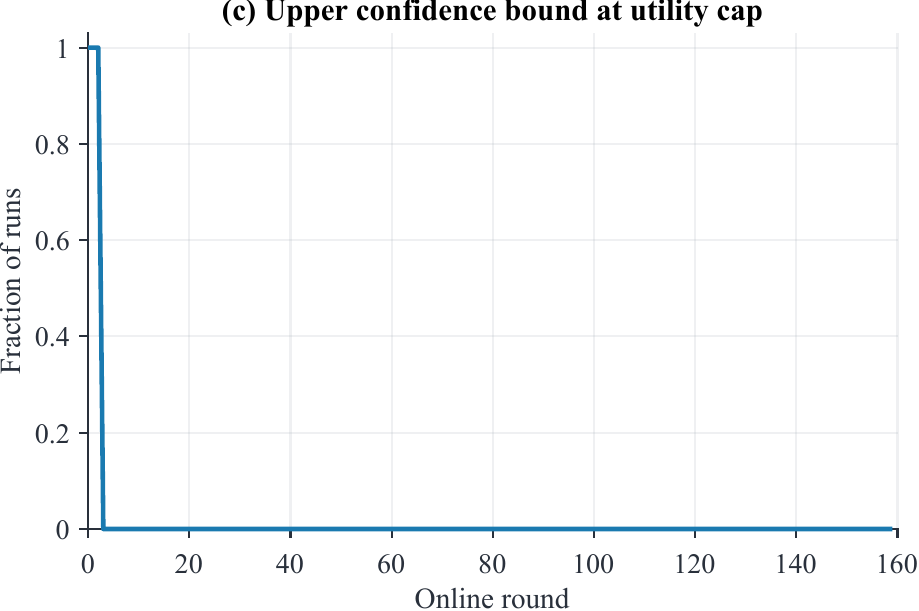}
\caption{Certificate informativeness under the 300-seed, 160-round audit. Panel (a) reports the seed-averaged predecision radius and its statistical and calibrated-bias components; panel (b) reports the seven-term budget ledger; panel (c) shows the fraction of runs whose selected upper confidence bound reaches the utility cap. The separate 24-seed pilot diagnostics are reported in Figure~\ref{fig:certificate-pilot-diagnostics}.}
\label{fig:certificate-informativeness-300}
\end{figure}

\begin{figure}[H]
\centering
\includegraphics[width=.32\linewidth]{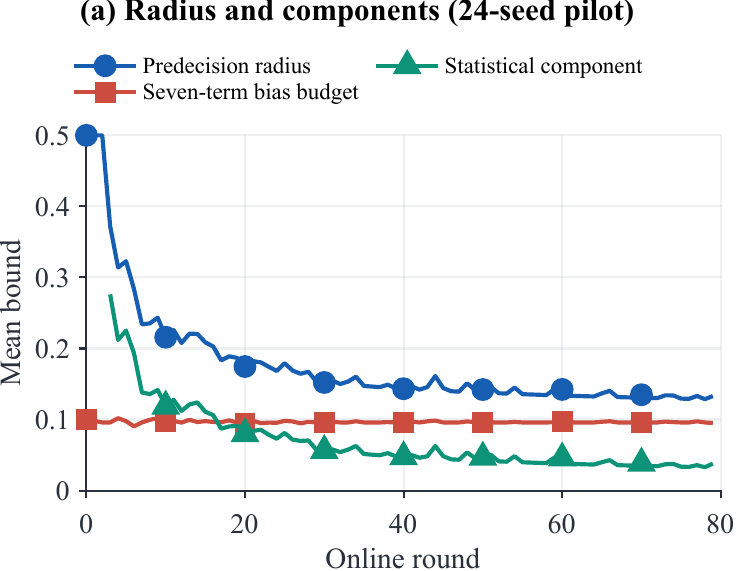}\hfill
\includegraphics[width=.32\linewidth]{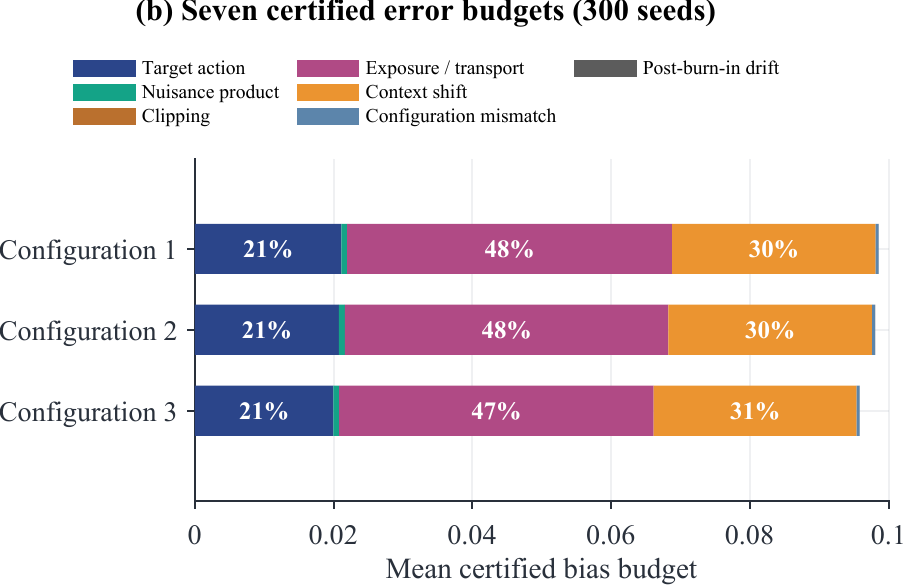}\hfill
\includegraphics[width=.32\linewidth]{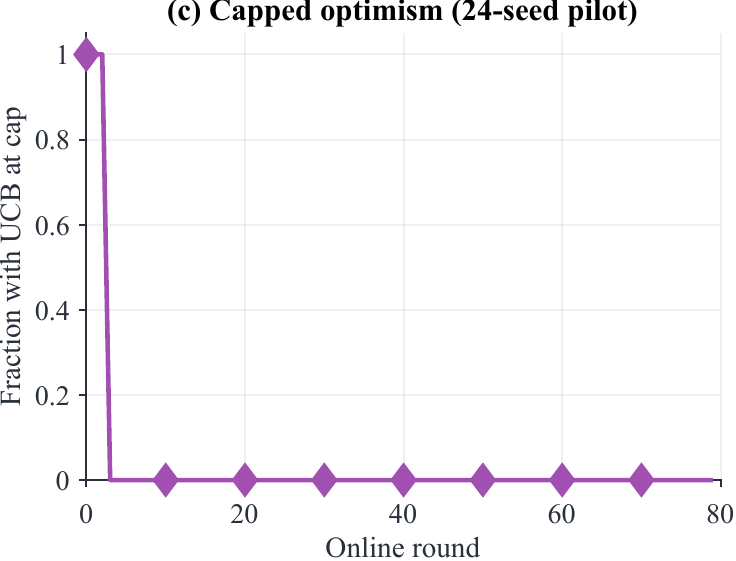}

\caption{Supplementary certificate diagnostics from the panel-source archive.
Panel (a) reports radius components in the 24-seed, 80-round pilot; panel (b) shows the configuration-specific bias-budget composition in the separate 300-seed audit; panel (c) reports capped-optimism frequency in the 24-seed pilot.
The protocols are intentionally distinguished and their outcomes are not pooled.
The 300-seed audit's aggregate informativeness summary appears in Figure~\ref{fig:certificate-informativeness-300}.}
\label{fig:certificate-pilot-diagnostics}
\end{figure}

\FloatBarrier

%% file: appendix/appendix_xxv5_calibration_misspecification.tex
\subsection{Coverage under Hidden Calibration Misspecification}
\label{app:seven-term-misspec}

We directly stress the finite-state seven-term certificate by violating one assumption at a time while leaving the corresponding provenance flag falsely asserted as valid.
Each condition contains 200 seeds, 120 online rounds per seed, 1,200 scored windows per decision, 1,000 fitting samples per cell, and 50,000 samples per cell claimed by the audit interface.
The six hidden violations are: (i) the target action law differs from the audited target law; (ii) the target context distribution shifts after audit; (iii) the target interventional response shifts after audit; (iv) a context-dependent target--logging response transport effect is omitted; (v) the actual retained-block persistence is larger than the declared dependence envelope; and (vi) training estimates are reused as if they were independent audit estimates while the larger audit sample count is still claimed.
The perturbation magnitudes are fixed as follows: for the third configuration, the target action probability for action 0 is reduced by 0.26 (with the complementary action probability adjusted accordingly); the target context law is changed to $(0.88,0.12)$; the target response is shifted upward by 0.17 and clipped to $[0,1]$; the transport stress shifts context-0 responses downward by 0.18 and context-1 responses upward by 0.18; the true persistence parameter is increased from the declared $0.45$ to $0.92$ with the same retained separation; and the audit-independence stress reuses 1,000 training estimates while falsely declaring 50,000 independent audit samples.
The simulator's true target values are used only for post-hoc coverage and regret evaluation.

\begin{table}[!htbp]
\centering
\small
\caption{Full-path coverage conditional on deliberately false certificates being issued. Each condition has 200 seeds.}
\label{tab:seven-term-misspec-coverage}
\resizebox{\linewidth}{!}{%
\begin{tabular}{lrrrr}
\hline
Violation & False issuance & Full-path covered & Coverage (95\% Wilson CI) & Mean final radius (95\% $t$ CI)\\
\hline
Target action law & 200/200 & 200/200 & 100.0\% ($[98.1,100]$) & 0.323 ($[0.318,0.328]$)\\
Target context law & 200/200 & 200/200 & 100.0\% ($[98.1,100]$) & 0.324 ($[0.319,0.329]$)\\
Target response & 200/200 & 199/200 & 99.5\% ($[97.2,99.9]$) & 0.317 ($[0.313,0.321]$)\\
Exposure/transport & 200/200 & 200/200 & 100.0\% ($[98.1,100]$) & 0.324 ($[0.320,0.329]$)\\
Dependence envelope & 200/200 & 200/200 & 100.0\% ($[98.1,100]$) & 0.324 ($[0.320,0.329]$)\\
Audit independence & 200/200 & 200/200 & 100.0\% ($[98.1,100]$) & 0.239 ($[0.238,0.240]$)\\
\hline
\end{tabular}}
\end{table}

The observed miss count is one out of 1,200 false certificate issuances, and it occurs in the target-response condition.
This should not be interpreted as evidence that the certificate remains valid under misspecification: the theoretical guarantee does not apply when the stated assumptions are false.
In this particular stress design, the reported intervals are wide (mean final average radius approximately 0.239--0.324), which often preserves empirical coverage despite invalid provenance.
Thus the key failure is not necessarily immediate undercoverage; it is that the interface accepts a false attestation and cannot independently discover a concealed violation.
The width result must be read alongside coverage rather than as a robustness guarantee.

As a control, we separately declared each corresponding required provenance flag invalid.
The implementation rejected all six disclosed-invalid controls (6/6).
This tests the software gate behavior only: it does not show that the gate can detect a false assertion that remains marked valid.

\begin{figure}[!htbp]
\centering
\begin{subfigure}[t]{0.48\linewidth}
\centering
\includegraphics[width=\linewidth]{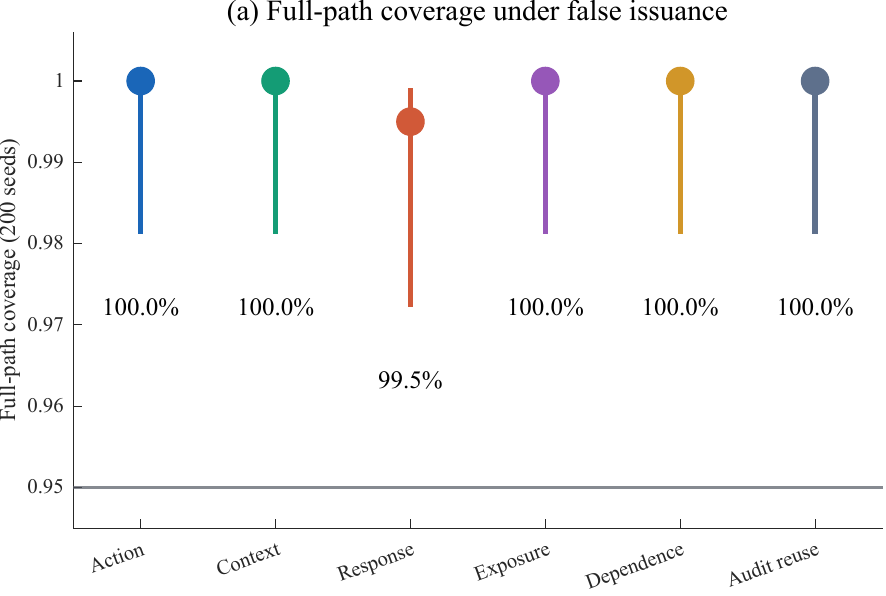}
\end{subfigure}\hfill
\begin{subfigure}[t]{0.48\linewidth}
\centering
\includegraphics[width=\linewidth]{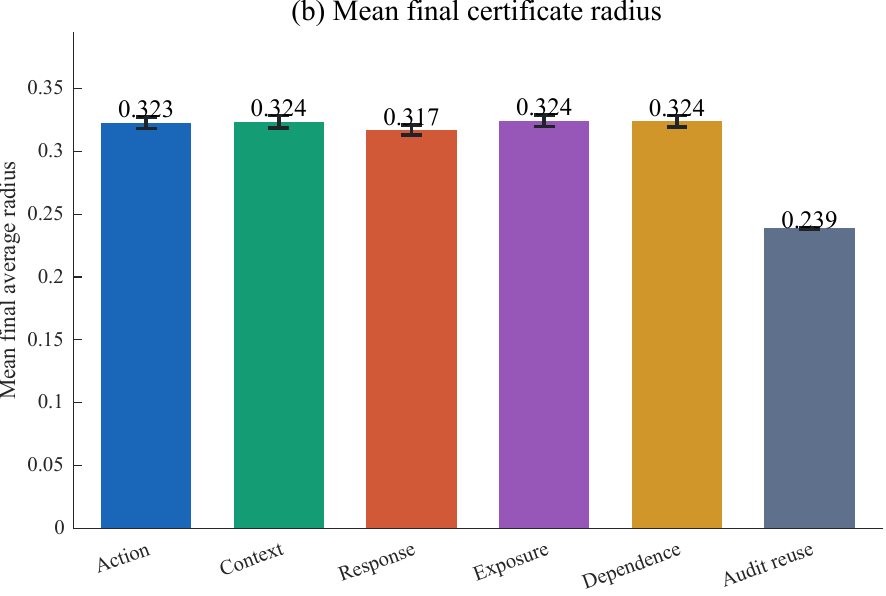}
\end{subfigure}

\caption{\textbf{Certificate diagnostics under hidden assumption violations.}
(a) Full-path coverage conditional on false certificate issuance under six hidden assumption violations, with Wilson 95\% confidence intervals.
(b) Mean final average certificate radius under the same violations, with 95\% Student-$t$ intervals across seeds.}
\label{fig:seven-term-misspec-coverage}
\vspace{-3pt}
\end{figure}

This experiment is a finite-simulation diagnostic, not a universal misspecification theorem.
The severity and form of each perturbation are fixed as stated above; the observed coverage rates cannot be extrapolated to arbitrary violations.
In particular, high empirical coverage under these perturbations is partly attributable to conservative, relatively wide intervals and does not validate the false assumptions.

\FloatBarrier

%% file: appendix/appendix_xxv5_severity_addendum.tex
\subsubsection{Severity of concealed target-law misspecification}
\label{app:seven-term-severity}
We further examine how the empirical validity of a previously issued seven-component certificate responds to the magnitude of an unreported target-law change.
We retain the finite-state certificate interface and the same frozen training, independent audit, and logged-score streams across severity levels.
Only the simulator's target-regime law is changed without updating the attestation.
For each of 64 paired seeds, the online algorithm evaluates three configurations for 120 rounds with 1,200 scoring windows per round.
We prespecify magnitudes $\xi\in\{0,0.10,0.20,0.35,0.50\}$ and test four target-law mechanisms: target-response means, target-action assignment probabilities, target-context mixture probabilities, and the action-specific target/logging response transport.
The same online trajectory is tested against the resulting target value $W^\xi_\infty(z;0)$, so the comparisons isolate a hidden identification failure rather than algorithmic random-seed changes.

\begin{figure}[H]
\centering

\begin{minipage}[c]{0.53\textwidth}
\centering
\captionsetup{font=footnotesize,skip=4pt}

\captionof{table}{Full-path simultaneous target-value coverage for concealed misspecification; 64 paired seeds per scenario (each row uses the same 64 seeds, so the rows are not independent replications).
All incorrect assumptions at positive severity are deliberately asserted valid by the attestation.
Values in parentheses are the maximum population target-value shifts induced by the mechanism.}
\label{tab:seven-term-severity}

\small
\setlength{\tabcolsep}{4pt}
\resizebox{\linewidth}{!}{%
\begin{tabular}{lccc}
\toprule
Hidden target-law change & $\xi=0.20$ & $\xi=0.35$ & $\xi=0.50$ \\
\midrule
Target response & 54/64 (.200) & 1/64 (.346) & 0/64 (.453) \\
Target action & 64/64 (.034) & 64/64 (.059) & 64/64 (.084) \\
Target context & 64/64 (.001) & 64/64 (.002) & 64/64 (.003) \\
Exposure/transport & 64/64 (.098) & 63/64 (.171) & 15/64 (.244) \\
\bottomrule
\end{tabular}%
}
\end{minipage}
\hfill
\begin{minipage}[c]{0.43\textwidth}
\centering
\captionsetup{font=footnotesize,skip=4pt}

\includegraphics[width=\linewidth]{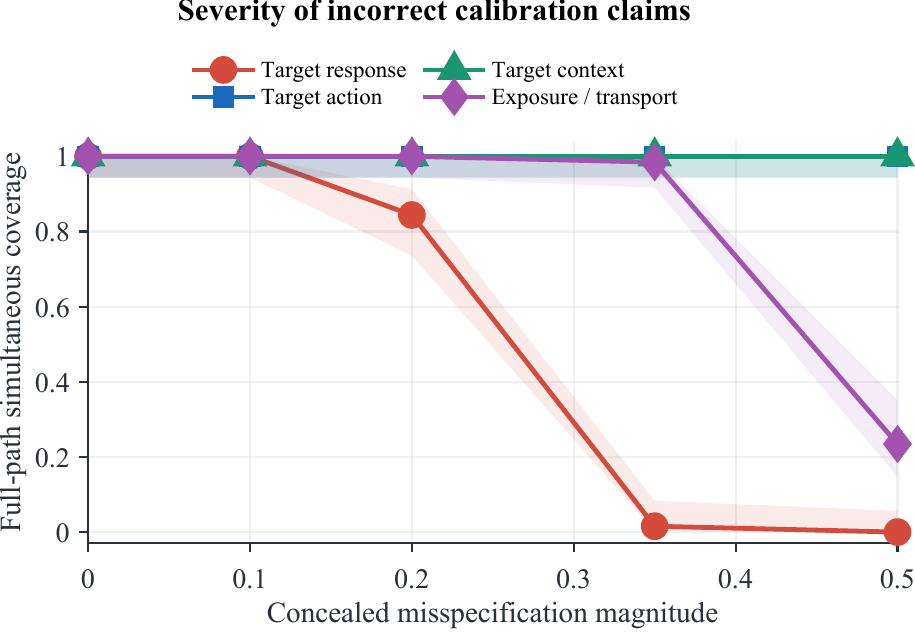}

\captionof{figure}{Sensitivity of the issued seven-component certificates to concealed, predeclared target-law shifts.
Bands are Wilson 95\% intervals across 64 paired seeds; the same seeds are reused across severity levels.
Full-path coverage is simultaneous over all configurations and examined rounds.
The curve does not represent a validity guarantee under misspecification.}
\label{fig:seven-term-severity}

\end{minipage}
\end{figure}

At the largest target-response shift, none of the 64 paths remains covered, whereas the same certificate covers all paths with shifts at most $0.10$.
For exposure/transport shifts of magnitude $0.50$, coverage is $15/64$.
The target-context perturbation leaves the target value almost unchanged (maximum shift $0.0031$), and its high coverage must not be interpreted as robustness to a comparably large target-value error.
The mean final certificate radius is $0.326$ and is unchanged across severity because it is computed from the unmodified attested evidence.
The stress test therefore characterizes the failure boundary of an \emph{incorrectly attested} certificate; it does not weaken the conditional coverage guarantee under correctly verified assumptions in Theorem~1.

%% file: appendix/32_limitations_and_societal_implications.tex
\section{Limitations and Societal Implications}
\label{app:limitations-impacts}

\subsection{Limitations}
\paragraph{Conditional validity of causal certificates.}
The general value certificate depends on correctly specified and independently justified allowances for joint action--exposure overlap, target-law estimation, exposure and response transport, context shift, clipping, and temporal dependence. Observable discrepancy statistics and fitted effective sample sizes are diagnostics, not substitutes for validated bounds. If any required allowance cannot be calibrated, the resulting interval should not be interpreted as an operational coverage guarantee. The anytime finite-sample certificate in Proposition~1 applies only to its stated known-local-interference model; it does not certify the general black-box setting.

\paragraph{Scope and computational burden.}
An incomplete exposure map can omit influential neighboring actions, and action-law correction alone cannot identify the no-exploration steady-state context distribution. Broad interference or weak joint overlap may therefore leave target values insufficiently identified. Evaluating multiple configurations also requires nuisance fitting, joint-exposure weighting, and calibration or auditing; these costs can increase with configuration count, exposure complexity, and dependence. We do not establish a general bound on wall-clock, memory, or audit cost for large production networks.

\paragraph{Limits of empirical generalization.}
The controlled ring comparisons use 24 paired seeds over 200 rounds; the independent coverage audit is a separate protocol; and METR-LA/FlockLab experiments are 180-round semi-synthetic replays with eight seeds per condition and model-generated counterfactual outcomes. They are not live deployment tests or evidence that the general certificate is calibrated on either trace. Under low mismatch, Greedy-Raw can outperform CAUSAL-UCB in target regret; under high mismatch, lower target regret can come with greater measurement-induced deployment loss. Hence neither uniform superiority nor deployment safety follows from these experiments.

\subsection{Potential Societal Impacts and Responsible Deployment}
Correcting the discrepancy between logged exploratory rewards and no-exploration target values could improve the auditability of network-wide configuration choices and make measurement costs more visible. However, in operational networks, exploration itself may degrade service, and optimizing an aggregate utility can conceal disproportionate effects on individual nodes, users, or regions. Misspecified exposure representations or unverified transport assumptions could also produce overconfident decisions. Depending on the application, fine-grained network telemetry may create privacy or security risks.

Before any real-world use, system operators should assess application-specific safety constraints, impose exploration budgets and fallback or rollback rules, examine disaggregated service-quality outcomes, minimize access to sensitive telemetry, and independently validate the certificate's required assumptions and error allowances. These are proposed deployment precautions, not safeguards experimentally implemented or certified in the present study.